\documentclass[mnsc,nonblindrev]{informs3} 

\RequirePackage{amsmath}
\RequirePackage{amssymb}
\RequirePackage{bm} 
\RequirePackage{url}
\usepackage{multirow}
\usepackage{natbib}
\usepackage{graphicx}
\usepackage{subfigure}
\usepackage{makecell}
\usepackage{booktabs}
\usepackage{array}
\usepackage{url}
\usepackage{algorithm}
\usepackage{algorithmic}

\let\hat\widehat
\let\tilde\widetilde

\newcommand{\gb}{\mathbf{g}}
\newcommand{\hb}{\mathbf{h}}

\newcommand{\vb}{\mathbf{v}}
\newcommand{\wb}{\mathbf{w}}
\newcommand{\xb}{\mathbf{x}}

\newcommand{\bS}{\bm{S}}

\newcommand{\bZ}{\bm{Z}}

\newcommand{\cE}{\mathcal{E}}
\newcommand{\cF}{\mathcal{F}}
\newcommand{\cG}{\mathcal{G}}

\newcommand{\cM}{\mathcal{M}}
\newcommand{\cN}{\mathcal{N}}

\newcommand{\GG}{\mathbb{G}}

\newcommand{\RR}{\mathbb{R}}

\newcommand{\bbeta}{\bm{\beta}}

\newcommand{\sign}{\mathop{\mathrm{sign}}}

\newcommand{\overbar}[1]{\mkern 1.5mu\overline{\mkern-1.5mu#1\mkern-1.5mu}\mkern 1.5mu}

\ifx\cond\undefined
\newtheorem{cond}{Condition}
\fi
\usepackage{amsmath,amssymb,bbm}
\usepackage{natbib}
\usepackage{multirow}
\usepackage{subfigure}
\usepackage{makecell}
\usepackage{booktabs}
\usepackage{array}
\usepackage{tabularx}
\usepackage{tabulary}
\usepackage{caption}
\usepackage{booktabs}
\usepackage{url}
\usepackage{algorithm}
\usepackage{algorithmic}
\usepackage{bm}
\usepackage{wrapfig}
\usepackage{lipsum}
\usepackage{mathrsfs}
\usepackage{dsfont}
\usepackage{color}

\usepackage{xr}

\usepackage[usenames,dvipsnames,svgnames,table]{xcolor}
\usepackage[colorlinks,
linkcolor=red,
anchorcolor=blue,
citecolor=blue
]{hyperref}

\newcommand{\nn}{\nonumber}

\def\##1\#{\begin{align}#1\end{align}}
\def\$#1\${\begin{align*}#1\end{align*}}

\def\sn{\sum_{i=1}^n}

\newcommand{\BB}{\mathbb{B}}

\newcommand{\E}{\mathrm{E}}

\usepackage{relsize}
\newcommand{\T}{{\mathsmaller {\rm T}}}

\newcommand{\wt}{\widetilde}

\newcommand{\Rom}[1]{\text{\uppercase\expandafter{\romannumeral #1\relax}}}

\def\bfblue#1{{\color{blue} #1}}

\usepackage{enumitem}
 \newcommand{\cc}{{\rm c}}
\newcommand{\hQ}{\hat{Q}}

\newenvironment{myabstract}[1][Abstract]{\begin{center}\bfseries #1 \end{center}
    \begin{quotation}%
}{%
    \end{quotation}%
}
\newenvironment{mykeywords}{%
    \section*{Keywords:}%
    \begin{quote}%
}{%
    \end{quote}%
}

\OneAndAHalfSpacedXI

\usepackage{natbib}
 \bibpunct[, ]{(}{)}{,}{a}{}{,}%
 \def\bibfont{\small}%
\TheoremsNumberedThrough     
\ECRepeatTheorems

\EquationsNumberedThrough    

\MANUSCRIPTNO{}
\usepackage{fancyhdr}
\footrule
\begin{document}


\RUNAUTHOR{Zhao, Zhou and Wang}

\RUNTITLE{Privacy-Preserving Data-driven Newsvendor}

\TITLE{Private Optimal Inventory Policy Learning for Feature-based Newsvendor with Unknown Demand}

\ARTICLEAUTHORS{%
\AUTHOR{Tuoyi Zhao}
\AFF{Department of Management Science, Miami Herbert Business School, University of Miami, Coral Gables, FL 33146, \EMAIL{txz311@miami.edu}}
\AUTHOR{Wen-Xin Zhou}
\AFF{Department of Information and Decision Sciences, University of Illinois at Chicago, Chicago, IL 60607, \EMAIL{wenxinz@uic.edu}} 
\AUTHOR{Lan Wang}
\AFF{Department of Management Science, Miami Herbert Business School, University of Miami, Coral Gables, FL 33146, \EMAIL{lxw611@miami.edu}}
}

\ABSTRACT{%
The data-driven newsvendor problem with features has recently emerged as a significant area of research, driven by the proliferation of data across various sectors such as retail, supply chains, e-commerce, and healthcare. Given the sensitive nature of customer or organizational data often used in feature-based analysis, it is crucial to ensure individual privacy to uphold trust and confidence. Despite its importance, privacy preservation in the context of inventory planning remains unexplored.  
A key challenge is the nonsmoothness of the newsvendor loss function, which sets it apart from existing work on privacy-preserving algorithms in other settings. This paper introduces a novel approach to estimate a privacy-preserving optimal inventory policy within the $f$-differential privacy framework, an extension of the classical $(\epsilon, \delta)$-differential privacy with several appealing properties.
We develop a clipped noisy gradient descent algorithm based on convolution smoothing for optimal inventory estimation to simultaneously address three main challenges: (1) unknown demand distribution and nonsmooth loss function; (2) provable privacy guarantees for individual-level data; and (3) desirable statistical precision.
We derive finite-sample high-probability bounds for optimal policy parameter estimation and regret analysis.
By leveraging the structure of the newsvendor problem, we attain a faster excess population risk bound compared to that obtained from an indiscriminate application of existing results for general nonsmooth convex loss. Our bound aligns with that for strongly convex and smooth
loss function.
Our numerical experiments demonstrate that the proposed new method can achieve desirable privacy protection with a marginal increase in cost.}

\begin{center}
    \Large
     Private Optimal Inventory Policy Learning for Feature-based Newsvendor with Unknown Demand
    \\[8pt]
    \normalsize 
    Tuoyi Zhao\footnote{ Department of Management Science, University of Miami. txz311@miami.edu.}, Wen-Xin Zhou \footnote{ Department of Information and
Decision Sciences, University of Illinois at Chicago. wenxinz@uic.edu.} and Lan Wang\footnote{ Department of Management Science, University of Miami. lxw611@miami.edu.  The research of Zhao and Wang was partially supported by Grant NSF FRGMS-1952373.}
    \\[8pt]
    \small
\end{center}
\begin{myabstract}
    The data-driven newsvendor problem with features has recently emerged as a significant area of research, driven by the proliferation of data across various sectors such as retail, supply chains, e-commerce, and healthcare. Given the sensitive nature of customer or organizational data often used in feature-based analysis, it is crucial to ensure individual privacy to uphold trust and confidence. Despite its importance, privacy preservation in the context of inventory planning remains unexplored.  
A key challenge is the nonsmoothness of the newsvendor loss function, which sets it apart from existing work on privacy-preserving algorithms in other settings. This paper introduces a novel approach to estimate a privacy-preserving optimal inventory policy within the $f$-differential privacy framework, an extension of the classical $(\epsilon, \delta)$-differential privacy with several appealing properties.
We develop a clipped noisy gradient descent algorithm based on convolution smoothing for optimal inventory estimation to simultaneously address three main challenges: (1) unknown demand distribution and nonsmooth loss function; (2) provable privacy guarantees for individual-level data; and (3) desirable statistical precision.
We derive finite-sample high-probability bounds for optimal policy parameter estimation and regret analysis.
By leveraging the structure of the newsvendor problem, we attain a faster excess population risk bound compared to that obtained from an indiscriminate application of existing results for general nonsmooth convex loss. Our bound aligns with that for strongly convex and smooth
loss function.
Our numerical experiments demonstrate that the proposed new method can achieve desirable privacy protection with a marginal increase in cost.
\end{myabstract}
\rule{\textwidth}{0.4pt}
\begin{mykeywords}
    newsvendor; differential privacy; 
data-driven decision-making;
convolution smoothing; regret analysis.
\end{mykeywords}
\section{Introduction}
\label{sec:1} 

The newsvendor problem, a classical example of the inventory-control problem, is of fundamental importance to operations management. In recent years, there has been growing interest in data-driven feature-based newsvendor problems, due to the vast amount of data generated by retail and supply chains, e-commerce, banking, financial, hospitals, and other business domains. The goal of the feature-based newsvendor problem is to estimate the optimal inventory policy based on the historical demand data as well as the observed features (e.g., product characteristics, customer characteristics) associated with the demand. The ability to determine optimal inventory levels based on features (or contextual information) such as location and usage is essential for supply chain planning. Recently, \citet{ban2019} carefully justified from a theoretical perspective the value of incorporating features in the newsvendor problem when such information is available. They proved that ignoring features may lead to estimation bias, which does not diminish as the number of observations gets large.

In a host of applications, feature-based inventory analysis involves sensitive customer or organizational data. Examples include but are not restricted to the following:
\begin{itemize}
    \item {\it Healthcare}. In hospitals, nurse staffing in the emergency room can be formulated as a feature-based newsvendor problem, see, for example, \citet{he2012timing}, \citet{green2013} and \citet{ban2019}. The features can include the hospital inflow and outflow conditions, surgical case volume, behavioral health patients' boarding information,
    doctor staffing information, and nurses' credentials. As another application, surgical procedures require a large number of consumable supplies that need to be kept in hospital inventory and transported to the operating rooms.  \citet{gorgulu2022newsvendor} formulate the problem of preparing a surgical preference card, a list of items for each surgery, as a newsvendor problem. The features include surgery type, patient disease status, and physicians' past records, among others. Hospitals in general would prefer to keep the related information internal and any physician- and patient-level information private.
    \item {\it E-commerce}. Companies such as Chewey, an online retailer of pet supplies or grocery stores with online ordering and delivery service, often leverage customer-level data for inventory management. This helps not only coordinate shipping from its local warehouses but also design targeted marketing campaigns (e.g., sending coupons to different groups of customers based on predicted individualized inventory levels). 
    \item {\it Finance}. In portfolio management, mutual funds hold a certain percentage of their assets in cash to meet redemption demand from investors. The decision on how much cash to reserve can be formulated as a feature-based newsvendor problem. If not enough cash is held, the fund must sell some of its holdings and will incur transaction costs. In the recent Silicon Valley Bank’s downfall in March 2023, the bank had to sell its securities to raise cash to meet a wave of withdrawals from customers. A strategy of inventory policy can be planned based on the financial and operational variables and clients' behaviors.
\end{itemize}

In the context of the applications discussed above, the preservation of privacy is of critical importance. Despite this, systematic studies in the realm of inventory planning are lacking.
The protection of individual privacy is essential in maintaining customer trust and confidence, and in helping the business avoid financial losses and reputation damage \citep{Williams.2017, fainmesser2022digital, hu2022privacy}. The prevalence of individual-level data and the increased awareness of privacy concerns motivate us to develop a principled privacy-preserving framework for data-driven feature-based newsvendor problems with unknown demand.  We focus on the scenario where there is a trusted curator, such as the company's in-house business analytics team, responsible for processing and analyzing the data. The primary objective of this paper is to develop a data-driven approach that generates valuable outputs.
These outputs assist a decision maker (hereafter referred to as ``DM") in estimating the optimal inventory level, all while safeguarding the privacy of historical individual-level data. 
In essence, our approach controls the likelihood of an adversary making harmful inferences about a data subject based on a differentially private data release, ensuring it remains a small probability event.

\begin{figure}[ht]
\FIGURE
  {\includegraphics[width=0.6\linewidth]{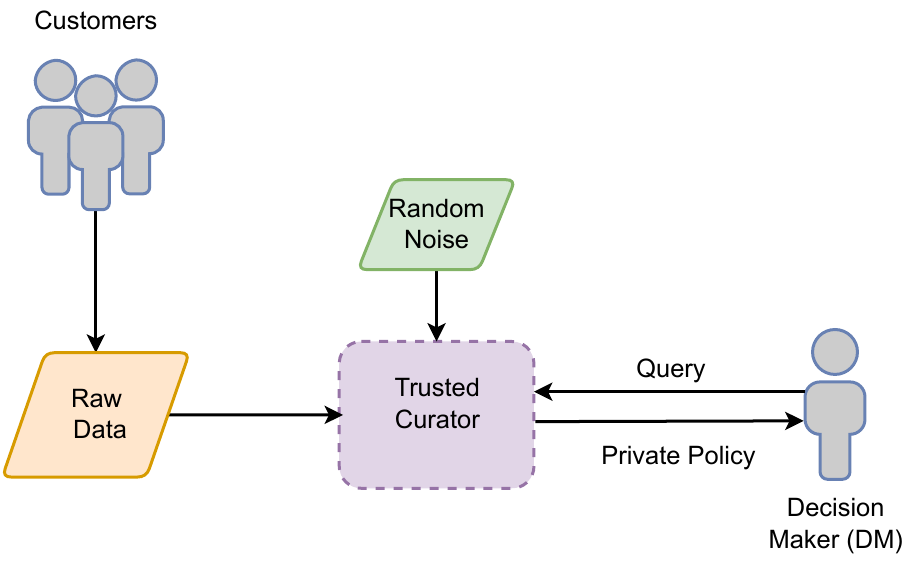}}
  {Illustration of privacy protection\label{Fig_privacy}}
{} 
\end{figure}

In more detail, we study a feature-based newsvendor problem where the demand distribution is unknown to the DM. The trusted curator has access only to $n$ past records (historical data) she collects $\{d_i, \BFx_i\}$, $i=1,\ldots,n$ where $d_i$ is the observed demand, and $\BFx_i$ is the associated vector of features (or covariates). 
When presented with a new query, the curator utilizes a data-driven iterative algorithm to 
release a private output. This output guides the DM in determining the optimal inventory level, while ensuring that the sensitive information from the historical data remains non-inferable from the output. The iterative algorithm introduces a carefully tuned amount of random noise
to the statistical outputs, aiming to strike a balance between privacy protection and statistical accuracy. See Figure~\ref{Fig_privacy} for an illustration.

Distinct from prior work on privacy-preserving algorithms in other business applications, we 
confront the challenge of the nonsmoothness of the newsvendor loss function. In this setting, we 
leverage the recently introduced concept
of $f$-differential privacy \citep{dong2022gaussian} and propose a noisy clipped gradient descent algorithm based on convolution smoothing for optimal inventory estimation. The new approach simultaneously addresses the three main challenges: (1) unknown demand distribution and nonsmooth loss function; (2) provable privacy guarantees for individual-level data; and (3) desirable statistical precision. Importantly, our theoretical and numerical results demonstrate that a reasonable degree of privacy protection can be achieved with
minimal sacrifice of data utility, particularly when the size of the historical dataset is large.

\subsection{Contributions}
Our major results and contributions are summarized as follows.

\subsubsection{Provable privacy-protection guarantee in the $f$-differential privacy framework.}
 
 To establish rigorous privacy protection properties, we adopt a recently introduced novel privacy framework named $f$-differential privacy ($f$-DP, \citet{dong2022gaussian}), which generalizes the classical $(\epsilon, \delta)$-DP notion \citep{dwork2006our,dwork2006calibrating} with several attractive properties, see Section 4.1 for a more in-depth introduction. The $(\epsilon, \delta)$-DP notion was proposed by the computer science community and has become a popular framework for provable privacy protection against arbitrary adversaries while allowing the release of analytical summaries. It provides a statistical hypothesis testing interpretation for differential privacy, thereby making the privacy guarantees easily understandable. Despite its great success, a major shortcoming of
 $(\epsilon, \delta)$-DP is its inability to tightly handle composition (a.k.a. repeated application of the mechanism to the same data set). Using the output of an $(\epsilon, \delta)$-DP
mechanism, the power of any $\alpha$-level test is bounded by $e^{\epsilon}\alpha+\delta$. Recall that the composition of $(\epsilon_1,\delta_1)$- and $(\epsilon_2,\delta_2)$-DP mechanisms results in an $(\epsilon_1 + \epsilon_2, \delta_1 + \delta_2)$-DP mechanism. The resulting power bound $e^{\epsilon_1+\epsilon_2} \alpha + \delta_1 +\delta_2$ of any $\alpha$-level test no longer tightly characterizes the trade-off between significance level and power.
As a fundamental observation, \cite{dong2022gaussian} pointed out that $(\epsilon, \delta)$-DP is mis-parameterized in the sense that the guarantees of the composition of $(\epsilon_i, \delta_i)$-DP mechanisms cannot be characterized by any single pair of parameters $(\epsilon, \delta)$. Many recent efforts have been devoted to developing relaxations of DP for which composition can be handled exactly. These notions of DP no longer have hypothesis testing interpretations; rather, they are based on studying divergences that satisfy a certain information processing inequality. We refer to Section 1 of \cite{dong2022gaussian} for an in-depth discussion on this matter.


The main idea of the $f$-DP is the usage of the so-called trade-off functions as a measure of indistinguishability of two neighboring data sets rather than a few parameters as $(\epsilon, \delta)$-DP and other prior relaxations do. It preserves the hypothesis testing interpretation of differential privacy. Furthermore, it captures all the desirable properties of prior differential privacy definitions, in particular, composition, amplification by sampling, and Gaussian mechanism, tightly and analytically. It provides a powerful technique to import existing results proven for the $(\epsilon, \delta)$-DP to $f$-DP. 

 In the powerful and versatile $f$-DP framework, we rigorously establish that our data-driven algorithm provides the desired privacy guarantees. 

\subsubsection{A computationally efficient algorithm to estimate the feature-based optimal inventory policy with unknown demand function.}

Traditionally, the newsvendor problem is solved based on the assumption that the demand distribution is known up to a small number of parameters. The commonly used data-driven estimation procedures often consist of two steps: the first step estimates the parameters using the observed data and the second step performs the optimization to estimate the optimal order quantity.  However, in reality, the true demand distribution is hardly ever known to the DM. 

For the data-driven feature-based newsvendor problem, \citet{ban2019} proposed a one-step estimation procedure based on empirical risk minimization (ERM) and established its connection to quantile regression \citep{KB1978}. In terms of computation, \citet{ban2019} reframed the ERM problem as a linear program and utilized existing linear programming solvers. These general-purpose solvers are capable of generating solutions with high precision (low duality gap). However, in the context of machine learning, this is inefficient for two reasons. First, generic toolboxes are often unaware of the problem structure and tend to be too slow or encounter memory issues. Secondly,  high precision is not always necessary for machine learning problems, and a duality gap of the order of machine precision may not be required \citep{Bachetal2012}. More importantly, commonly used algorithms for solving linear programs, such as simplex-based methods and interior point methods, may not be readily adaptable for privacy preservation purposes.

To address the computational and privacy concerns mentioned above, we propose a new approach that utilizes convolution-smoothing \citep{fernandes2021smoothing,HPTZ2020}. This approach transforms the non-differentiable newsvendor loss function into a twice-differentiable, convex, and locally strongly convex surrogate, allowing for fast and scalable gradient-based algorithms for optimization. Additionally, to ensure privacy protection while maintaining computational tractability, inspired by \cite{SCS2013}, \cite{BST2014} and \cite{lee2018concentrated}, we adopt a noisy optimization approach by adding Gaussian noise to the gradient of the smoothed empirical cost in each iteration. By carefully selecting the scale of the added noise and the number of iterations, we can achieve the desired privacy level along a sequence of outputs. The algorithm is designed for efficient implementation, and in this paper, we provide both privacy-protection guarantees and statistical accuracy guarantees for the output of this novel algorithm.

\subsubsection{Finite-sample performance bounds \bfblue{and excess risk} analysis.}

Under a linear demand model with an unknown error distribution and a potentially large number of features, we analyze the convergence of the proposed algorithm and its finite sample performance error bounds. We also derive its regret bound, which is the difference between its expected cost and the optimal cost of the clairvoyant who knows the underlying demand distribution. The regret upper bound is of the order $O(\log (n)\max\{(\frac{p+\log n}{\mu n})^2, \frac{p}{n}\} )$, where $n$ is the size of the historical data set, $p$ is the number of the features, $\mu$ is the privacy parameter. As we discuss in Section~\ref{intro_privacy}, $\mu=0.5$ (or less) indicates a reasonable degree of privacy protection in practice. The term $O( \log(n)(\frac{p+\log n}{\mu n})^2 )$  corresponds to the additional regret due to privacy protection, which goes to zero quickly as $n$ gets large for a reasonable choice of $\mu$. The theory and our numerical results suggest that privacy protection can be achieved with a reasonably small additional cost.

The idea of convolution smoothing was initially proposed in the optimization community by \cite{CM1995, CM1996}, where sigmoid functions were used as smooth approximations of the plus function $\max\{ x, 0\}$. However, the impact of smoothing on statistical performance, in terms of estimation error bounds or regret bounds, remained largely unknown until recent studies in the context of quantile regression by \cite{fernandes2021smoothing}, \cite{tan2022high} and \cite{HPTZ2020}.

In this paper, we conduct a comprehensive analysis of the noisy clipped gradient descent iterates, as opposed to the hypothesized empirical risk minimizer. By exploring the specific structure of the newsvendor problem, we achieve a faster excess population risk bound compared to the results obtained by indiscriminately applying existing results developed for general nonsmooth convex loss. Our bound matches with what one would obtain when the loss function is both strongly convex and smooth. The combination of carefully selected smoothing and noise scale parameters allows for control over the trade-off between statistical efficiency and the level of privacy. A key aspect of our analysis is a novel characterization of the local strong convexity and smoothness of the smoothed cost function, which subtly depends on the order of the smoothing parameter. The technical devices we employ in this paper to establish the theoretical framework are distinct from earlier works, such as those presented in \citet{ban2019}, and yield sharper results, as elaborated in Section~\ref{sec:5}. Furthermore, we relax the i.i.d. error condition in their paper. By allowing the error distribution to be heteroscedastic, we permit the features to influence not only the location of the demand distribution but also its dispersion. Furthermore, we do not require the error distribution to be bounded.

\subsection{Notation and Organization}

The following general notation will be used throughout the paper. We use $\BFI_p$ to denote the $p\times p$ identity matrix. For a vector $\BFu \in \mathbb{R}^p$ ($p\geq 2$), we write $\| \BFu \|^2 = \BFu^\T \BFu$. For a positive definite matrix $\BFA$, we write $\| \BFA \|_2 = \max_{\BFu: \| \BFu\|_2 = 1} \| \BFA \BFu \|_2$ and $\| \BFu \|_{\BFA} = \sqrt{\BFu^\T \BFA \BFu}$. We use $\mathbb{S}^{p-1}$ to denote the unit sphere in $\mathbb{R}^p$, that is, $\mathbb{S}^{p-1} = \{ \BFu \in \mathbb{R}^p: \| \BFu \|_2 =1 \}$. For two sequences of positive numbers $\{ a_n \}_{n\geq 1}$ and $\{ b_n \}_{n\geq 1}$, we write $a_n \lesssim b_n$ if there exists some constant $C>0$ independent of $n$ such that $a_n \leq C b_n$ for all $n$; write $a_n \gtrsim b_n$ if $b_n \lesssim a_n$, and write $a_n \asymp b_n$ if $a_n\lesssim b_n$ and $b_n \lesssim a_n$. For an event or set $E$, let $\mathbbm{1}(E)$ or $\mathbbm{1}\{ E \}$ denote the indicator function.

The paper is organized as follows. In Section 2, we review the related literature. Section 3 presents the model and introduces the underlying assumptions. In Section 4, we introduce the basics of $f$-differential privacy, present the new privacy-preserving algorithm for the feature-based newsvendor problem, and provide theoretical justifications for privacy-preserving guarantees. In Section 5, we provide high-probability bounds for the estimated private parameter indexing the optimal inventory policy and
the regret analysis. We analyze the performance of our approach through an extensive numerical study and a real data example in Section 6. Section 7 contains some concluding remarks. The technical details are given in the appendices.

\section{Related Review}
We briefly review related research in data-driven newsvendor problems and differential privacy for operations management.

\noindent
{\it Data-driven newsvendor problem with unknown demand}.
Earlier work on newsvendor problems often assumes the demand distribution is known. There has also been extensive literature on relaxing the known demand distribution assumption but without using any feature information. \citet{ban2019} provided an excellent literature review and broadly characterized these methods
into three categories: the Bayesian approach, the minimax approach, and the data-driven approach. Our proposed method is more closely related to the data-driven approach where the DM uses the observed sample to make decisions, see \cite{burnetas2000adaptive,  godfrey2001adaptive, powell2004learning, levi2007provably, kunnumkal2008using, huh2009nonparametric, levi2015data}, among others. Related to this line of work, \cite{liyanage2005practical, hannah2010nonparametric,see2010robust, beutel2012safety, ban2019, oroojlooyjadid2020} incorporated feature-based information. \citet{ban2019} provided a systematic study on the benefits of incorporating features, proposed new algorithms, and derived performance bounds. \citet{oroojlooyjadid2020} considered a deep-learning approach.

Our approach differs from the aforementioned work in several major aspects. First, our algorithm provides individual-level data privacy protection. To the best of our knowledge, this is the first time in the literature of newsvendor problems the issue of privacy is systematically investigated. Second, we provide a theoretical error bound for the $T$-step output of the proposed algorithm directly. In contrast, the theory of the early work is for the limiting (or theoretical) solution of their proposed algorithms. Third, we substantially relax the technical conditions on the random error distribution for the theory compared with the earlier work.

\noindent
{\it Differential privacy for operations management}. In the last decade, privacy preservation has 
received substantial attention in theoretical computer science, database, and cryptography literature. There exist different notions of privacy. {\it Differential privacy}, a seminal concept introduced in \citet{dwork2006our,dwork2006calibrating}, has emerged as the foundation to develop a rigorous framework
for provable privacy protection against arbitrary adversaries. The most commonly used form of 
differential privacy relies on two parameters $\epsilon\geq 0$ and $0\leq \delta\leq 1$, and is often also referred to as the $(\epsilon, \delta)$-differential privacy. This concept has an intuitive hypothesis interpretation. Suppose an attacker would like to distinguish two neighboring data sets which differ by only one observation. Formulated as a hypothesis testing problem, accepting the null hypothesis means
the attacker cannot tell the two data sets apart. Then for any level $\alpha$ test ($0<\alpha<1$) based on the output of a privacy-preserving algorithm satisfying $(\epsilon, \delta)$-differential privacy, its power (a.k.a. the probability of rejecting the null hypothesis when the two data sets are different)
is upper bounded by $e^{\epsilon}\alpha+\delta$. Moreover, $(\epsilon, \delta)$-differential privacy is immune to postprocessing, that is, combining two differential private algorithms preserves differential privacy. Although $(\epsilon, \delta)$-differential privacy provides an elegant formalism for privacy protection, it is known to suffer from the major drawback that it does not tightly 
handle composition. This makes it challenging to provide a tight analysis of the cumulative privacy loss over multiple computations thus limiting its applicability to practically useful privacy-preserving algorithms which often involve injecting privacy protection into different modules and iterative steps. 

Although several relaxations of the $(\epsilon, \delta)$-differential privacy have been proposed, they do not handle well fundamental primitives associated with differential privacy such as privacy amplification by subsampling.
This motivated us to adopt a recently proposed 
new notion of $f$-differential privacy  \citep{dong2022gaussian} which 
extends $(\epsilon, \delta)$-differential privacy and overcomes the above limitations. Similarly to $(\epsilon, \delta)$-differential privacy, $f$-differential privacy characterizes privacy preservation from the hypothesis testing perspective. Rather than using a pair of parameters $(\epsilon, \delta)$ to balance between type I and type II errors, $f$-differential privacy uses a trade-off function. This functional extension of differential privacy avoids the drawbacks mentioned above. We refer to Section~\ref{intro_privacy} for more detailed discussions on the properties of $f$-differential privacy.
The notion of $f$-differential privacy was recently published as a discussion paper in the leading statistical journal {\it Journal of the Royal Statistical Society, Series B}. 
One of the discussants wrote ``One can expect the latter ($f$-differential privacy) to become
a dominant approach in this literature given its appealing intuitive hypothesis-testing interpretation,
exact composition property, the central limit role for composition, and computational tractability
for approximating privacy losses."

\noindent
{\it Differentially private convex optimization}. Our work is also related to the literature on differentially private convex optimization. Differentially private empirical risk minimization (ERM) is a well-studied area. The earlier popular approaches include output perturbation \citep{chaudhuri2008privacy, wu2017bolt} and objective Perturbation \citep{chaudhuri2008privacy, chaudhuri2011differentially, abadi2016deep, jain2014near, iyengar2019towards, slavkovic2021perturbed}.

Motivated by \cite{SCS2013} and \cite{BST2014}, we consider a noisy gradient descent algorithm. 
Differential privacy with various types of gradient descent algorithms have been studied by
\cite{SCS2013}, \cite{BST2014}, \cite{bassily2019private}, \cite{balle2020privacy}, \cite{lee2018concentrated}, \cite{wang2017differentially}, and \cite{wang2018revisiting}, among others.
The methods in \citet{BST2014}, \citet{bassily2019private}, \citet{feldman2020private} and others do not directly apply to our setting. Most of the prior work requires strong convexity and other smoothness
 conditions that are not satisfied by the newsvendor loss function. In the case for which Lipschitz continuity suffices, the known excess loss rate is suboptimal in our setting as they do not explore the specific structure for the newsvendor loss as we do. Unlike the earlier literature, we do not assume the gradient is bounded by a constant and carefully analyzed a clipped DP gradient descent algorithm.
In the statistical literature, \cite{ABL2021} recently investigated optimization-based approaches for Gaussian differentially private $M$-estimators. However, the objective function in our setting does not satisfy the local strong convexity and smoothness in their paper. However, our proof technique deviates from theirs. We will discuss the main challenges and our proof strategies in Section \ref{sec:5.3}.
Despite the nonsmooth newsvendor loss, our novel analysis based on restricted strong convexity and smoothness leads to a faster excess population risk rate, which is only obtained in Section~5 of \cite{feldman2020private} when the loss function is both $\lambda$-strongly convex and $\beta$-smooth.
From this perspective, our results and proof techniques bring new insights and results to the differential private convex optimization literature.


Our work is also related to the growing but still limited literature on privacy preservation in operations management. \citet{chen2022privacy} addressed privacy preservation for personalized pricing with demand following a generalized linear model. They proposed the new notion of anticipating $(\epsilon, \delta)$-differential privacy that is tailored to the dynamic pricing problem. 
\citet{lei2020privacy} and \citet{chen2022differential} considered 
personalized pricing using the notion of central differential privacy and local differential privacy.
The aforementioned papers do not face the challenge of nonsmooth loss function as we have here.
However, our work is not a mere application of existing results on private nonsmooth convex optimization, nor does it utilize standard arguments such as uniform stability. Furthermore,
in contrast to these existing works, we adopt the $f$-differential privacy framework 
to study the performance of the new privacy-preserving algorithm and establish its provable privacy protection guarantees.

\section{Problem Formulation}

\subsection{Feature-based Newsvendor Problem}
We consider the classical single-period newsvendor problem setting. The DM needs to determine the ordering level $q$ based on the observable demand $d$ and feature vector $\BFx$.
Both $d$ and $\BFx$ are random. We assume that the distribution of the demand $d$ is unknown. In the feature-based newsvendor problem, given a realization of the feature vector
$\BFx \in \mathbb{R}^p$, the DM sets the ordering level $q(\BFx)$ to
minimize the conditional expected cost function
\begin{equation*}
    \E \{ C(q(\BFx), d ))|\BFx \} =\E\{[h(q(\BFx)-d )^{+}+b( d -q(\BFx))^{+}]|\BFx\},
\end{equation*}
where $h$ is the per-unit holding cost, $b$ is the per-unit lost-sales penalty cost, $t^{+}=\max\{t,0\}$, and the expectation is taken with respect to the conditional distribution of $d$ given $\BFx$.

Similarly to \citet{beutel2012safety} and \citet{ban2019}, we consider a linear decision function of the
 form 
 \begin{equation*}
   q(\BFx)=  \BFx^\T   \BFbeta  = \sum_{j=1}^p x_{j} \beta_j, 
 \end{equation*}
 where the $p$-dimensional feature vector $\BFx= (x_{1},\ldots, x_{p})^\T$ has $x_{1}\equiv 1$, and $\BFbeta = (\beta_1, \ldots, \beta_p)^\T$ is the coefficient vector. Considering the linear decision space is not a restriction in theory or practice. By replacing the features with their transformations (e.g., via series functions), the framework can be adapted to accommodate nonlinear decision rules. More specifically, one may approximate a nonlinear function $\BFx \mapsto q(\BFx)$ by linear forms $\BFz(\BFx)^\T \BFbeta$, where $\BFx \mapsto \BFz(\BFx) := (z_1(\BFx), \ldots, z_k(\BFx))^\T$ is a vector of approximating functions and $k = k_n \geq 1$ may increase with $n$. Then, we denote the transformed features as $\{\BFz_i = \BFz(\BFx_i)\}_{i=1}^n$. Popular choices of the series approximating functions include B-splines (or regression splines), polynomials, Fourier series, and compactly supported wavelets. We refer to \cite{N1997} and \cite{C2007} for a detailed description of these series functions. Under the linear decision model, we define the parameter indexing the optimal decision rule as
\begin{equation}
\BFbeta^*=\argmin_{\BFbeta \in \mathbb{R}^p  }  \, C(\BFbeta) := \E \big\{ h( \xb^\T \BFbeta-d )^{+}+b( d - \xb^\T \BFbeta )^{+}  \big\}   , \label{def:true.beta}
\end{equation}
where the expectation is taken with respect to the joint distribution of $(d, \xb)$. 
Write $\varepsilon =d- \BFx^\T \BFbeta^*$.
Then the linear decision function (\cite{ban2019}) is
equivalent to assuming that the conditional $b / (b+h)$ quantile of $\varepsilon$ given $\BFx$ is zero. Unlike \cite{ban2019}, we do not assume independence between $\xb$ and $\varepsilon$.

\subsection{Convolution Smoothing for Empirical Risk Minimization}

In the case without features, it is well known that the
optimal decision is given by the $b / (b+h)$ quantile of the demand distribution. It can be estimated by the data-driven sample average approximation (SAA) \citep{levi2015data},
an approach without making any parametric distributional assumption on $d$.  In the setting with features, \citet{ban2019} extended it
to the conditional case, proposed a linear programming-based empirical risk minimization algorithm (NV-ERM), and established the connection to conditional quantile regression. 
More explicitly, one can rewrite
$C(q(\BFx), d)=(b+h)\rho_{\tau}(d -q(\BFx))$,
where $\tau= b / (b+h)$, and $\rho_\tau(u) = u \{ \tau - \mathbbm{1}(u<0)\}$ is referred to as the 
quantile loss function or the check function
corresponding to the $\tau$-th quantile \citep{KB1978}.

For an arbitrary $\BFbeta \in \RR^p$, let 
$\varepsilon_i(\BFbeta) := d_i -  \BFx_i^\T \BFbeta$,
$i=1,\ldots, n$. Consider the empirical cumulative distribution function (ECDF) of the $\varepsilon_i(\BFbeta)$: $\hat F(u;\BFbeta)
=(1/n) \sn \mathbbm{1}\{ \varepsilon_i(\BFbeta)  \leq u \}$, $u \in \RR$.
Then, the 
empirical risk minimization approach (NV-ERM) of
\citet{ban2019} minimizes the following empirical loss function
\#
\hat C(\BFbeta) = (b+h)\int_{-\infty}^\infty \rho_\tau(u) \,{\rm d} \hat F(u;\BFbeta) , \label{empirical.qr}
\#
which can be solved via a linear program reformulation. The empirical loss function is nonsmooth
and poses significant challenges to developing a privacy-protection procedure.
To come up with an efficient algorithm
to estimate the optimal decision with 
provable privacy-preserving  guarantees, we
adopt convolution smoothing to address the challenge associated with the non-differentiability of
the loss function. This aims to simultaneously 
achieve two goals: (1) to have a feasible algorithm with 
privacy-preserving guarantee; (2) to have
an algorithm with a statistical accuracy guarantee
as measured by the accuracy of estimating $\BFbeta^*$
and the regret, which is the cost gap
between the estimated policy from the algorithm and the clairvoyant benchmark.

The idea of convolution smoothing originates from \cite{CM1995, CM1996} in a special case, and has been re-examined from a statistical perspective by \cite{fernandes2021smoothing}.
Specifically, let
$\hat F_{\varpi}(\cdot;\BFbeta)$ be a 
smoothed estimator of the distribution function of $\varepsilon_i(\BFbeta)$ based on the classical Rosenblatt--Parzen kernel density estimator. That is, for $u\in \RR$,
\#
	\hat F_{\varpi}(u;\BFbeta) = \int_{-\infty}^u \hat f_{\varpi}(t;\BFbeta) \,{\rm d}   t~~\mbox{ with }~~ \hat f_{\varpi}(t;\BFbeta) = \frac{1}{n} \sn K_{\varpi} \big(  t-\varepsilon_i(\BFbeta) \big) , \nn
\#
where $K_{\varpi}(u) := (1/{\varpi}) K(u/{\varpi})$, 
$K:\mathbb R \to \mathbb [0,\infty)$ is a symmetric, non-negative kernel that integrates to one,  and ${\varpi}={\varpi}_n >0$ is a smoothing parameter. We consider the following smoothed counterpart of $\hat C(\BFbeta)$:
\begin{equation}
	\hat C_{\varpi}(\BFbeta) :=   (b+h)\int_{-\infty}^\infty \rho_\tau(u) \,{\rm d} \hat F_{\varpi}(u;\BFbeta) = \frac{b+h }{n } \sn  (\rho_\tau * K_\varpi) (d_i - \BFx_i^\T \BFbeta) ,  \label{smooth.qloss}
\end{equation}
where ``$*$" denotes the convolution operator that for any two measurable functions $f$ and $g$, $(f * g)(u) = \int_{-\infty}^\infty f(v) g(u-v) {\rm d} v$. Therefore, $\hat C_{\varpi}$ is also referred to as the convolution-smoothed loss/cost function. Commonly used kernel functions in optimization and statistics include (i) Gaussian kernel $K(u)=(2\pi)^{-1/2}e^{-u^2/2}$, (ii) Laplacian kernel $K(u) = e^{-|u|}/2$, (iii) logistic kernel $K(u) =e^{-u}/(1+e^{-u})^2$, (iv) uniform kernel $K(u) =(1/2)\mathbbm{1}(|u|\leq 1)$, and (v) Epanechnikov kernel $K(u) = (3/4) (1-u^2)\mathbbm{1}(|u|\leq 1)$. The following lemma shows that, given any symmetric kernel $K$ and smoothing parameter $\varpi>0$, the resulting smoothed loss $\rho_\tau *K_\varpi$ provides an upper approximation of  $\rho_\tau$ with uniform approximation error that scales with $\varpi$.

\begin{lemma} \label{lem:loss.approxi}
Let $K$ be a symmetric, non-negative kernel function with $\kappa_1 := \int_{-\infty}^\infty |u| K(u) {\rm d}u <\infty$. For any $\varpi >0$, it holds uniformly over $u\in \mathbb{R}$ that $\rho_\tau(u) \leq (\rho_\tau *K_\varpi) (u) \leq \rho_\tau(u) + \kappa_1 \varpi /2$.
\end{lemma}

From Lemma~\ref{lem:loss.approxi} we see that uniformly over $\BFbeta \in \mathbb{R}^p$,
$\hat C (\BFbeta) \leq  \hat C_{\varpi}(\BFbeta) \leq   \hat C (\BFbeta)  + 0.5 \kappa_1(b+h)  \varpi$. The analysis of the discrepancy between the respective minimizers necessitates a more nuanced examination and additional assumptions on the data-generating process. In the following subsection, we elucidate the assumptions necessary for the analysis of the privacy-preserving algorithm proposed for feature-based newsvendor problems. We also discuss their connections with the assumptions imposed in \cite{ban2019}.

\begin{figure}[ht]
\FIGURE
{\includegraphics[width=0.5\linewidth]{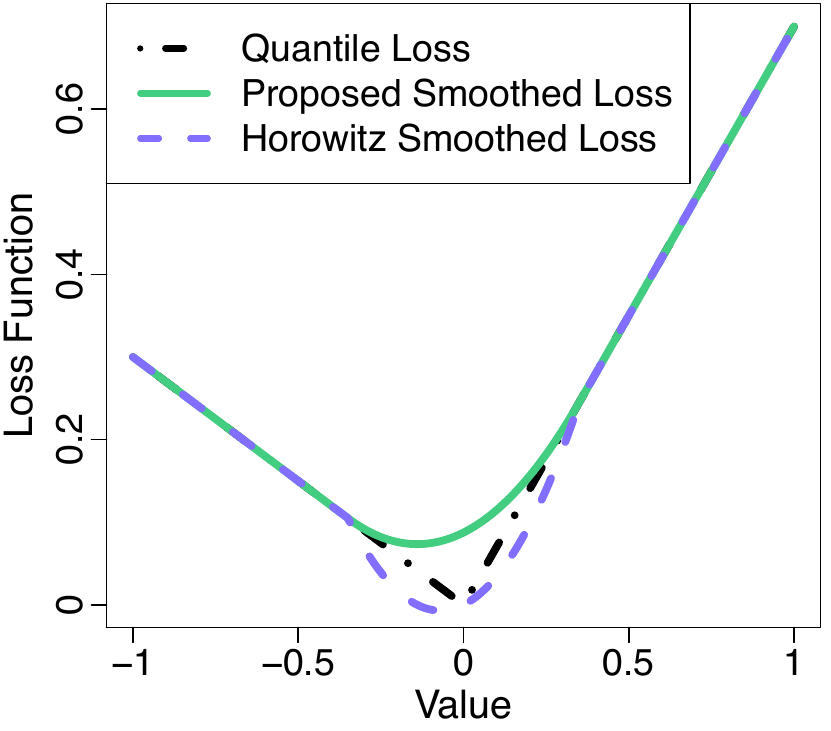}}
  {Illustration of two smoothed check/quantile loss functions \\ \label{Fig_loss}}
{} 
\end{figure}

\medskip
\begin{remark}
To address the non-differentiability (at the origin) of the check loss, \cite{H1998} proposed a more direct approach by replacing the indicator function $\mathbbm{1}(u<0)$ in $\rho_\tau(u)$ with $G(-u/\varpi)$, where $G(\cdot)$ is a smooth, non-decreasing function that takes values between 0 and 1, and $\varpi>0$ is a smoothing parameter. However, Horowitz’s smoothing gains smoothness at the cost of convexity, which inevitably raises optimization challenges, 
particularly when dealing with a large number of features is large. On the contrary, provided a non-negative kernel is used, the convolution-smoothed loss $\rho_\tau * K_\varpi$ remains convexity, as its second derivative $(\rho_\tau * K_\varpi)^{(2)}(u) = K_\varpi(u) = K(u/\varpi) /\varpi$ is everywhere non-negative.  See Figure~\ref{Fig_loss} below for a visualization of Horowitz’s and convolution-smoothed check losses.

Using an approximate Hessian matrix, \cite{CLZ2019} proposed a Newton-type algorithm to solve Horowtiz's smoothed empirical loss minimization problem. In their convergence analysis, they assumed the initial estimator to be consistent, albeit at a sub-optimal rate. In contrast, our analysis imposes minimal assumptions on the initial value, as demonstrated in Theorems~\ref{thm:convergence} and \ref{thm:initial.convergence} in Section~\ref{sec:5.3}. To achieve differential privacy in their algorithm, it may be necessary to inject noise into both the gradient and the approximate Hessian of the empirical loss function. However, the theoretical analysis of their algorithm in the presence of inconsistent initial estimates is currently unknown.
\end{remark}

\medskip

\begin{remark}
The choice of the smoothing technique has an impact on the optimality of the algorithm.
Section A of \citet{BST2014} considered an example in the setting of hinge loss with 
Huberization method (quadratic smoothing)  and argued that it does not allow one to get the optimal excess risk bounds. An alternative popular smoothing method for minimizing a non-differentiable convex objective $\BFbeta \mapsto \hat C(\BFbeta)$ is through Moreau-Yosida inf-convolution (or envelope) \citep{Nesterov2005}. Recall that $ \rho_\tau(u) = \tau u \mathbbm{1}(u > 0) - (1-\tau) u\mathbbm{1}(u \leq  0)$. Its Moreau-Yosida envelop is given by
$$
    \rho_\tau^\gamma(u) = \min_{v\in \mathbb{R}} \bigg\{ \rho_\tau(v) + \frac{1}{2\gamma} (u- v)^2 \bigg\},
$$
where $\gamma>0$ is the regularization parameter. It follows from standard calculations that 
\begin{align} \label{MY.smoothed.loss}
   \rho_\tau^\gamma(u) = \begin{cases}
    - (1-\tau) u - \frac{1}{2}(1-\tau)^2 \gamma ~&\mbox{ if } u <  - (1-\tau)\gamma , \\
    \frac{1}{2\gamma}u^2 , ~&\mbox{ if }   u \in [  - (1-\tau) \gamma ,  \tau \gamma ],  \\
    \tau u  - \frac{1}{2} \tau^2 \gamma ~&\mbox{ if }  u >   \tau \gamma
    \end{cases}
\end{align}
and $( \rho_\tau^\gamma)'(u) = \tau \mathbbm{1}(u > \tau\gamma) - (1-\tau) \mathbbm{1}\{ u < - (1-\tau)\gamma \}  + \frac{1}{\gamma} u \mathbbm{1}\{   - (1-\tau)\gamma \leq  u \leq \tau \gamma \}$. 
When the convex loss is nonsmooth, i.e., not everywhere differentiable, \cite{bassily2019private} proposed a variant of the noisy stochastic gradient descent (SGD) algorithm and established upper bounds on the expected excess risk. Applying their general results to the check loss $\rho_\tau$, it follows that the expected excess risk of the output from their $(\epsilon, \delta)$-DP algorithm is of order 
$$
    O\Bigg( \sqrt{\frac{p}{n}} + \sqrt{\log(1/\delta)} \frac{p}{\epsilon n} \Bigg).
$$
In Section~\ref{sec:5.2}, by proving a form of restricted strong convexity that holds with high probability, we will show that the expected excess risk of the proposed noisy GD estimator satisfies a faster rate, which is of the order
$$
O \Bigg( \log(n) \bigg\{ \frac{p}{n} + \bigg(\frac{p+\log n}{\mu n} \bigg)^2 \bigg\} \Bigg) ,
$$
where $\mu>0$ is the privacy parameter for the Gaussian mechanism.
\end{remark}

\subsection{Assumptions}
\label{sec:3.3}

Suppose we observe an i.i.d. sample $\{ (d_i, \BFx_i) \}_{i=1}^n$ from $(d,\BFx)$ that follows a linear demand model $d= \BFx^\T \BFbeta^* + \varepsilon$, where the observation noise $\varepsilon$ is such that the conditional $\tau$-th quantile of $\varepsilon$ given $\BFx$ is 0 and $\tau=\frac{b}{b+h}$.
Moreover, we assume that the conditional density function of $\varepsilon|\BFx$, denoted by $f_{\varepsilon |\BFx}$, exists and satisfies some regularity conditions described below. Compared to the linear model (18) considered in \cite{ban2019}, here we do not assume the independence between the observation noise $\varepsilon$ and the random feature $\BFx$. Under the above model, the $\tau$-th conditional quantile of
$d$ given $\BFx$ is $\BFx^\T\BFbeta^*$. Specifically, let $F_d(\cdot | \BFx)$ be the conditional distribution function of $d$ given $\BFx$. Then, the conditional $\tau$-th conditional quantile of $d$ given $\BFx$ is formally defined as $Q_d(\tau|\BFx) = \inf\{ u : F_d(u | \BFx) \geq \tau \}$.
Write $\BFx=(x_1, \BFx_-^\T)^\T$, where $x_1 \equiv 1$ and $\BFx_- = (x_2,\ldots, x_p)^\T$ consists of the remaining random features. For theoretical analysis, we assume without loss of generality that $\mu_j  := \E(x_j) = 0$ for $j=2,\ldots, p$. Otherwise, it suffices to work with the demeaned model $Q_d(\tau|\BFx) =   \beta^\flat_1 +   \sum_{j=2}^p (x_j - \mu_j) \beta^*_j$, where $\beta^\flat_1 = \beta^*_1 +\sum_{j=2}^p   \mu_j \beta^*_j$.

To facilitate the analysis, we adopt the following technical conditions.
 
\begin{cond}[Kernel function] \label{cond.kernel} Let $K(\cdot)$ be a symmetric, non-negative, and Lipschitz continuous kernel function, that is, $K(u)= K(-u)$, $K(u) \geq 0$ for all $u$ and $\int_{-\infty}^\infty K(u)   {\rm d}u  = 1$. Moreover, assume $\kappa_u := \sup_{u\in \RR} K(u) $ and $\kappa_\ell := \int_{-\infty}^\infty |u|^\ell K(u) {\rm d}u$, $\ell=1,2 $ are bounded.
\end{cond}

  \begin{cond}[Feature distribution] \label{cond.subgaussian}
 The random predictor $\BFx_- \in \RR^{p-1}$ is {\it sub-Gaussian}: there exists $ \upsilon_1 \geq 1$ such that $\E e^{\lambda   \BFu^\T   \BFw_-  } \leq e^{\lambda^2 \upsilon_1^2 /2 }$ for all $\lambda \in \RR$ and $\BFu \in \mathbb{S}^{p-2}$, where $\BFw_-  = \BFS^{-1/2} \BFx_-$ and $\BFS = \E( \BFx_- \BFx_-^\T)$ is positive definite. 
 \end{cond}
 
Under Condition~\ref{cond.subgaussian},  the $p\times p$ matrix $\Sigma = \E(\BFx \BFx^\T)$ is also positive definite. Let $\BFw= \Sigma^{-1/2} \BFx=(1, \BFw_-^\T)^\T$ denote the standardized feature vector satisfying $\E(\BFw \BFw^\T) = \BFI_p$ and $\E(\BFw) = (1, \BFzero_{p-1}^\T)^\T$.

\begin{cond}[Observational noise  $\varepsilon=d - \BFx^\T \BFbeta^*$]  \label{cond.density}

There exist constants $l_0>0$ and $f_u\geq f_l>0$ such that $|f_{\varepsilon |\BFx}(u) - f_{\varepsilon |\BFx}(u) | \leq l_0 | u-v|$, $f_{\varepsilon |\BFx}(u) \leq f_u$  for all $u, v \in \RR$ almost surely (over $\BFx$), and 
\#
  \inf_{t\in[0,1], \,    \BFv \in \mathbb{S}^{p-1} }   \E \big\{  f_{\varepsilon |\BFx}(t \langle \BFw, \BFv \rangle ) \langle \BFw, \BFv  \rangle^2 \big\}  \geq f_l . \label{lower.density.bound}
\#
\end{cond}

Below, we discuss how Conditions~\ref{cond.subgaussian} and \ref{cond.density} are comparable to Assumptions 1 and 2 in \cite{ban2019}. For the random feature vector $\BFx=(1, \BFx_-^\T)^\T \in \RR^p$, \cite{ban2019} assumed that the coordinates of $\BFx_-$ are normalized with mean zero and standard deviation one, and $\| \BFx\|_2 \leq C\sqrt{p}$ for some constant $C>0$. Condition~\ref{cond.subgaussian}, on the other hand, requires feature vector $\BFx$ to be sub-Gaussian, extending the concept of sub-Gaussian random variables to higher dimensions through one-dimensional marginals. Examples of such sub-Gaussian vectors include: (i) Gaussian and Bernoulli random vectors, (ii) spherical random vectors,1 (iii) random vectors uniformly distributed on the Euclidean ball centered at the origin with radius $\sqrt{p}$, and (iv) random vectors uniformly distributed on the unit cube $[-1, 1]^p$. We refer to Chapter 3.4 in \cite{V2018} for detailed discussions on multivariate sub-Gaussian distributions. For sub-Gaussian features vectors $\BFx_i$'s, from Theorem~2.1 in \cite{HKZ2012} it follows that $\max_{1\leq i\leq n}\| \BFx_i \|_2 \lesssim \sqrt{p+\log n}$ with high probability. For the observation noise $\varepsilon$,  \cite{ban2019} assumed that its density function, denote by $f_\varepsilon$, is bounded away from zero on some compact interval $[\underbar{$D$}, \overline D]$. In Condition~\ref{cond.density}, it is worth noticing that the constant $f_l$ may depend on both the conditional distribution of $\varepsilon$ given $\BFx$, and the distribution of $\BFx\in \RR^p$. To see this, define $\iota_\delta = \inf\{  \iota >0 :  \E     \langle \BFw , \BFu \rangle^2 \mathbbm{1}_{ \{|\langle \BFw , \BFu \rangle| > \iota \}}   \leq \delta ~\mbox{ for all } \BFu \in \mathbb{S}^{p-1} \}$ for $\delta \in (0,1]$. It is easy to see that $\delta \mapsto \iota_\delta$ is non-decreasing and $\iota_\delta \leq (m_q/\delta)^{1/(q-2)}$ for any $q>2$, where $m_q := \sup_{\BFu \in \mathbb{S}^{p-1}} \E | \langle \BFw, \BFu \rangle|^q$. Then, a sufficient condition for \eqref{lower.density.bound} is $\min_{|t|\leq  \iota_\delta} f_{\varepsilon |\BFx}(t) \geq c_\delta  >0$ almost surely (over $\BFx$) for some $0<\delta<1$, because
\$
 & \inf_{t\in[0,1], \,    \BFv \in \mathbb{S}^{p-1} }   \E \big\{  f_{\varepsilon |\BFx}(t \langle \BFw, \BFv \rangle ) \langle \BFw, \BFv  \rangle^2 \big\}   \geq  c_\delta  \inf_{ \BFv \in \mathbb{S}^{p-1} }   \E \big\{ \langle \BFw, \BFv  \rangle^2 \mathbbm{1}  ( | \langle \BFw, \BFv  \rangle| \leq \iota_\delta    ) \big\}  \geq (1-\delta)c_\delta  .
\$

\section{A Privacy-Preserving Algorithm for Feature-based Newsvendor Problem}
\label{sec2}

\subsection{Preliminaries on $f$-differential Privacy}
\label{intro_privacy}

We will first review the $(\epsilon, \delta)$-differential privacy concept introduced in \citet{dwork2006calibrating, dwork2006our}. Let $\mathcal{S}$ denote a data set consisting of observations $\{\BFz_1,\ldots, \BFz_n\}$, where $\BFz_i=(d_i, \BFx_i)$, $i=1,\ldots,n$. A pair of data sets $\mathcal{S}$ and $\mathcal{S}'$ are said to be neighboring data sets if they differ in only one data point.

\begin{definition}
($(\epsilon, \delta)$-differential privacy or $(\epsilon, \delta)$-DP)
A randomized algorithm $M$ is $(\epsilon, \delta)$-differentially private if 
for any neighboring data sets $\mathcal{S}$ and $\mathcal{S}'$, and any event 
$\mathcal{E}$, we have
\begin{equation*}
    P(M(\mathcal{S})\in \mathcal{E}) \leq e^{\epsilon}P(M(\mathcal{S}')\in \mathcal{E})+\delta,
\end{equation*}
where $\epsilon\geq 0$ and $0\leq \delta\leq 1$ are constants.
\end{definition}

An intuitive way to understand the concept of $(\epsilon, \delta)$-differential privacy is via the lens of a hypothesis testing problem for distinguishing two neighboring  data sets $\mathcal{S}$ and $\mathcal{S}'$:
\begin{equation*}
    H_0: \mbox{the underlying data set is $\mathcal{S}$ } \quad 
    \mbox{versus} \quad
    H_1: \mbox{the underlying data set is $\mathcal{S}$}'. 
\end{equation*}
Consider any given test procedure $\phi$ based on the output of the randomized algorithm $M$, and denote its type I error and type II error by $\alpha_{\phi}$ and $\beta_{\phi}$, respectively. It can be shown that for a test $\phi$ based on the output of any $(\epsilon, \delta)$-differentially private algorithm, its power is bounded by $\min\{e^{\epsilon}\alpha_{\phi}+\delta, 1-e^{-\epsilon}(1-\alpha_{\phi}-\delta)\}$. If $\epsilon$ and $\delta$ are both small, then any $\alpha$-level ($0<\alpha<1$) test will be nearly powerless.

\citet{dong2022gaussian} extended \citet{dwork2006calibrating, dwork2006our}
by introducing the trade-off function to characterize the trade-off between the type I and type II errors.

\begin{definition}
(trade-off function \citep{dong2022gaussian}) For any two probability distributions $P$
and $Q$, the trade-off function $T(P,Q): [0, 1]\rightarrow [0,1]$
is defined as $T(P,Q)(\alpha)=\inf\{\beta_{\phi}: \alpha_{\phi}\leq \alpha\}$,
where the infimum is taken over all measurable rejection rules to distinguish $P$ and $Q$.
\end{definition}

The greater the trade-off function is, the harder it
is to distinguish the two distributions via hypothesis testing. 
\citet{dong2022gaussian} showed that a function 
$f: [0,1]\rightarrow [0,1]$ is a trade-off function if and only if $f$
is convex, continuous, nonincreasing and $f(x)\leq 1-x$ for all $x\in [0,1]$.
In the following, for any two functions $f$ and $g$ defined on [0,1], we write
$g\geq f$ if $g(x)\geq f(x)$, $\forall \ x\in [0,1]$. 
We abuse the notation a little by identifying 
$\mathcal{S}$ and $\mathcal{S}'$ with their 
respective probability distributions.

\begin{definition}
($f$-differential privacy or $f$-DP \citep{dong2022gaussian})
A randomized algorithm $M$ is said to be $f$-differentially private if 
for any neighboring data sets $\mathcal{S}$ and $\mathcal{S}'$
\begin{equation*}
    T(\mathcal{S},\mathcal{S}')\geq f
\end{equation*}
for some trade-off function $f$.
\end{definition}

If $f=T(P, Q)$ for some distributions $P$ and $Q$,
then a mechanism $M$ is $f$-DP if distinguishing any two neighboring datasets based on the output of $M$ 
is at least as difficult as distinguishing $P$ and $Q$
based on a single draw.
This functional perspective avoids some of the pitfalls associated with $(\epsilon, \delta)$-differential
privacy. $f$-DP is a generalization of  
$(\epsilon, \delta)$-DP. A result of 
\citet{wasserman2010statistical}
indicates that
a mechanism $M$ is $(\epsilon, \delta)$-DP
if and only if $M$ is $f_{\epsilon, \delta}$-DP
where 
$f_{\epsilon, \delta}(\alpha)
=\max\{0, 1-e^{\epsilon}\alpha_{\phi}-\delta, e^{-\epsilon}(1-\alpha_{\phi}-\delta)\}$.

\begin{definition}
(Gaussian differential privacy \citep{dong2022gaussian})
A randomized algorithm $M$ is said to satisfy 
$\mu$-Gaussian differential privacy if 
for any neighboring data sets $\mathcal{S}$ and $\mathcal{S}'$
\begin{equation*}
    T(\mathcal{S},\mathcal{S}')\geq G_{\mu}
\end{equation*}
where $G_{\mu}(\alpha)=\Phi(\Phi^{-1}(1-\alpha)-\mu)$
and $\Phi$ is the distribution function of the
standard normal distribution.
\end{definition}

\begin{figure}[ht]
\FIGURE
  {\includegraphics[width=0.5\linewidth]{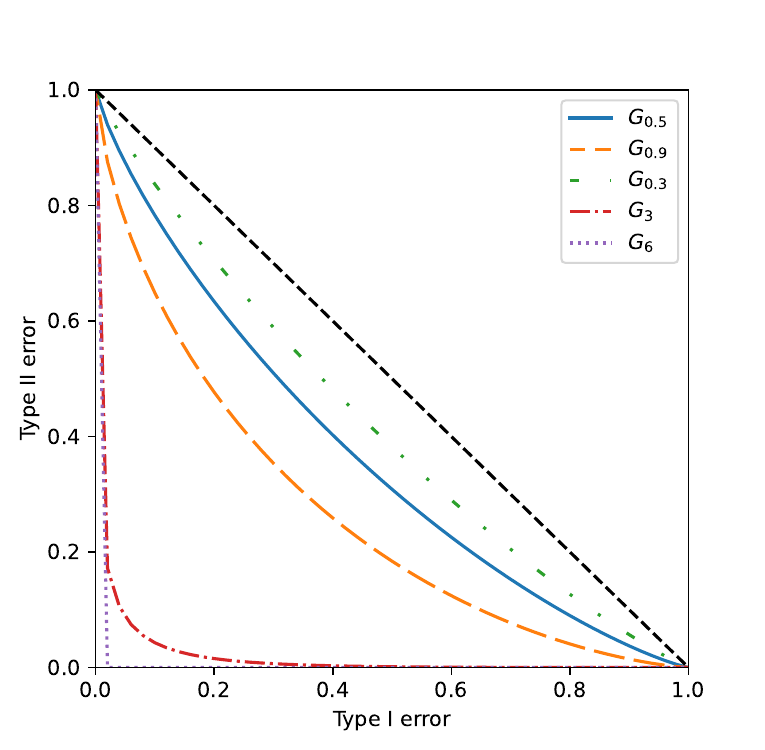}}
  {Trade-off functions for GDP with privacy parameter
$\mu$=0.3, 0.5, 0.9, 3 and 6, respectively.\label{Fig_GDP}}
{} 
\end{figure}

GDP provides a parametric family of $f$-DP
that guarantees and enjoys many desirable properties. As a rule of thumb, $\mu\leq 0.5$ guarantees a reasonable amount of privacy, 
$\mu=1$ is borderline private,
and $\mu>6$ promises almost no privacy guarantee. The trade-off of Type I and type II errors is illustrated in 
Figure~\ref{Fig_GDP}.

\subsection{Proposed Differentially Private Algorithm}

To obtain a differentially private counterpart of $\hat \BFbeta_{\varpi} = \argmin_{\BFbeta} \hat C_{\varpi}(\BFbeta)$, we draw inspiration from works \cite{BST2014} and \cite{SCS2013} and utilize a noisy optimization approach that involves adding Gaussian noise during each iteration of the gradient descent method.
Minimizing the loss function $\hat C(\BFbeta)$ used in \cite{ban2019} poses a challenge as it is convex but not differentiable everywhere, and subgradient methods commonly employed for such cases exhibit slow (sublinear) convergence, leading to computational instability. 
In this work, we propose a differentially private algorithm that leverages convolution smoothing and noisy gradient descent. Its privacy protection guarantee is established in Theorem~\ref{prop:GDP}
in Section~\ref{protection}.
We further investigate its finite sample performance in Section~\ref{sec:5} and demonstrate that the proposed approach achieves a balanced trade-off between statistical accuracy and computational stability, owing to the effectiveness of convolution smoothing.

Given a non-negative kernel function $K(\cdot)$ introduced earlier, we define
$$ 
	 \bar{K}(u) = \int_{-\infty}^u K(v){\rm d}v ~~\mbox{ and }~~ \bar{K}_{\varpi}(u) = \bar{K}(u/{\varpi}), 
$$
so that $\bar{K}'_{\varpi}(u) = K_{\varpi}(u)$. If the $p \times p$ matrix $\Sigma= \E(\BFx\BFx^\T)$ were known, we consider the following noisy smoothed gradient descent method, starting at iteration 0 with an initial estimate $\BFbeta^{(0)}$. At iteration $t= 0 , 1, 2, \ldots, T-1$, we update the solution as follows
\begin{equation}
 \BFbeta^{(t+1)}  = \BFbeta^{(t)} -  \frac{\eta_0}{n} \Sigma^{-1/2}\bigg[  \sn  \big\{ \bar{K}_{\varpi}(\BFx_i^\T \BFbeta^{(t)} - d_i ) - \tau \big\}    w_B(\BFw_i)    +    \sigma    \BFg_t \bigg]  , \label{noisy.gd}
\end{equation}
where $\BFw_i = \Sigma^{-1/2} \BFx_i$ denotes the standardized covariates vector in the sense that $\E(\BFw_i \BFw_i^\T) = \BFI_p$, $w_B(\BFu) = \BFu / \max\{ 1 , \| \BFu \|_2/ B \}$, and $\BFg_t \in \RR^p$ ($t=0, 1,\ldots, T-1$) are independent $\cN(\BFzero,   \BFI_p )$ vectors.  Here, $\eta_0>0$ is the step size, $T\geq 1$ is a prespecified number of iterations,  $B\geq 1$ is a truncation parameter,  and $\sigma$ is a positive constant
adjusting the level of noise injected into the gradient perturbation.
We summarize this procedure in Algorithm~\ref{algo:pqr}.

The presence of $\Sigma$ in Algorithm~\ref{algo:pqr} is primarily for theoretical convenience as it allows us to establish upper bounds on the estimation error $\E\{\BFx^\T(\hat \BFbeta_\varpi - \BFbeta^*)\}^2$ by expressing it as $\| \Sigma^{1/2} (\hat \BFbeta_\varpi - \BFbeta^*  ) \|_2^2$, where the expectation is only taken with respect to $\BFx$ that is independent of $\hat \BFbeta_\varpi$.
When $\Sigma$ is unknown, we consider the following update 
\begin{equation}
\BFbeta^{(t+1)}  = \BFbeta^{(t)} -   \frac{\eta_0}{n}   \bigg[ \sn  \big\{ \bar{K}_{\varpi}(\BFx_i^\T \BFbeta^{(t)} - d_i ) - \tau \big\}    w_B(\BFx_i)  +    \sigma  \BFg_t  \bigg]   \label{noisy.gd2}
\end{equation}
with a slight abuse of notation.

In both \eqref{noisy.gd} and \eqref{noisy.gd2},   the use of covariate clipping/truncation guarantees a bounded $\ell_2$-sensitivity of the gradient function, which is the key to achieving differential privacy. For the initial value, one can take $\BFbeta^{(0)}$ to be either $\BFzero$ or a random guess that is uniformly distributed over the unit sphere.

\begin{algorithm}[!t]
    \caption{ {\small Private ERM via Noisy Smoothed Gradient Descent}}
    \label{algo:pqr}
    \textbf{Input}: dataset $\{(d_i, \BFx_i)\}_{i=1}^n$,  probability level $\tau\in (0,1)$,  bandwidth $\varpi >0$,  initial value  ${\BFbeta}^{(0)}$,  step size $\eta_0>0$, noise scale $\sigma>0$,  truncation level $B\geq 1 $,  number of iterations $T \geq 1$.
    
    \begin{algorithmic}[1]
      \FOR{$t = 0,  1 \ldots, T-1$}
       \STATE  Generate standard multivariate normal vector $\BFg_t \sim \mathcal{N}( \BFzero , \BFI_p)$;
          \STATE Compute clipped/truncated covariates $\overline \BFw_i = \BFw_i \min \{ 1,  B / \| \BFw_i \|_2 \}$ for $i=1,\ldots, n$;
       \STATE  Compute $\BFbeta^{(t+1)} = \BFbeta^{(t)} - (\eta_0/n) \cdot  \Sigma^{-1/2} \big[  \sn \{   \bar K_{\varpi}( \BFx_i^\T \BFbeta^{(t)} - d_i ) - \tau \} \overline{\BFw}_i  + \sigma \BFg_t \big]$;
      \ENDFOR
    \end{algorithmic}
      \textbf{Output}: $\BFbeta^{(T)}$.
\end{algorithm}

\subsection{Privacy-protection Guarantee of the Proposed Algorithm}\label{protection}

We first review some useful results about $f$-DP.

\begin{proposition} \label{GaussMechianism}
(Theorem 1 in \citet{dong2022gaussian})
Define the Gaussian mechanism that operates on a statistic $\theta$
as $M(\mathcal{S})=\theta(\mathcal{S})+\xi$,
where $\xi\sim \cN(0, \mbox{sens}(\theta)^2/\mu^2)$,
and the sensitivity $\mbox{sens}(\theta)=\sup_{\mathcal{S},\mathcal{S}'}
|\theta(\mathcal{S})-\theta(\mathcal{S}')|$ with the supremum over all neighborhood data sets $\mathcal{S}$ and $\mathcal{S}'$.
Then $M$ is $\mu$-GDP.
\end{proposition} 

A nice property of $f$-DP is that the composition of private mechanisms is closed and tight in the $f$-DP framework. 
A two-step composition can be written as
$M(\mathcal{S})=(y_1, M_2(\mathcal{S}, y_1))$,
where $y_1=M_1(\mathcal{S})$ with
$M_1: X\rightarrow Y_1$ being the first mechanism,
and $M_2: X\times Y_1\rightarrow Y_2$
is the second mechanism. 
In general, given a sequence of
mechanisms $M_i: X\times Y_1\times \cdots \times Y_{i-1}\rightarrow Y_i$, $i=1,\dots,n$, we consider the $n$-fold composed mechanism:
$M:X\rightarrow Y_1\times \cdots \times Y_n$.

\begin{definition}
The tensor product of two trade-off functions $f=T(P,Q)$
and $g=T(P',Q')$ is $f\bigotimes g=T(P\times P', Q\times Q')$.
\end{definition}

\begin{proposition} 
(Theorem 4 in \citet{dong2022gaussian})
Let $M_i(\cdot, y_1,\ldots, y_{i-1})$ be $f_i$-DP for all
$y_1\in Y_1, \ldots, y_{i-1}\in Y_{i-1}$. Then the $n$-fold composed mechanism 
$M: X\rightarrow Y_1\times \cdots Y_n$ is $f_1\bigotimes  \cdots 
\bigotimes f_n$-DP.
\end{proposition} 

\begin{corollary} (Corollary 2 in \citet{dong2022gaussian})
The $n$-fold composition of $\mu_i$-GDP mechanisms is
$\sqrt{\mu_1^2+\ldots+\mu_n^2}$-GDP.
\end{corollary}

\begin{remark}
\citet{dong2022gaussian} revealed that the Gaussian mechanism in Proposition \ref{GaussMechianism} satisfies $G_\mu(\alpha)=\inf _{\text {neighboring } S, S^{\prime}} T\left(M(S), M\left(S^{\prime}\right)\right)(\alpha)$. For all possible type I error rate $\alpha$, the infimum could be achieved at two neighboring data sets which satisfy $|\theta(S)-\theta(S^{\prime})| = \text{sens}(\theta)$. This implies that the characterization by GDP is precise in the pointwise sense. GDP offers the tightest possible privacy bound of the Gaussian mechanism. 
\end{remark}

Theorem~\ref{prop:GDP} below establishes the Gaussian differential privacy property of the proposed algorithm. For any given
sample size $n$, number of iterations $T$ and noise scale
$\sigma$, the output of the algorithm $\BFbeta^{(T)}$
satisfies the Gaussian differential privacy property with privacy level $\mu$ as outlined in the theorem.

\begin{theorem} \label{prop:GDP} (Privacy-protection guarantees)
Given an initial estimate $\BFbeta^{(0)} \in \RR^p$ and a dataset $\bZ_n = \{ (d_i, \BFx_i ) \}_{i=1}^n$,  consider the noisy gradient descent iterates $\{ \BFbeta^{(t)} \}_{t=0, \ldots, T}$ defined in \eqref{noisy.gd}. Given a privacy level $\mu>0$,
if $\sigma>0$
satisfies $\sigma \geq  2\bar \tau   B T^{1/2}/\mu$ with $\bar \tau = \max(\tau, 1-\tau)$, then   
the final output $\BFbeta^{(T)}$ is $\mu$-GDP.
\end{theorem}

\proof{Proof of Theorem~\ref{prop:GDP}}
Consider two datasets $\bZ_n$ and $\bZ'_n$ that differ by one datum,  say $(d_1, \BFx_1) \in \bZ_n$ versus $(d_1', \BFx_1')\in \bZ_n'$.  At the first iteration, note that
\$
	 & \| \Sigma^{1/2} \BFbeta^{(1)} (\bZ_n) -\Sigma^{1/2} \BFbeta^{(1)} (\bZ'_n) \|_2  \\
	 &  =  \frac{\eta_0}{n}  \bigg\| \big\{ \bar{K}_{\varpi}(\langle \BFx_1  ,  \BFbeta^{(t)}  \rangle - d_1) - \tau \big\}    w_B(\BFw_1)  - \big\{ \bar{K}_{\varpi}( \langle \BFx_1' ,  \BFbeta^{(t)}  \rangle - d'_1 ) - \tau \big\}    w_B(\BFw'_1)   \bigg\|_2 \\
	 & \leq  2  \max( 1-\tau, \tau)  B  \frac{\eta_0}{n} = 2\bar \tau B  \frac{\eta_0  }{n} .
\$
By Proposition~1,  $\Sigma^{1/2} \BFbeta^{(1)}$ is $(T^{-1/2} \mu)$-GDP as long as $\sigma \geq 2\bar \tau   B T^{1/2}  / \mu$. After post-processing (by a deterministic map), $\BFbeta^{(1)}$ is also $(T^{-1/2} \mu)$-GDP.

By definition,  the second iterate $\BFbeta^{(2)} = \BFbeta^{(2)} (\bZ_n)$ takes $\BFbeta^{(1)}$ as  input  in addition to the dataset.  Together, Proposition~2 and Corollary~1 show that the two-fold composed (joint) mechanism $(\BFbeta^{(1)} , \BFbeta^{(2)} )$ is $\sqrt{\mu^2/T + \mu^2/T}$-GDP.
Using the same argument repeatedly, we conclude that the $T$-fold composed mechanism $(\BFbeta^{(1)} ,  \ldots, \BFbeta^{(T)} )$ is $\mu$-GDP, and  hence  so is  $\BFbeta^{(T)}$. \Halmos
 \endproof

\section{Theoretical Performance}
\label{sec:5}

Theorem~\ref{prop:GDP} in Section~\ref{protection} establishes the privacy-protection guarantees 
of the proposed new algorithm.
This section provides a statistical analysis of the privacy-preserving coefficient estimate $\BFbeta^{(T)}$ under Conditions~\ref{cond.kernel}--\ref{cond.density} from Section~\ref{sec:3.3}. 
Section~\ref{sec:5.1} provides upper bounds for the finite-sample bias of the estimated optimal policy both in high probability and under expectation. Section~\ref{sec:5.2} provides the regret analysis. To prove these bounds, the main technical challenge is that the empirical cost function after convolution smoothing does not satisfy the local strong convexity as required in \cite{ABL2021}.
We provide main proof strategies and important intermediate results in Section~\ref{sec:5.3}.

\subsection{Performance Bound on Estimation Error}
\label{sec:5.1}

The parameter indexing the clairvoyant optimal policy, where the demand distribution is known a priori is given by
(\ref{def:true.beta}): 
$
    \BFbeta^*=\argmin_{\BFbeta:  q(\BFx)=  \BFx^\T   \BFbeta} \E \{ C(q(\BFx), d))|\BFx \} .
$ 
Given a feature vector $\BFx$, the clairvoyant optimal
to-order-quantity is $\BFx^\T \BFbeta^*$. Given an estimate $\hat \BFbeta$ of the unknown parameter $\BFbeta^*$ based on a random, independent
sample $\{ (d_1, \xb_1), \ldots, (d_n, \xb_n) \}$ of size $n$ drawn from the linear demand model, we use the following two metrics to evaluate its performance: (i) the estimation error $\| \hat \BFbeta - \BFbeta^* \|_2$ under the vector $\ell_2$-norm or its variant, and (ii) the excess (population) risk $C(\hat \BFbeta ) - C(\BFbeta^*)$, where $C(\cdot )$ is defined in \eqref{def:true.beta}.

With a predetermined bandwidth $\varpi>0$, step size $\eta_0>0$, truncation level $B\geq 1$, number of iterations $T\geq 1$, and noise scale $\sigma=2\bar \tau B T^{1/2}/\mu$, let $\BFbeta^{(T)}$ be the $\mu$-GDP estimator of $\BFbeta^*$ obtained from Algorithm~\ref{algo:pqr}, where $\bar \tau = \max(\tau, 1-\tau)$.
Our first theorem provides upper bounds for the estimation error $\|  \BFbeta^{(T)}  - \BFbeta^* \|_\Sigma$ both {\it in high probability} and {\it under expectation}, where $\| \cdot \|_\Sigma$ denotes the $\Sigma$-induced norm, that is, $\| \BFu \|_\Sigma = \sqrt{\BFu^\T \Sigma \BFu}$ for $\BFu \in \mathbb{R}^p$.

\begin{theorem} \label{thm:high.prob.bound}
In addition to Conditions~\ref{cond.kernel}--\ref{cond.density}, assume $\kappa_l := \min_{|u|\leq 1} K(u) >0$. Let the triplet of parameters $(\varpi, B, T)$ and sample size $n$ satisfy 
\#
    \varpi  \asymp  \bigg(  \frac{p+ \log n}{n} \bigg)^{1/4}, \quad B \asymp \sqrt{p+\log n}, \quad T \asymp \log n  ~~\mbox{ and }~~ n \gtrsim  T^{1/2}    \frac{p+ \log n}{  \mu f_l  } . \label{sample.size.cond2}
\#
Moreover, let the step size $\eta_0$ satisfy $0<\eta_0 \leq 1/\max(  2 f_u ,  f_l + \bar \tau  )$. Then, the $\mu$-GDP estimated coefficient $\BFbeta^{(T)}$ obtained from noisy gradient descent initialized at any $\BFbeta^{(0)} \in \BFbeta^* + \Theta_\Sigma(1)$ satisfies
\$
	 \| \BFbeta^{(T)} - \BFbeta^* \, \|_\Sigma \leq C_0 \bigg( \eta_0 T^{1/2} \frac{ p + \log n }{   \mu n} +  \frac{1}{f_l} \sqrt{\frac{p \log n}{n}} \, \bigg)  \mbox{ with probability at least }  1- \frac{C_1}{n^2}   
\$
and
\$
 \E \| \BFbeta^{(T)} - \BFbeta^* \, \|_\Sigma  \leq C_2 \sqrt{\log n}\bigg( \eta_0\frac{ p + \log n }{   \mu n} + \frac{1}{f_l}\sqrt{\frac{p }{n}} \, \bigg) ,
\$
where $C_0$, $C_1$ and $C_2$ are positive constants independent of $(n, p)$.
\end{theorem}

As a benchmark, we use $\hat \BFbeta_\varpi$ to denote the non-private empirical (smoothed) risk minimizer, that is, $\hat \BFbeta_\varpi = \argmin_{\BFbeta \in \RR^p} \hat C_\varpi(\BFbeta)$ with $\hat C_\varpi(\cdot)$ defined in \eqref{smooth.qloss}.
From the proof of Theorem~\ref{thm:high.prob.bound} we see that the second term on the right-hand side upper bound is the bound on the estimation error of $\hat \BFbeta_\varpi$, which is of order $\sqrt{p/n}$ with a properly chosen smoothing parameter $\varpi$ \citep{HPTZ2020}. The first term quantifies the ``cost of privacy" of the noisy gradient descent algorithm for solving the feature-based newsvendor problem. For sufficiently small values of $\mu$, the obtained upper bound (on $\| \cdot \|_\Sigma$-error) matches the minimax lower bound, up to logarithmic factors, for $(\mu, \delta)$-DP estimation of $\BFbeta^*$ under a linear model with normal errors; see Theorem~4.1 in \cite{CWZ2021}. By Corollary 1 in \cite{dong2022gaussian}, an algorithm is $\mu$-GDP if and only if $(\mu, \delta(\mu))$-DP, where $\delta(\mu) = \Phi(-1+\mu/2) - e^{\mu} \Phi(-1-\mu/2)$.

\subsection{Regret Analysis}
\label{sec:5.2}

We next provide a finite-sample analysis of the regret $C(   \BFbeta^{(T)}) - C( \BFbeta^*)$, where $C(\BFbeta) = \E \{ \hat C(\BFbeta) \}$ is as in \eqref{def:true.beta}. 
The regret is the difference between the expected cost obtained with the estimated privacy-preserving optimal policy and the clairvoyant optimal expected cost with known demand distribution but without privacy protection.

Without loss of generality (up to a constant scale),
we consider the regret of $\BFbeta^{(T)}$, defined as $Q(\BFbeta^{(T)}) - Q(\BFbeta^*) = Q(\BFbeta^{(T)}) - \inf_{\BFbeta \in \mathbb{R}} Q(\BFbeta)$, where $Q(\BFbeta) = (b+h)^{-1}  C(\BFbeta) =\E \{ \rho_\tau(d- \BFx^\T \BFbeta) \}$ with $\tau=b/(b+h)$ is the population cost (without smoothing). Recall that the coefficient $\BFbeta^*$ indexing the optimal inventory policy satisfies the first-order condition $\nabla Q(\BFbeta^*) = \BFzero$. By the mean value theorem and Condition~\ref{cond.density} that $\sup_{u\in \mathbb{R} } f_{\varepsilon | \BFx}(u) \leq f_u$, it can be shown that $Q(\BFbeta) - Q(\BFbeta^*) = Q(\BFbeta) - Q(\BFbeta^*) - \langle \nabla Q(\BFbeta^*), \BFbeta - \BFbeta^* \rangle \leq 0.5 f_u \| \BFbeta - \BFbeta^* \|_\Sigma^2$ for any $\BFbeta \in \mathbb{R}^p$. This, combined with Theorem~\ref{thm:high.prob.bound}, implies the following high probability upper bound on the regret along with an expected regret bound.

\begin{theorem} \label{thm:regret.bound}
(High probability bound for excess risk)
Under the same set of assumptions in Theorem~\ref{thm:high.prob.bound},  we have
\begin{align}
      Q(\BFbeta^{(T)}  ) - Q(\BFbeta^*)  \lesssim  \log(n) \bigg\{   \frac{1}{f_u}\bigg( \frac{ p + \log n}{\mu n} \bigg)^2  + \frac{f_u}{f_l^2} \frac{p}{  n}  \bigg\}  \label{regret.ubd}
\end{align}
with probability at least $1-C_1 n^{-2}$.
\end{theorem}

\begin{corollary} (Excess population risk bound)
Under the conditions of Theorem~\ref{thm:regret.bound},
the excess population risk bound satisfies
\begin{align}
\E \{ Q(\BFbeta^{(T)}  ) - Q(\BFbeta^*) \} \lesssim  \log(n) \{ f_u^{-1} (p+\log n)^2/(\mu n)^2 + (f_u/f_l^2)  \cdot   p /n \}.
\end{align}
\end{corollary}

For fixed cost parameters $b$ and $h$, the bound \eqref{regret.ubd} indicates that if the number of relevant features $p$ is small relative to the number of observations in the sense that $p\log n = o(n)$, the expected cost of the $\mu$-GDP estimated decision converges to that of the optimal decision at a fast rate $O( p/n +  \mu^{-2} (p/n)^2 )$, up to logarithmic factors. Without privacy guarantees, \cite{ban2019} obtained a similar performance bound, which implies consistency but under a stronger condition on the number of features, that is, $p^2 = o(n)$.

\subsection{Proof Strategy and Key Intermediate Results}
\label{sec:5.3}

The statistical analysis of the noisy gradient descent iterates $\{ \BFbeta^{(t)} \}_{t=1,\ldots, T}$ depends crucially on the landscape of the empirical loss function, as highlighted in recent research \citep{CWZ2021, ABL2021}. Unlike many commonly used loss functions in statistical learning,  such as squared loss, Huber loss (and its smoothed variants), and logistic loss, the quantile loss $\rho_\tau$ exhibits piecewise linearity, which lacks local strong convexity.  Instead, its ``curvature energy'' is concentrated in a single point. Notably, the local strong convexity and smoothness of $\hat Q_\varpi(\cdot)$ are intricately influenced by the bandwidth used in the analysis.

In this study, we present the key findings for analyzing the landscape of $\hat Q_\varpi(\cdot)$.  Specifically, we illustrate that by conditioning on a series of ``good events'' related to the empirical smoothed cost function,   the proposed noisy gradient descent iterates exhibit favorable convergence properties. Moreover, we demonstrate that these good events will occur with high probability under Conditions~\ref{cond.kernel}--\ref{cond.density}.   We conclude this section with a supplementary analysis of the initialization of the algorithm.

Given a kernel function $K(\cdot)$ and bandwidth $\varpi >0$, we define the empirical smoothed loss
$$ 
\hat Q_\varpi(\BFbeta) =  (b+h)^{-1} 
\hat C_\varpi(\BFbeta)   = \frac{1}{n} \sn  \underbrace{ (  \rho_\tau * K_\varpi  ) }_{=: \, \ell_\varpi } ( d_i - \BFx_i^\T \BFbeta),
$$
where ``$*$" is the convolution operator. Its gradient and Hessian are given, respectively, by
\$
 \nabla \hQ_\varpi(\BFbeta) =  \frac{1}{n} \sn \big\{ \bar{K}_\varpi( \BFx_i^\T \BFbeta - d_i ) - \tau \big\} \BFx_i ~~\mbox{ and }~~ \nabla^2 \hQ_\varpi(\BFbeta) = \frac{1}{n} \sn K_\varpi( d_i - \BFx_i^\T \BFbeta ) \BFx_i \BFx_i^\T .
\$
Let $Q(\BFbeta) = \E\{ \hat Q(\BFbeta) \}$ and $Q_\varpi(\BFbeta) = \E \{ \hQ_\varpi(\BFbeta)\}$ be the population quantile and smoothed quantile losses, respectively. Note that although the parameter $\BFbeta^*$ indexing the theoretically optimal inventory policy satisfies the moment condition $\nabla   Q(\BFbeta^*) = \BFzero$, in general, $\nabla Q_\varpi(\BFbeta^*) \neq \BFzero$. Therefore, we use
$$
 b^* := \| \Sigma^{-1/2} \nabla Q_\varpi(\BFbeta^*) \|_2 = \|  \E    \{ \bar{K}_\varpi( \BFx^\T \BFbeta^* - d) - \tau   \}     \BFw    \|_2
$$ 
to quantify the smoothing bias. Together, Condition~\ref{cond.kernel} and the Lipschitz continuity of $f_{\varepsilon|\BFx}(\cdot)$ ensure that $b^* \leq 0.5 l_0 \kappa_2  \varpi^2$; see Lemma~1.3 in the supplementary material.

For any $r>0$, define the local ellipses centered at the origin and $\BFbeta^*$, respectively, as
\$
	 \Theta_{\Sigma}(r) =  \big\{ \BFdelta \in \RR^p: \| \BFdelta \|_{\Sigma} \leq r \big\} ~~\mbox{ and }~~ \Theta_\Sigma^*(r)    = \big\{ \BFbeta \in \RR^p: \| \BFbeta - \BFbeta^* \|_{\Sigma} \leq r \big\}  .
\$
Moreover, for every $\BFbeta\in \RR^p$,  we write 
\#
	 \hat D_\varpi (\BFdelta) &   = \hQ_\varpi(\BFbeta) - \hQ_\varpi(\BFbeta^* )  ~~\mbox{ and }~~ D_\varpi(\BFdelta ) = \E\{ \hat D_\varpi(\BFdelta) \} ~~\mbox{ for }~ \BFdelta = \BFbeta - \BFbeta^*.     \label{def:Dh}  
\#
Given parameters $B,  R \geq 1$ and $\delta_0 , \delta_1 >0$,  define the ``good events''
\#
  \cE_0(B ) & = \Big\{ \max_{1\leq i\leq n}  \| \BFw_i \|_2 \leq B  \Big\}   , \label{def:E0} \\
 \cE_1(\delta_0,  \delta_1 )  
 &= \big\{    | \hat D_\varpi (\BFdelta ) - D_\varpi(\BFdelta) | \leq \delta_0  \| \BFdelta \|_{\Sigma}  \mbox{ for all } \BFdelta  \in \Theta_\Sigma(1)\setminus {\Theta}_\Sigma(1/n)    \big\}  \label{def:E1}   \\ 
   &~~~~~~ \cap  \big\{    \| \nabla \hQ_\varpi(\BFbeta) - \nabla Q_\varpi(\BFbeta) \|_{\Sigma^{-1}} \leq \delta_1   \mbox{ for all }  \BFbeta \in \Theta_\Sigma^*(1) \big\}  , \nn \\
 \cE_2(R) & =    \big\{    \| \nabla^2 \hQ_\varpi(\BFbeta) - \nabla^2 Q_\varpi(\BFbeta) \|_{\Sigma^{-1}} \leq  f_u  \mbox{ for all }  \BFbeta \in \Theta_\Sigma^*(R) \big\}  . \label{def:E2} 
\#  
Here we write $\| \BFA \|_{\Sigma^{-1}} = \| \Sigma^{-1/2} \BFA \Sigma^{-1/2} \|_2$ for any $p\times p$ matrix $\BFA$. In the following, we will restrict our analysis to the intersection of the above events.

\begin{proposition}   \label{prop:smooth.landscape} (Restricted strong convexity and smoothness)
Let $0< \varpi \leq f_l /(2 l_0\kappa_1 )$ and set $\phi_1= 0.5(f_l - l_0 \kappa_1 \varpi) \geq 0.25f_l > 0$. Then, conditioned on the event $   \cE_1(\delta_0,\delta_1) \cap \cE_2(R)$, we have
\#
  &  \hat Q_\varpi(\BFbeta) - \hQ_\varpi(\BFbeta^* )   \nn\\
  &\geq \begin{cases}
\phi_1 \| \BFbeta - \BFbeta^* \|_{\Sigma}^2 -  (\delta_0 + b^*  ) \| \BFbeta - \BFbeta^* \|_{\Sigma}    & \mbox{ for all }~ \BFbeta \in \Theta_\Sigma^*(1)\setminus  {\Theta}_\Sigma^*(1/n) \\  
 ( \phi_1 - \delta_0  - b^*   ) \| \BFbeta - \BFbeta^* \|_{\Sigma}  & \mbox{ for all }~ \BFbeta \in \Theta_\Sigma^*(1)^\cc 
\end{cases}  , \label{restricted.loss.difference} 
\#
\#
 & \hQ_\varpi(\BFbeta^*) - \hQ_\varpi(\BFbeta)  -  \langle \nabla \hQ_\varpi(\BFbeta), \BFbeta^* - \BFbeta \rangle    \nn \\
  &  ~~~~~~~~~~~~~~~~~~~   \geq  
     \phi_1 \| \BFbeta - \BFbeta^* \|_{\Sigma}^2 -  (\delta_0 +  \delta_1 ) \| \BFbeta - \BFbeta^* \|_{\Sigma} ~\mbox{ for all }~ \BFbeta \in \Theta_\Sigma^*(1)\setminus  {\Theta}_\Sigma^*(1/n) ,  \label{restricted.rsc.one-side} 
\#
and
\#
&       \hat Q_\varpi(\BFbeta_2) - \hQ_\varpi(\BFbeta_1 ) - \langle  \nabla \hQ_\varpi(\BFbeta_1) , \BFbeta_2 - \BFbeta_1 \rangle   \leq   f_u    \| \BFbeta_2 - \BFbeta_1 \|^2_{\Sigma} ~\mbox{ for all }~ \BFbeta_1, \BFbeta_2 \in \Theta_\Sigma^*(R) .  \label{restricted.smootheness}
\# 
\end{proposition}

Note that the lower bound \eqref{restricted.rsc.one-side} implies a restricted strong convexity (RSC) for $\hQ_\varpi(\BFbeta)$ when $\BFbeta \in \Theta_\Sigma^*(1) / \Theta_\Sigma^*( n^{-1} \vee   r_1  )$ with $r_1  = (\delta_0+\delta_1)/\phi_1$. The upper bound \eqref{restricted.smootheness} is related to the local strong smoothness of the empirical cost, which no longer holds without convolution smoothing. In addition, we define a ``good'' event on which the smoothed empirical loss $\hat Q_\varpi(\cdot)$ satisfies a refined RSC property. Given a radius $r > 0$ and a curvature parameter $\phi_2\in (0, f_l)$, define 
$$
\mathcal{E}_3(r, \phi_2) = \Bigg\{ \inf_{\BFbeta_1 \in \BFbeta^* + \Theta_\Sigma(r/2), \, \BFbeta_2 \in \BFbeta_1 + \Theta_\Sigma(r) }\frac{\hat Q_\varpi(\BFbeta_1) - \hat Q_\varpi(\BFbeta_2) - \langle \nabla \hat Q_\varpi(\BFbeta_2), \BFbeta_1 - \BFbeta_2 \rangle}{ \|\BFbeta_1 - \BFbeta_2  \|_\Sigma^2} \geq \phi_2 \Bigg\}.
$$
Now we are ready to present the following general upper bound on the estimation error conditioning on the above good events.

\medskip
\begin{theorem} \label{thm:convergence}
Assume Conditions~\ref{cond.kernel} and \ref{cond.density} hold, $\BFbeta^{(0)} \in \Theta_\Sigma^*(1)$  and let $(\varpi, \eta_0)$ satisfy $0< \varpi \leq f_l /( 2l_0 \kappa_1)$ and $0<\eta_0 \leq 1/ \max(2f_u, f_l+\bar \tau)$, where $\bar \tau = \max(\tau, 1-\tau)$. Set $R=2$, and let $\delta_0, \delta_1>0$ be such that
\#
 r_0 :=   (\delta_0+b^* )/\phi_1  < 1 ~~\mbox{ and }~~  r_1 :=  (\delta_0 +\delta_1)/\phi_1  \leq 1     , \label{parameters.cond1}
\#
where $\phi_1 = 0.5 (f_l - l_0 \kappa_1 \varpi ) $.
Moreover, let $\Delta = \phi_1 - \delta_0 - b^* \in (0, f_l/2)$ and $\epsilon = \eta_0 \phi_1   \in (0, 1/2)$. For any $z\geq 0$,  let the sample size satisfy
\#
   n \geq  \sigma  \frac{  p^{1/2} + \sqrt{2 (\log T + z) }    }{ \Delta  }   . \label{sample.size.requirement}
\#
Then, conditioned on  the event $\cE_0(B)\cap \cE_1(\delta_0,\delta_1 ) \cap \cE_2( 2)$, the noisy gradient descent iterate $\BFbeta^{(T)}$ with $T \geq 2 \log( n)/ \log( (1- \epsilon )^{-1} )$ satisfies
\#
	 \| \BFbeta^{(T) } - \BFbeta^* \|_\Sigma \leq r^* : =  \sqrt{  \frac{1}{n^2}  +(1+1/\epsilon)   \bigg\{       (2p/\epsilon + 3z)  \Big( \frac{\eta_0 \sigma}{n} \Big)^2  +  (r_0 \vee r_1)^2 \bigg\} }     \label{def:r*}
\#
with probability (over the i.i.d. normal vectors $\{\BFg_t\}_{t=0}^{T-1}$) at least $1-2e^{-z}$. Moreover, the non-private empirical (smoothed) risk minimizer satisfies $\| \hat \BFbeta_\varpi - \BFbeta^* \|_\Sigma\leq r_0$.

Let $r=r^* + r_0$. Conditioned further on $\mathcal{E}_3(r, \phi_2)$, $\BFbeta^{(T)}$ with $T\geq 3\log(n) / \log((1-\epsilon)^{-1})$ satisfies
\#
 \| \BFbeta^{(T)} - \hat \BFbeta_\varpi \|_\Sigma \leq r^\dagger := \sqrt{\frac{r^2}{n} + (1+1/\epsilon) (2p/\epsilon + 3z) \bigg(\frac{\eta_0 \sigma }{n} \bigg)^2 } 
 \label{def:rdagger}
\#
with probability (over the i.i.d. normal vectors $\{\BFg_t\}_{t=0}^{T-1}$) at least $1-2e^{-z}$.
\end{theorem}

\medskip
\begin{remark}
Under local strong convexity and smoothness conditions, \cite{ABL2021} established statistical convergence guarantees for private $M$-estimators obtained via noisy gradient descent. Let $\mathcal{L}_n: \mathbb{R}^p \to \mathbb{R}$ be a general (empirical) loss function of interest, and $\Theta \subseteq \mathbb{R}^p$ be the parameter space. As high-level conditions, \cite{ABL2021} assumed that $\mathcal{L}_n$ is locally $\tau_1$-strongly convex and $\tau_2$-smooth, that is,
$$
    \mathcal{L}_n(\BFbeta_1) -  \mathcal{L}_n(\BFbeta_2) \geq \langle \nabla  \mathcal{L}_n(\BFbeta_2) , \BFbeta_1-\BFbeta_2 \rangle + \tau_1 \| \BFbeta_1 - \BFbeta_2 \|_2^2 , \quad \forall \BFbeta_1, \BFbeta_2 \in  \{ \BFbeta \in \mathbb{R}^p: \| \BFbeta - \BFbeta^* \|_2 \leq r \}  
$$
for some $r>0$, and $
    \mathcal{L}_n(\BFbeta_1) -  \mathcal{L}_n(\BFbeta_2) \leq  \langle \nabla  \mathcal{L}_n(\BFbeta_2) , \BFbeta_1-\BFbeta_2 \rangle + \tau_2 \| \BFbeta_1 - \BFbeta_2 \|_2^2$, $\forall \BFbeta_1, \BFbeta_2 \in \Theta$. To our knowledge, it remains uncertain whether the aforementioned local strong assumption holds with high probability for either the empirical newsvendor loss $\hat Q$ or its convolution-smoothed counterpart $\hat Q_{\varpi}$. Therefore, a more delicate argument is required to analyze the convergence of noisy gradient descent iterates obtained from Algorithm~\ref{algo:pqr}. Our proof of Theorem~\ref{thm:convergence} crucially relies on the structural properties of $\hat Q_{\varpi}$ stated in Proposition~\ref{prop:smooth.landscape}. In Proposition~\ref{prop:event012} below, we will show that the event conditioned on in Proposition~\ref{prop:smooth.landscape} holds with high probability. 
\end{remark}

\medskip
\begin{remark}
The proof of the error bound $\| \hat \BFbeta_\varpi - \BFbeta^* \|_\Sigma \leq r_0$ in Theorem~\ref{thm:convergence}, which holds conditioning on event $\cE_1(\delta_0, \delta_1)$, extends the argument in \cite{HPTZ2020}. The main contribution of Theorem~\ref{thm:convergence} is to establish finite sample performance bounds for the noisy gradient descent iterates $\{ \BFbeta^{(t)}\}_{t=1}^T$ in a sequential manner, which involves a more intricate analysis compared to that for the non-private empirical (smoothed) risk minimizer $\hat \BFbeta_\varpi$. More specifically, the analysis conducted in \cite{HPTZ2020} necessitates that the empirical loss satisfies only condition \eqref{restricted.loss.difference},  while our approach requires a more comprehensive version of the restricted strong convexity property \eqref{restricted.rsc.one-side}. 
We also establish a connection between statistical theory and algorithmic complexity, demonstrating that to achieve a statistically efficient estimator as shown in Theorem~\ref{thm:high.prob.bound}, the computational complexity is of order $O(n p \log(n))$. In contrast, the conventional interior-point method commonly used for solving the LP reformulation of empirical check loss minimization demands a significantly higher average-case computational complexity of $O_P(n^{1.25} p^3 \log n)$ \citep{PK1997}.
\end{remark}

\medskip
The convergence result stated in Theorem~\ref{thm:convergence} relies on the assumption that the initial value $\BFbeta^{(0)}$ falls within the neighborhood $\Theta^*_\Sigma(1)$, which we term as the {\it tightening region}. In each iteration of the noisy gradient descent, the current estimate contracts towards the true parameter, progressively moving closer to the region of near-optimal convergence.

In general, let us define $R_0:= \| \BFbeta^{(0)} - \hat \BFbeta_\varpi \,  \|_\Sigma$ to be the distance between the initial value $\BFbeta^{(0)}$ and the non-private empirical risk minimizer $\hat \BFbeta_\varpi$.  The following result presents the number of iterations necessary for the noisy gradient descent to enter the tightening region.

\begin{theorem} \label{thm:initial.convergence}
Assume Conditions~\ref{cond.kernel} and \ref{cond.density} hold, and let $(\varpi, \eta_0)$ satisfy $0< \varpi \leq f_l /( 2l_0 \kappa_1)$ and $0< \eta_0 \leq \min\{ 1, 1/(2f_u) \}$. Without loss of generality, assume $R_0 = \| \BFbeta^{(0)} - \hat \BFbeta_\varpi \,  \|_\Sigma > 1$, and let
$$
	\Delta = \phi_1 - \delta_0 - b^* \in (0, f_l/2), \quad r_0 = (\delta_0 + b^*)/\phi_1 \in (0, 1) , 
$$ 
where $\phi_1 = (f_l - l_0 \kappa_1 \varpi ) /2$. Given $z\geq 0$, let the number of iterations $T_0$ and sample size $n$ satisfy
\#
 T_0 \geq \frac{R_0^2}{\eta_0 \Delta}~~\mbox{ and }~~
 n \geq 2 B_{T_0} \sigma \max\bigg\{  \frac{  2 R_0 + (\bar \tau B + 1/4) (T_0+1) \eta_0 }{\Delta} ,  \frac{e-1}{4-e}( \bar \tau B + 1/4 ) T_0  \bigg\} , \label{T0.n.requirement}
\#
where $B_{T_0} = \sqrt{p} + \sqrt{2(\log T_0 + z) }$.
Then, conditioned on $\cE_0(B) \cap \cE_1(\delta_0, \delta_1) \cap \cE_2(R)$ with $R= 2 R_0 + r_0$,  the noisy gradient descent iterate $\BFbeta^{(T_0)}$ satisfies 
$$
	  \hQ_\varpi(\BFbeta^{(T_0)}  ) - \hQ_\varpi(\hat \BFbeta_\varpi) \leq \Delta ~~\mbox{ and }~~ \| \BFbeta^{(T_0)}  - \BFbeta^* \|_\Sigma \leq 1 
$$ with probability (over normal vectors $\{ \BFg_t\}_{t=0}^{T_0-1}$) at least $1-e^{-z}$.
 \end{theorem}

The aforementioned high-level findings demonstrate that, given a sequence of ``good events'' associated with the empirical smoothed cost function, the proposed noisy gradient descent iterates exhibit desirable convergence properties. To complement this deterministic analysis, we further provide probabilistic bounds, which subsequently yield finite sample performance bounds as presented in Sections~\ref{sec:5.1} and \ref{sec:5.2}.

\begin{proposition} \label{prop:event012}
Assume Conditions~\ref{cond.kernel}--\ref{cond.density} hold.  Given $R>0$,  for any $z>0$ we have that with probability at least $1-5e^{-z}$,
\#
	 \max_{1\leq i\leq n } \| \Sigma^{-1/2} \BFx_i \|_2&  \leq C_0 \upsilon_1 \sqrt{ p + \log(n) + z  } ,  \nn \\
	 \sup_{\BFdelta \in \Theta_\Sigma(1) \setminus \Theta_\Sigma(1/n) } \frac{| \hat D_\varpi(\BFdelta) - D_\varpi(\BFdelta) | }{\| \BFdelta \|_\Sigma } & \leq C_1  \upsilon_1   \sqrt{\frac{p + \log_2(n) + z }{n }}, \nn \\
	  \sup_{\BFbeta \in \Theta^*_\Sigma(1) }  \| \nabla \hQ_\varpi(\BFbeta) - \nabla Q_\varpi(\BFbeta) \|_{\Sigma^{-1}}  & \leq C_2 \upsilon_1 \sqrt{\frac{p \log (n/\varpi) + z}{n}} +  \frac{f_u \varpi +  2 \kappa_u}{n} \nn 
\#
and  
\#
  & \sup_{\BFbeta \in \Theta^*_\Sigma(R) }  \| \nabla^2 \hQ_\varpi(\BFbeta) -\nabla^2 Q_\varpi(\BFbeta) \|_{\Sigma^{-1}} 
  \nn \\
 &  \leq C_2 \upsilon_1^2 \bigg\{    \sqrt{\frac{p \log(n/\varpi) + z}{n \varpi }} + \frac{p \log(n/\varpi) + z}{ n \varpi } \bigg\} + C_2'  R \frac{\upsilon_1 ( p + \log n + z)^{1/2} + l_0 m_3 \varpi^2 }{n^2} , \nn
\#
provided that $n\gtrsim \upsilon_1^4 (p + z)$, where $C_0, C_1, C_2, C_2'>0$ are absolute constants. 
Moreover,  $ b^*  = \| \nabla Q_\varpi(\BFbeta^*) \|_{\Sigma^{-1} }  \leq 0.5 l_0 \kappa_2 \varpi^2$ and $\| \nabla^2 Q_\varpi(\BFbeta^*) \|_{\Sigma^{-1}} \leq f_u$.
\end{proposition}

\begin{proposition}
\label{prop5}
 In addition to Conditions~\ref{cond.kernel}--\ref{cond.density}, assume 
\$
	\kappa_l := \min_{|u|\leq 1} K(u) >0~~\mbox{ and }~~ \inf_{|u|\leq 1 }\frac{1}{2 u} \int_{-u}^u f_{\varepsilon|\xb} (v) {\rm d} v \geq f'_l ~\mbox{ almost surely} 
\$
for some $f'_l>0$.  Let $r_{{\rm loc}} = \varpi/(16\max\{ m_4, 3\}^{1/4} \upsilon_1)$. Then, for any $z>0$, we have that with probability at least $1-e^{-z}$,
\#
\hat Q_\varpi(\BFbeta_1) - \hat Q_\varpi(\BFbeta_2) - \langle \nabla \hat Q_\varpi(\BFbeta_2) , \BFbeta_1 - \BFbeta_2 \rangle \geq \phi_2 \| \BFbeta_1 - \BFbeta_2 \|_\Sigma^2    
\#
holds uniformly over $\BFbeta_1 \in \BFbeta^* +  \Theta_\Sigma(r_{{\rm loc}}/2)$ and $\BFbeta_2 \in \BFbeta_1 + \Theta_\Sigma(r_{{\rm loc}})$,  provided that the ``effective sample size'' $n \varpi $ satisfies $n \varpi \gtrsim m_4^{1/2} \upsilon_1^2(  p + z)$, where $\phi_2>0$ is a constant depending only on $(\kappa_l, f'_l)$.
\end{proposition}

Let the sample size $n$ and bandwidth $\varpi = \varpi_n >0$ satisfy
$n \varpi \gtrsim p \log n$ and $\varpi  \lesssim \{(p+\log n)/n\}^{1/4}$.
Then, Proposition~\ref{prop:event012} implies $b^* \lesssim \sqrt{(p+\log n)/n}$ and event $\cE_0(B) \cap \cE_1(\delta_0, \delta_1) \cap \cE_2(2)$ with $(B, \delta_0, \delta_1)$ satisfying
\$
		B \asymp \sqrt{p + \log n} ,   \quad  \delta_0 \asymp \sqrt{\frac{p+ \log n}{n}} ~~\mbox{ and }~~
		\delta_1 \asymp  \sqrt{\frac{p \log n}{n }}
\$
occurs with high probability. Combining this with Theorem~\ref{thm:convergence} implies the finite sample performance bounds in Theorem~\ref{thm:high.prob.bound}.

Moreover, under the additional assumptions stated in Proposition~\ref{prop5}, there exist some curvature parameter $\phi_2 > 0$ and a local radius $r_{{\rm loc}} \asymp \varpi$ such that the event $\mathcal{E}_3(r_{{\rm loc}}, \phi_2)$ also occurs with high probability. This further implies that the $\mu$-GDP estimate $\BFbeta^{(T)}$, obtained from noisy gradient descent initialized at any $\BFbeta^{(0)} \in \BFbeta^* + \Theta_\Sigma(1)$, satisfies with probability at least $1-Cn^{-1}$ that
$$
 \| \BFbeta^{(T)} - \hat \BFbeta_\varpi \|_\Sigma \lesssim \frac{\sqrt{\log n}}{(f_l \wedge f_l')^2} \frac{p+\log n}{\mu n}. 
$$
Here, we implicitly assume that both the smoothing parameter $\varpi$ and the number of iterations $T$ are chosen appropriately. The above bound, in turn, implies that the excess (smoothed) empirical risk is bounded with high probability by
$$
    \hat C_\varpi(\bbeta^{(T)}) -  \hat C_\varpi(\hat \bbeta_\varpi) =   \hat C_\varpi(\bbeta^{(T)}) -  \min_{\bbeta \in \mathbb{R}^p}\hat C_\varpi( \bbeta) \lesssim \log n \bigg( \frac{p+\log n}{\mu n} \bigg)^2.
$$
The above rate also matches (up to a logarithm factor) the one in \citet{BST2014} for Lipschitz and strongly convex loss functions (after adjusting for scaling differences).

\section{Numerical and Empirical Studies}

In this section, we utilize synthetic data as well as real-world data to showcase the empirical performance of the proposed privacy-preserving feature-driven policy. We compare its performance against that of the theoretically optimal policy, which assumes known demand but lacks privacy protection measures. For the sake of simplicity and consistency, we employ the Gaussian kernel in all of our numerical experiments.

\subsection{Synthetic Data}

We consider the linear demand model $d = \BFx^\T\BFtheta^* + \varepsilon$, where $\BFtheta^* = (1.5,1,-2.5,-1.5,3)^\T\in \mathbb{R}^5$ and $\BFx^\T = (1,\BFz^\T)^\T$. The feature vector $\BFz\in \mathbb{R}^4$ is generated from a centered multivariate normal distribution with covariance matrix $\Sigma= ( 0.5^{|j-k|})_{1\leq j, k \leq 4}$. Independent of the feature vector $\BFx$, the observation noise variable $\varepsilon$  follows one of the following three distributions: (i) standard norm distribution $\cN(0,1)$, (ii) $t$-distribution with 3 degrees of freedom ($t_3$), and (iii) Gaussian mixture distribution $0.9 \cN(0,1)+0.1 \cN(0,100)$.

\begin{figure}[ht]
   \FIGURE
   {\begin{minipage}[t]{0.3\textwidth}
\includegraphics[width=\textwidth]{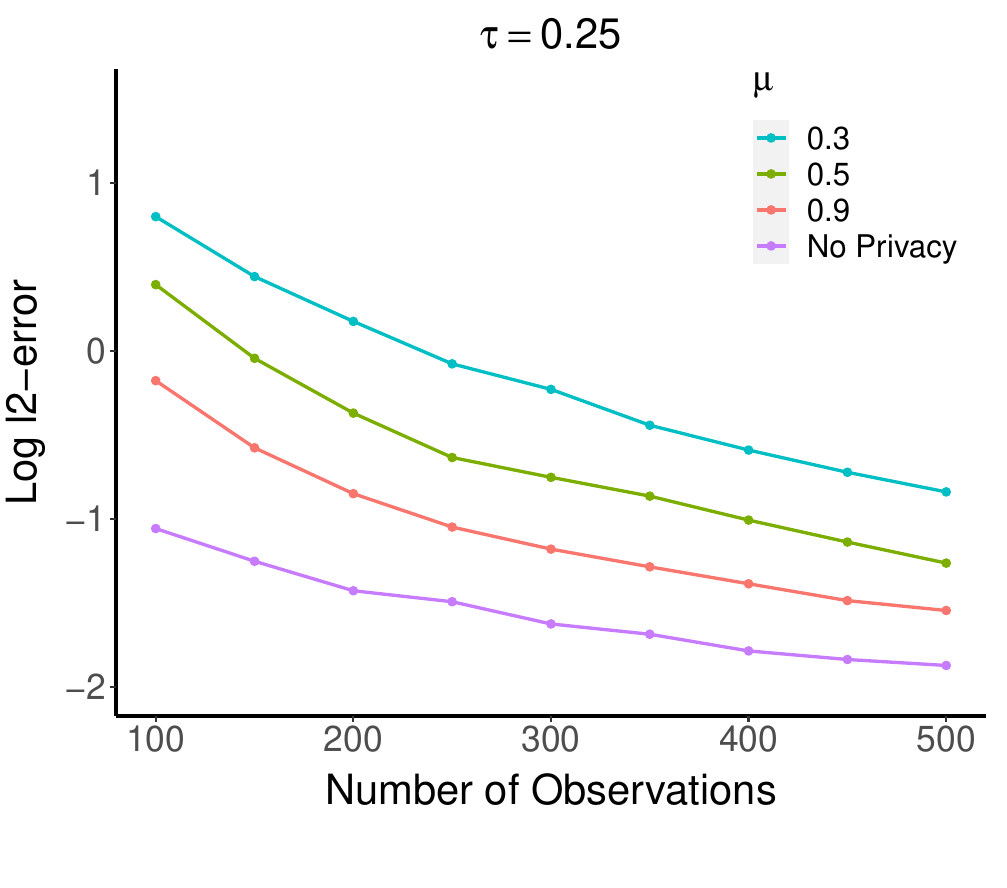}\\
\includegraphics[width=\textwidth]{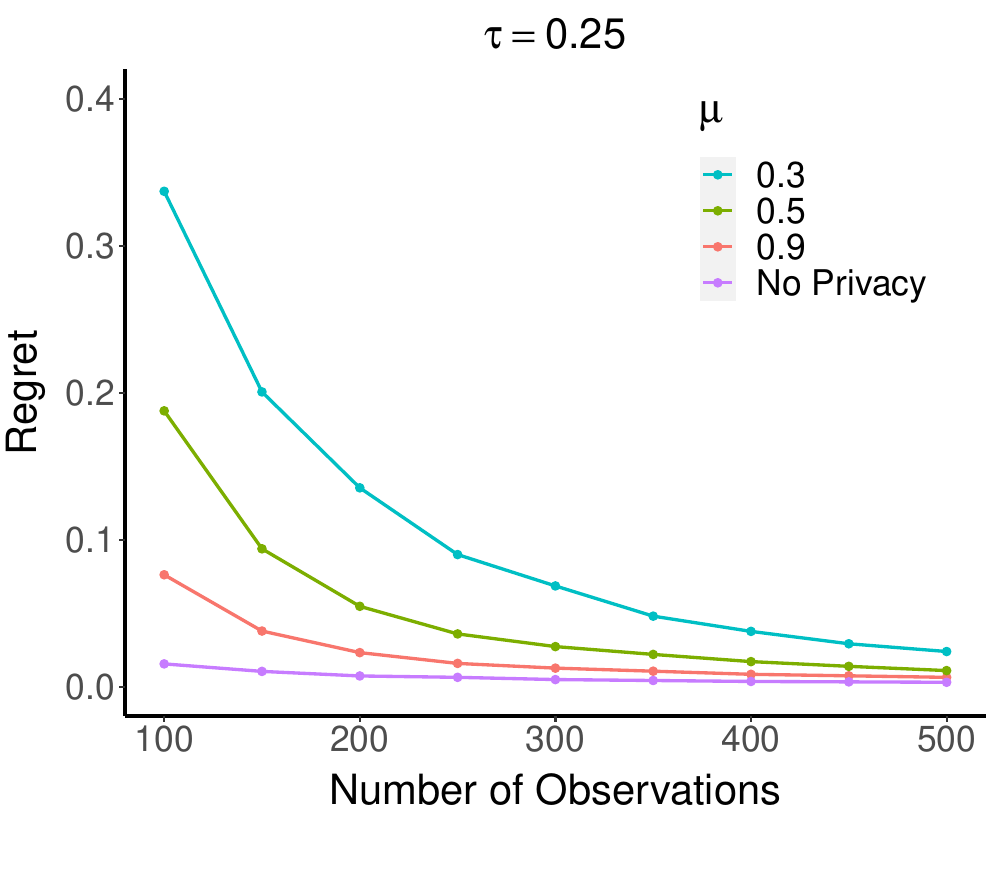}
\end{minipage}
\begin{minipage}[t]{0.3\textwidth}
\includegraphics[width=\textwidth]{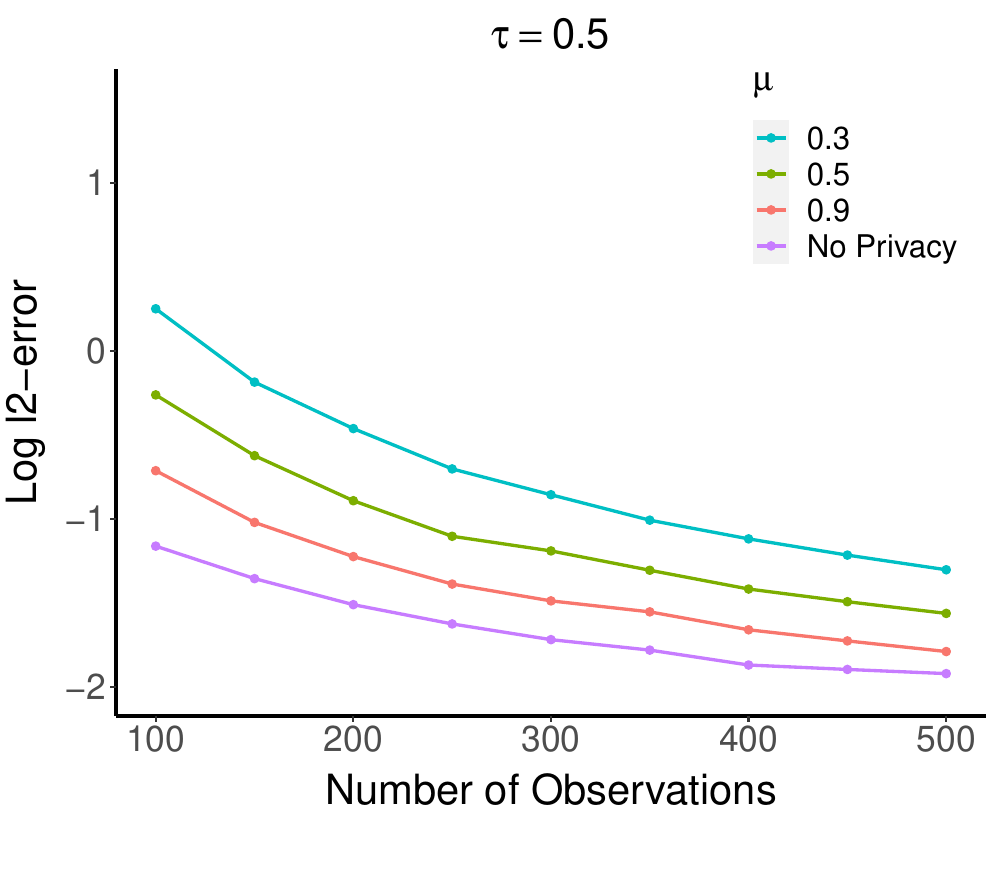}\\
\includegraphics[width=\textwidth]{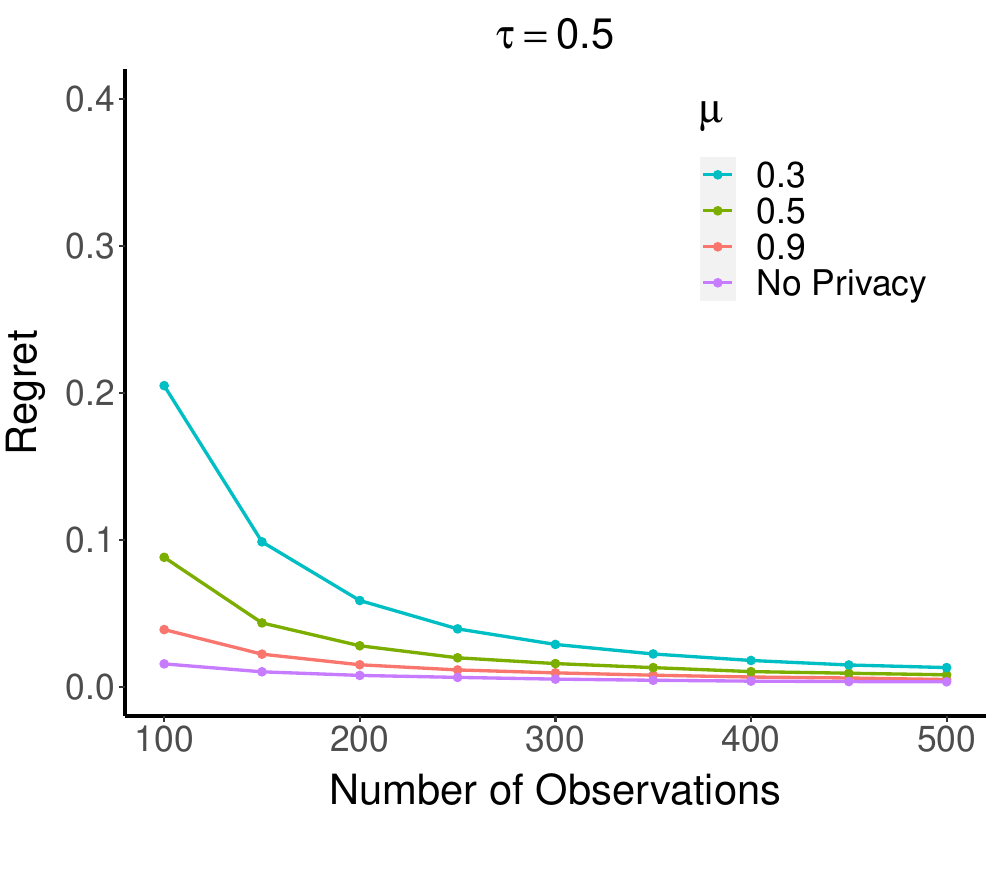}
\end{minipage}
\begin{minipage}[t]{0.3\textwidth}
\includegraphics[width=\textwidth]{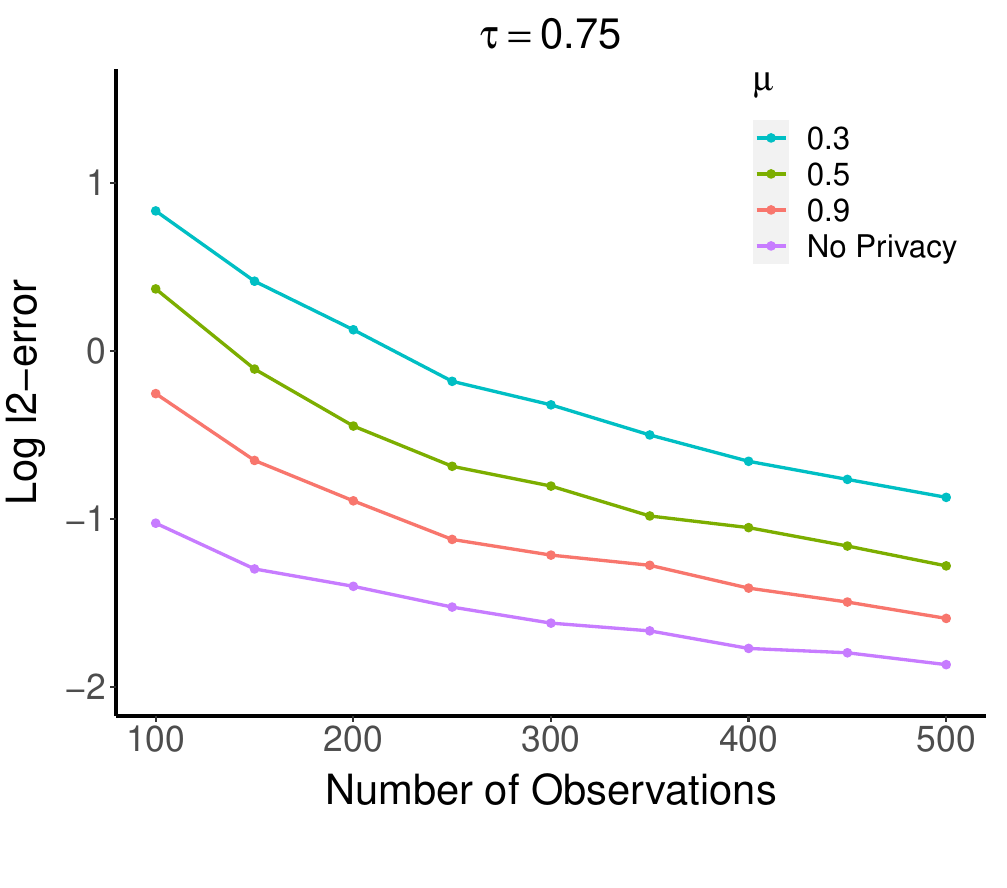}\\
\includegraphics[width=\textwidth]{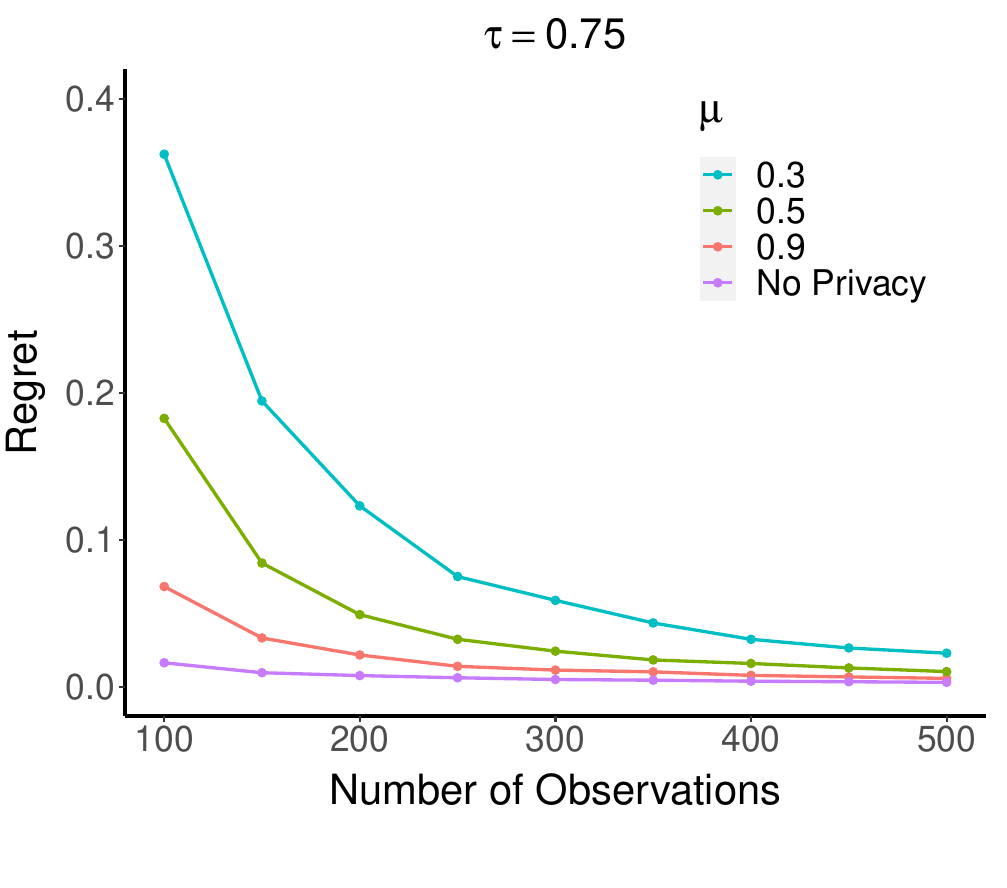}
\end{minipage}}
{Estimation errors and regrets of different estimators when $\varepsilon\sim \cN(0,1)$\label{Fig2}}
{Plots of logarithmic $\ell_2$ estimation error and regret versus the number of observations, averaged over 300 replications when $\varepsilon\sim \cN(0,1)$. }
\end{figure}

In all of our numerical experiments, we set $b+h = 1$ so that with the distribution of $\varepsilon$ known as a priori information, the optimal quantity to order can be determined by the $\tau$-th quantile of the conditional distribution of $d$ given $\BFx$, where $\tau =b$. Specifically, the clairvoyant optimal policy is $\BFx^\T \BFbeta^*$ with
$\BFbeta^* = \BFtheta^* + (Q_{\varepsilon}(\tau),0,0,0,0)^\T
$, where $Q_{\varepsilon}(\cdot)$ denotes the quantile function of $\varepsilon$.

\begin{figure}[ht]
   \FIGURE
   {\begin{minipage}[t]{0.3\textwidth}
\includegraphics[width=\textwidth]{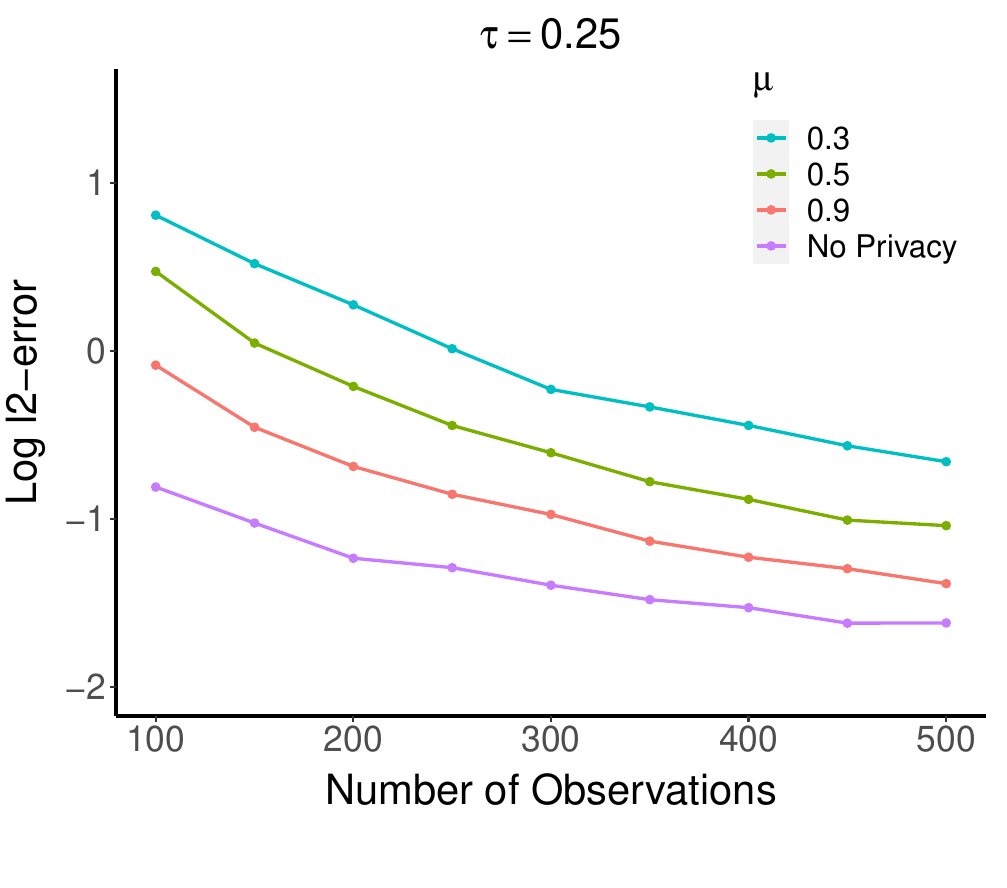}\\
\includegraphics[width=\textwidth]{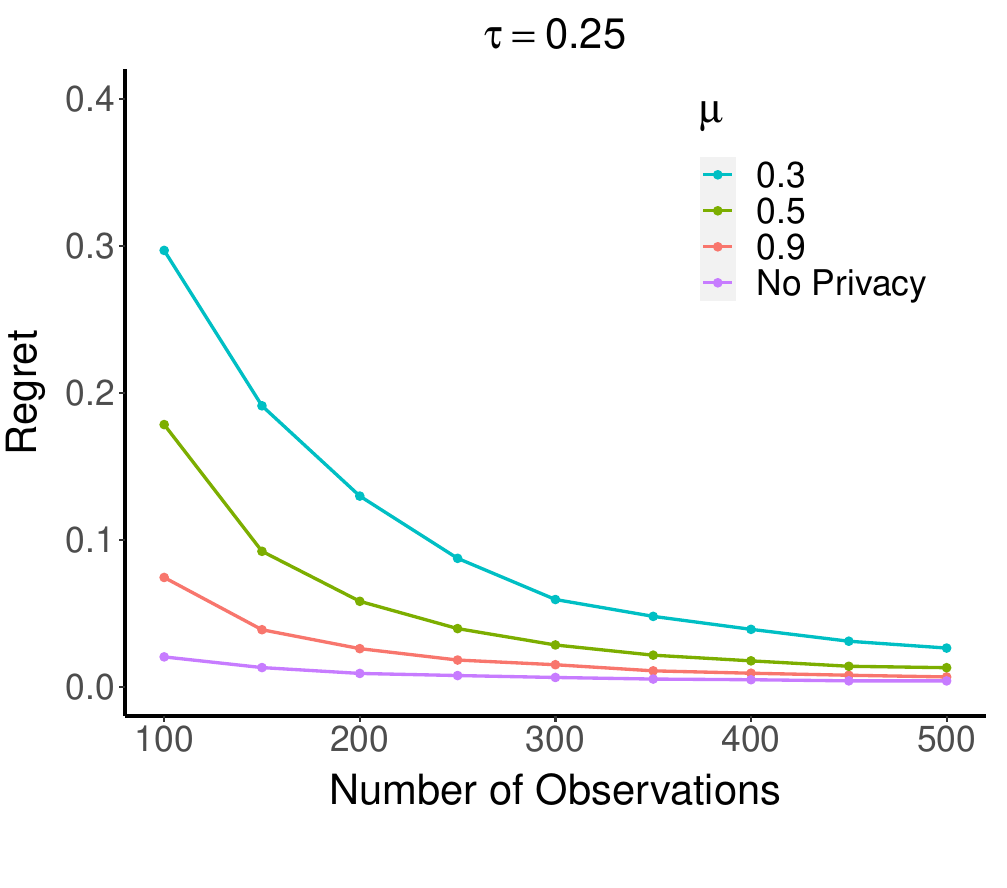}
\end{minipage}
\begin{minipage}[t]{0.3\textwidth}
\includegraphics[width=\textwidth]{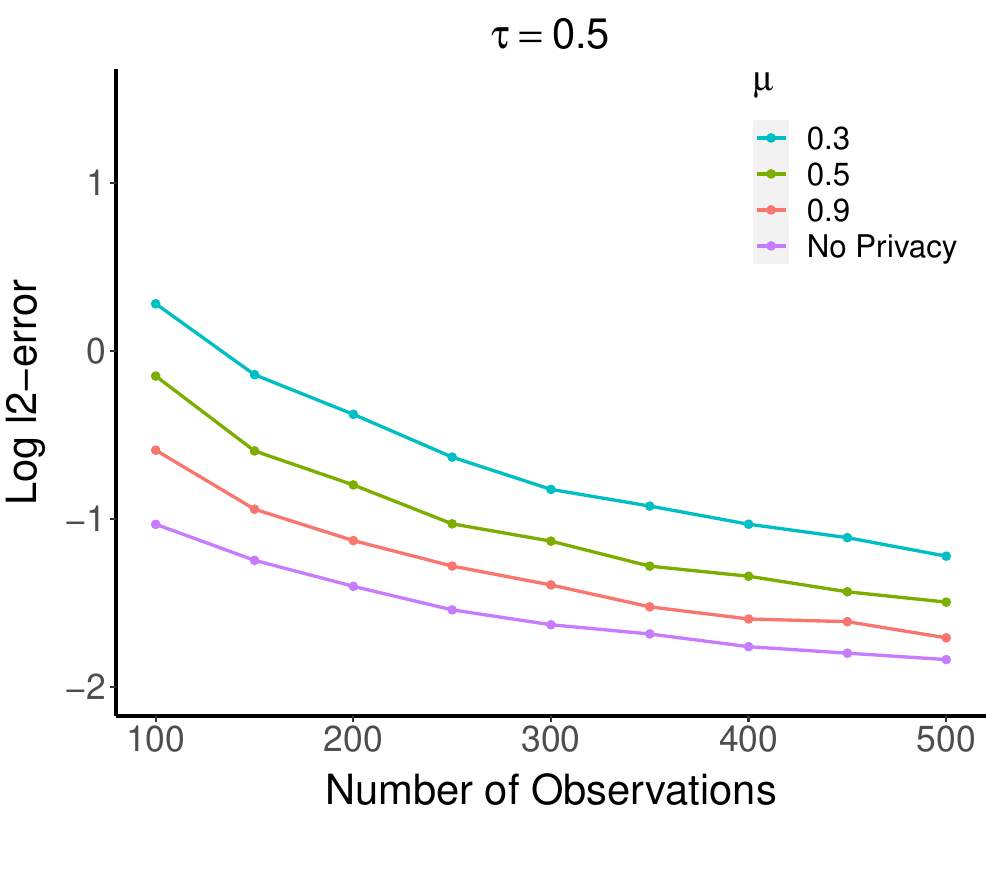}\\
\includegraphics[width=\textwidth]{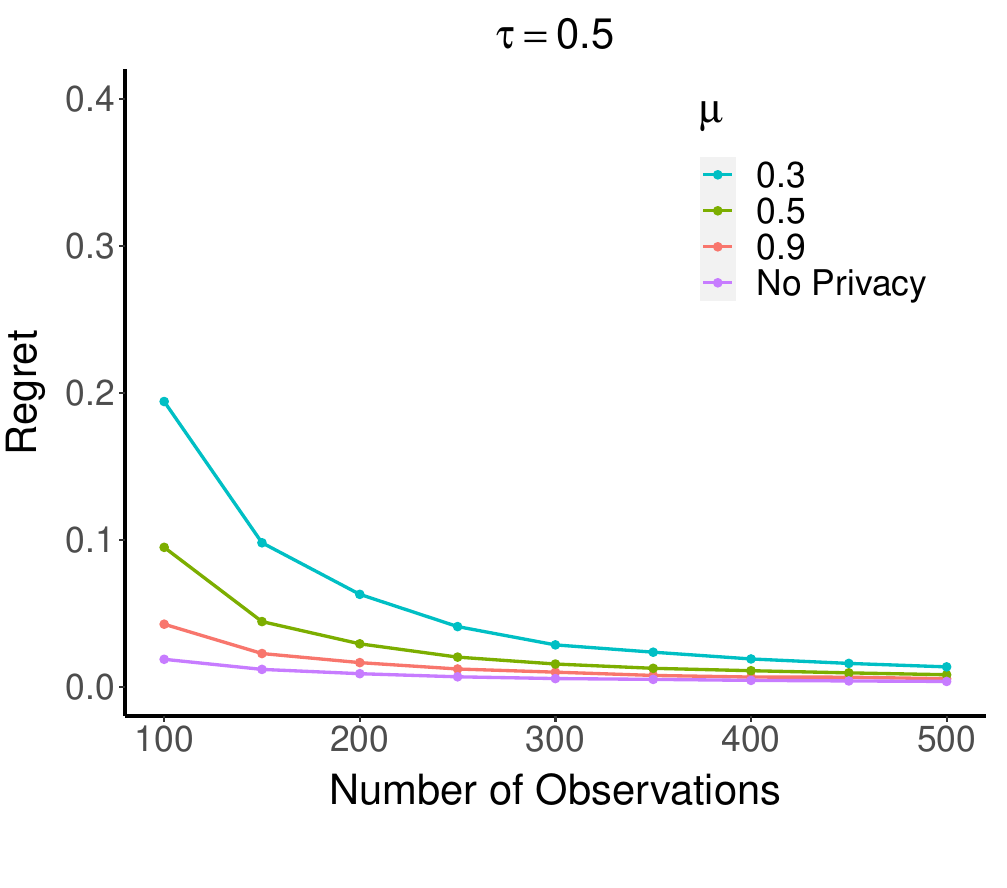}
\end{minipage}
\begin{minipage}[t]{0.3\textwidth}
\includegraphics[width=\textwidth]{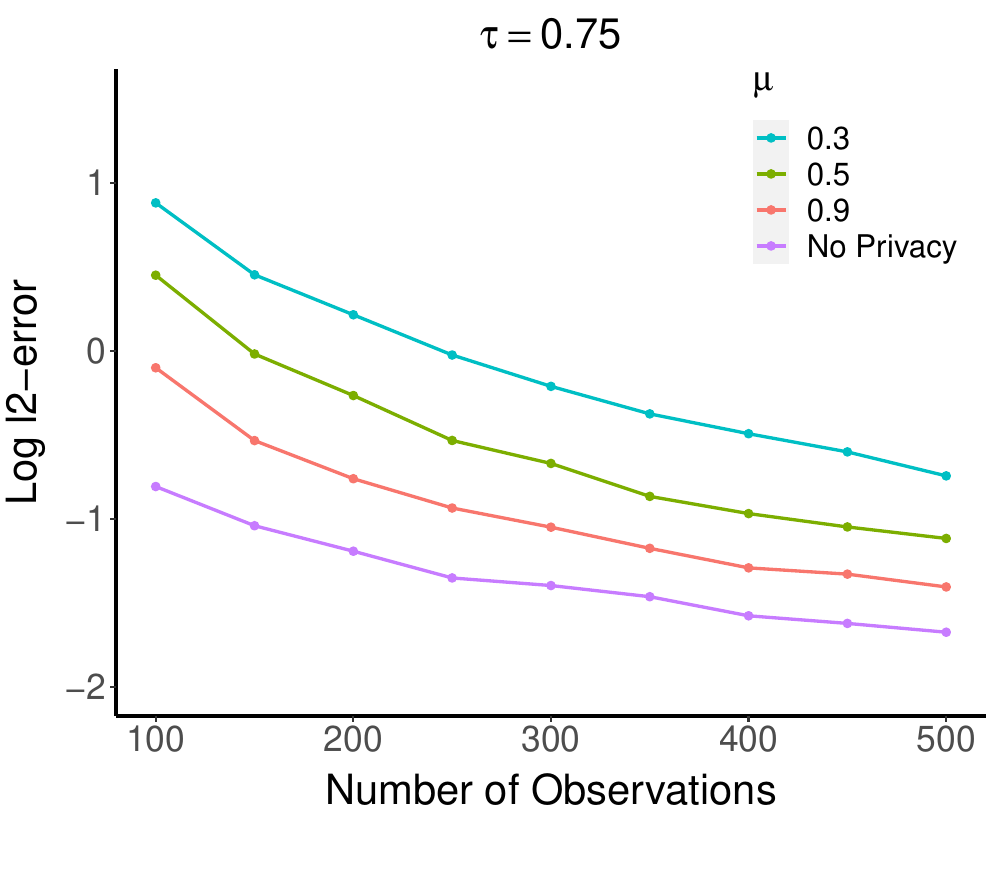}\\
\includegraphics[width=\textwidth]{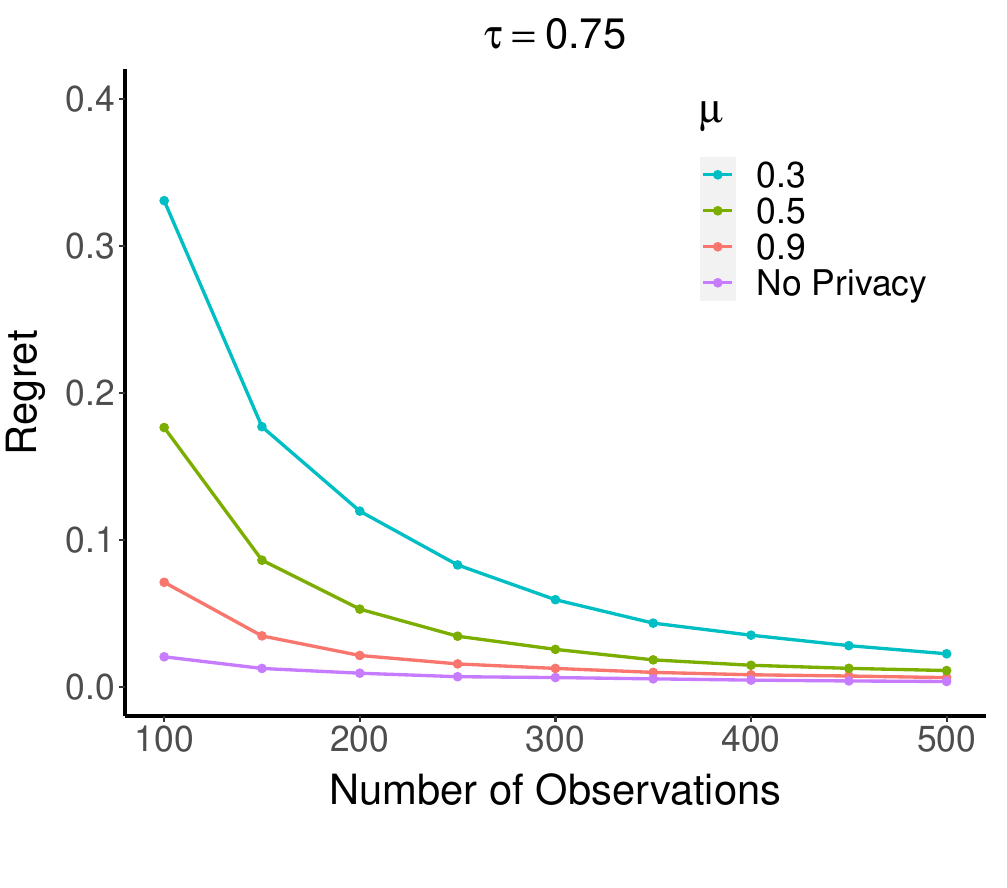}
\end{minipage}}
{Estimation errors and regrets of different estimators when $\varepsilon\sim t_3$\label{Fig3}}
{Plots of logarithmic $\ell_2$ estimation error and regret versus the number of observations, averaged over 300 replications when $\varepsilon\sim t_3$. }
\end{figure}

\begin{figure}[ht]
   \FIGURE
   {\begin{minipage}[t]{0.3\textwidth}
\includegraphics[width=\textwidth]{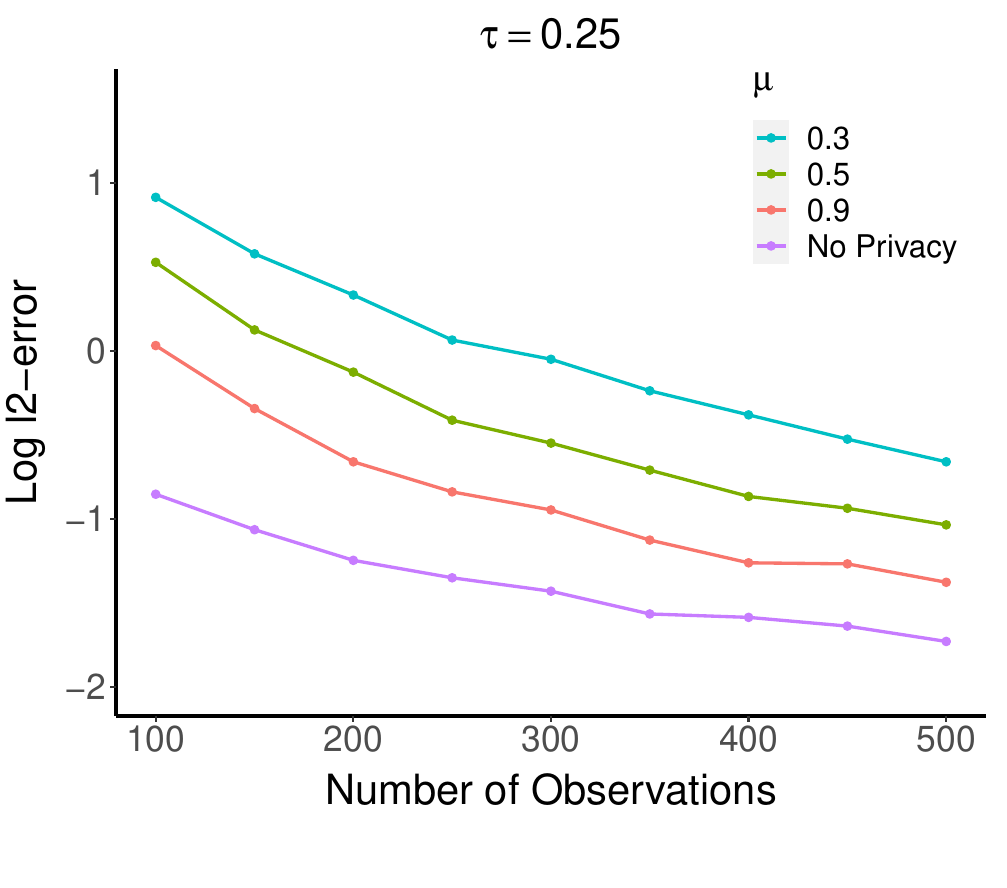}\\
\includegraphics[width=\textwidth]{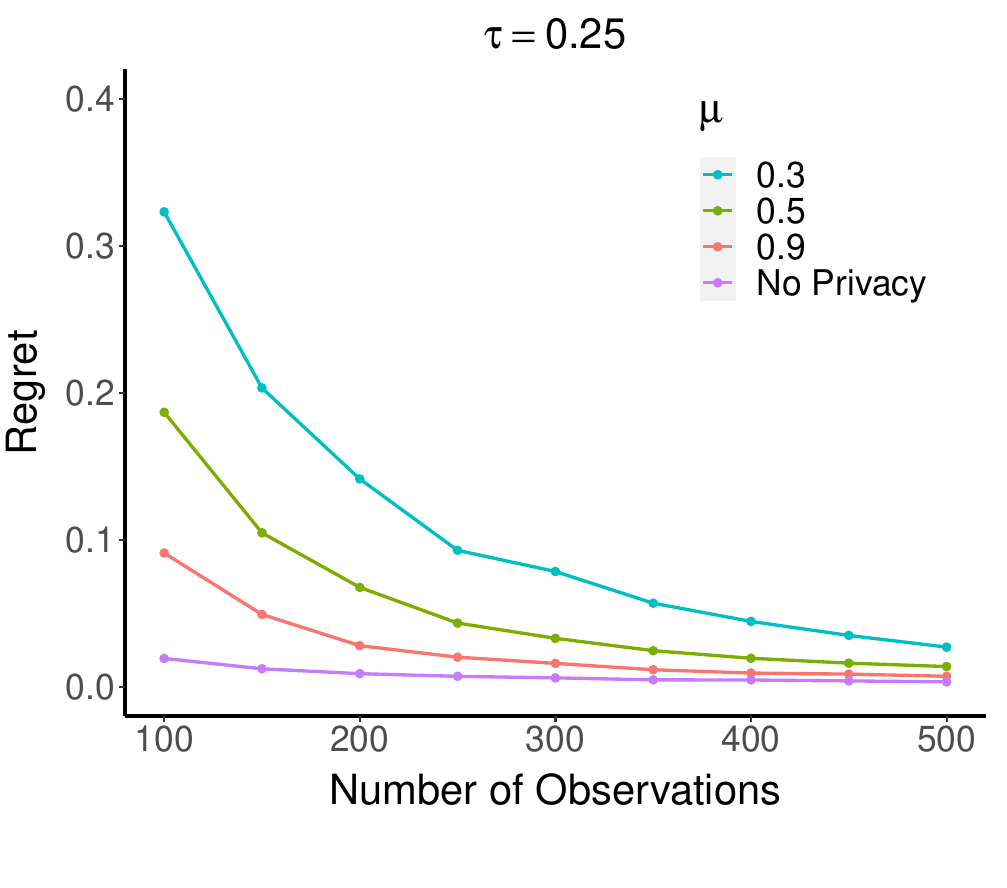}
\end{minipage}
\begin{minipage}[t]{0.3\textwidth}
\includegraphics[width=\textwidth]{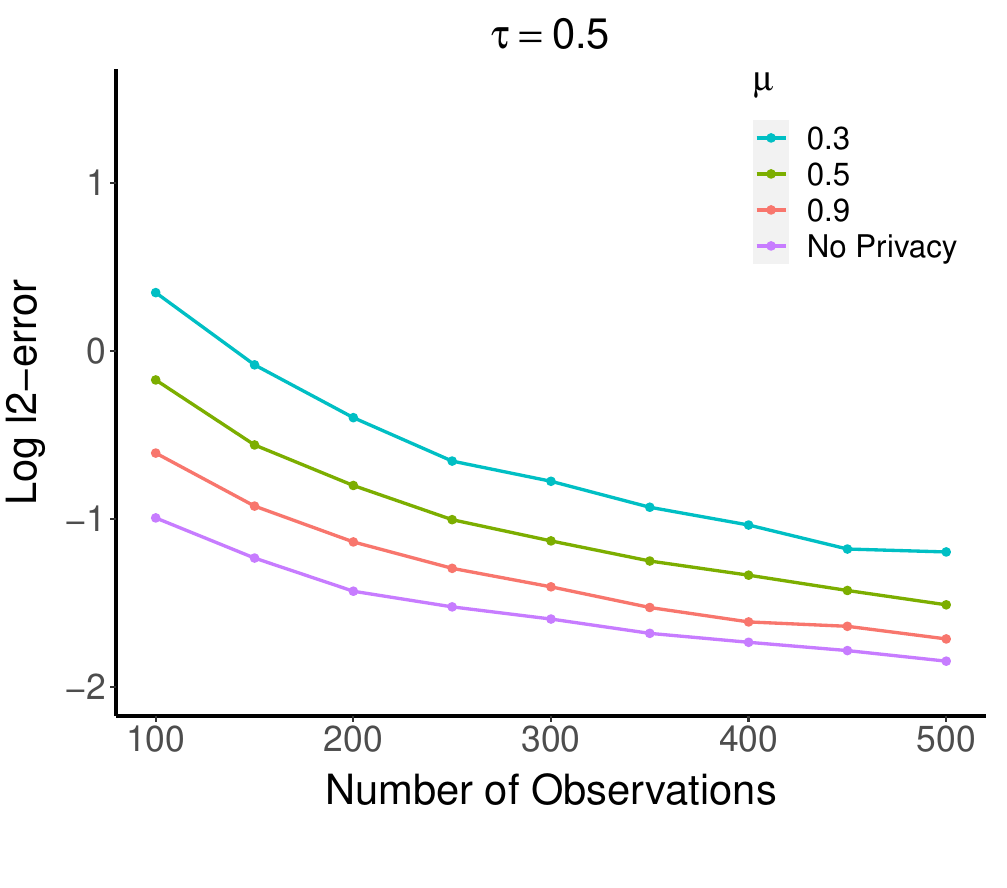}\\
\includegraphics[width=\textwidth]{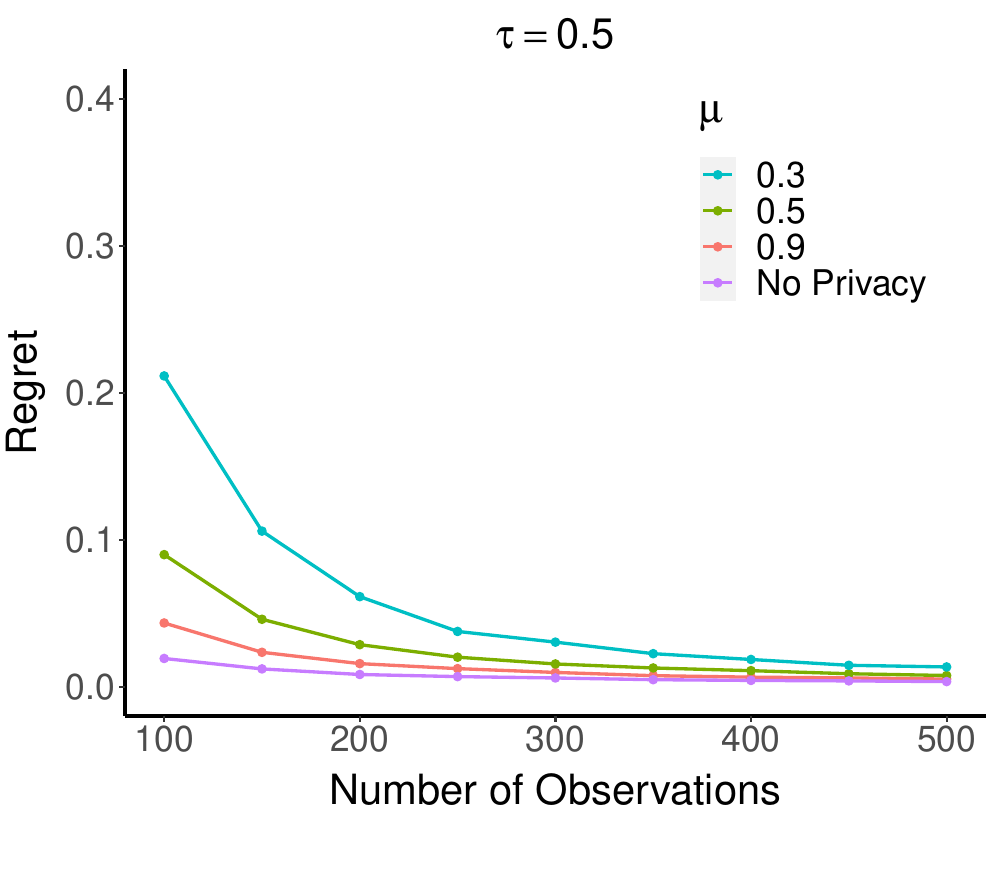}
\end{minipage}
\begin{minipage}[t]{0.3\textwidth}
\includegraphics[width=\textwidth]{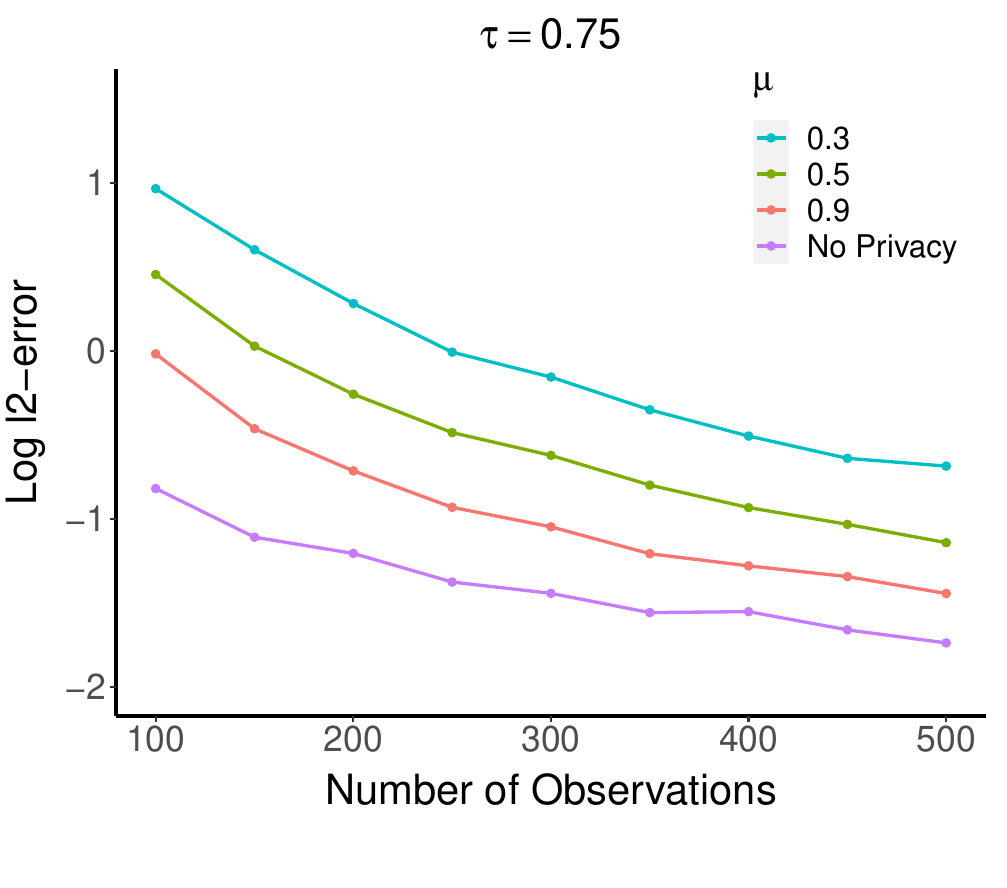}\\
\includegraphics[width=\textwidth]{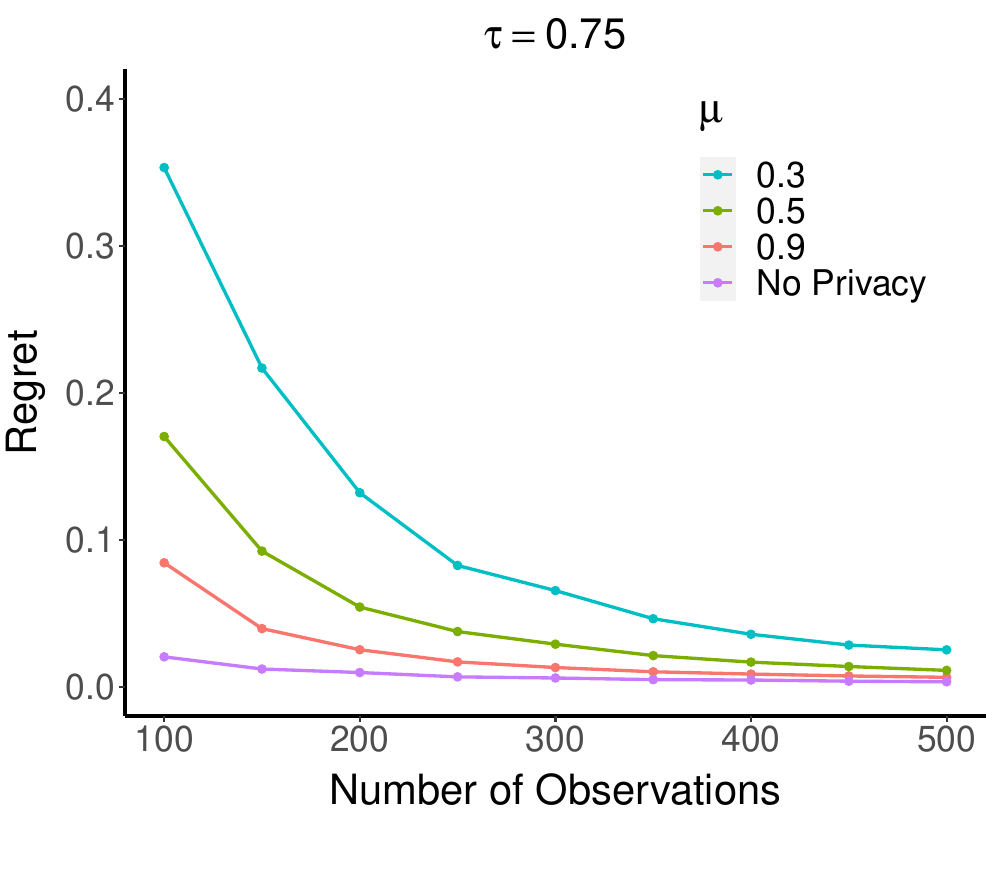}
\end{minipage}}
{Estimation errors and regrets of different estimators when $\varepsilon\sim 0.9 \cN(0,1)+0.1\cN(0,100)$\label{Fig4}}
{Plots of logarithmic $\ell_2$ estimation error and regret versus the number of observations, averaged over 300 replications when $\varepsilon\sim 0.9 \cN(0,1)+0.1\cN(0,100)$.}
\end{figure}

\begin{figure}[ht]
\FIGURE
{\includegraphics[width=1\linewidth]{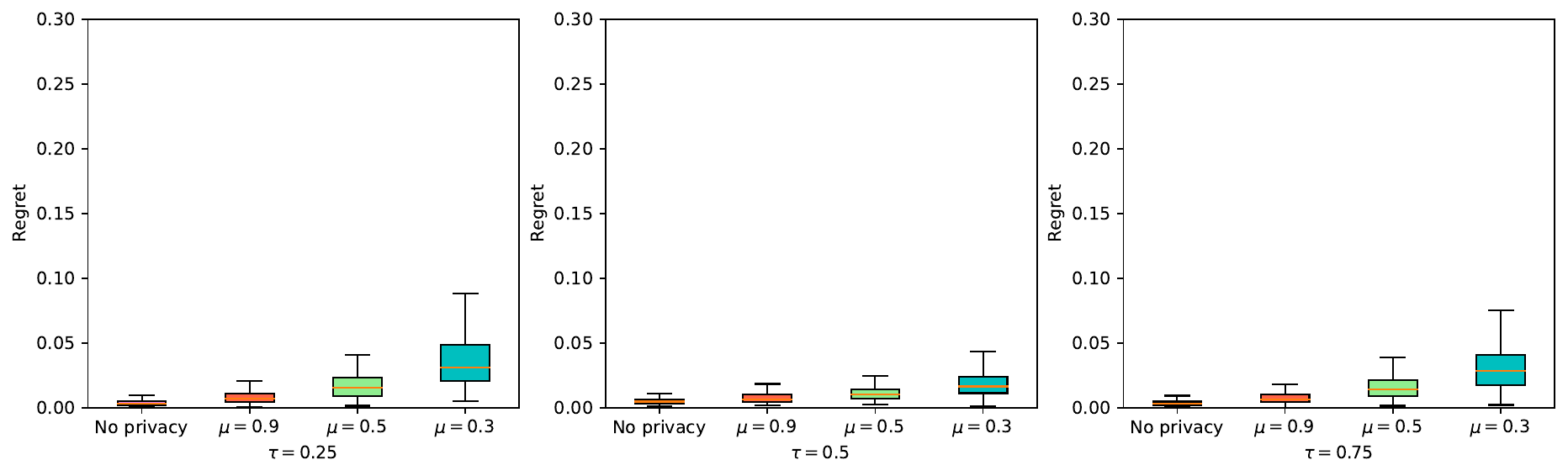}}
  {Regrets of different estimators when sample size equals 400 and $\varepsilon\sim \cN(0,1)$ \\ \label{norm-box}}
{Boxplots of regret with different privacy levels over 300 replications when the sample size is 400.} 
\end{figure}

\begin{figure}[ht]
\FIGURE
{\includegraphics[width=1\linewidth]{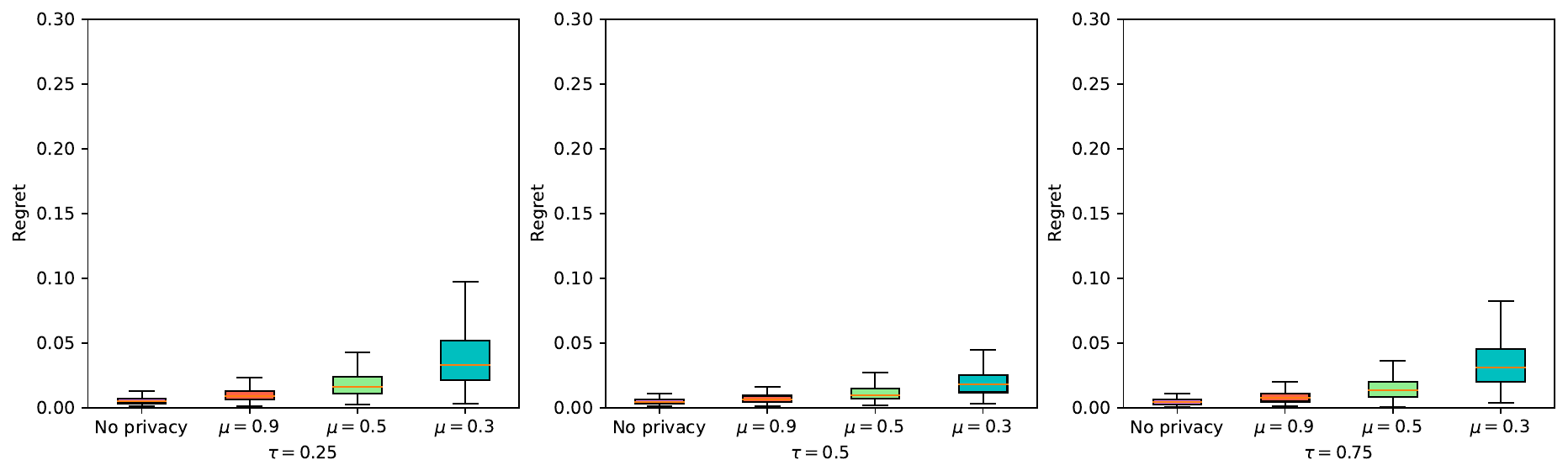}}
  {Regrets of different estimators when sample size equals 400 and $\varepsilon\sim t_3$ \\ \label{t-box}}
{Box-plots of Regret with different privacy levels over 300 replications when sample size equals 400. } 
\end{figure}

\begin{figure}[ht]
\FIGURE
{\includegraphics[width=1\linewidth]{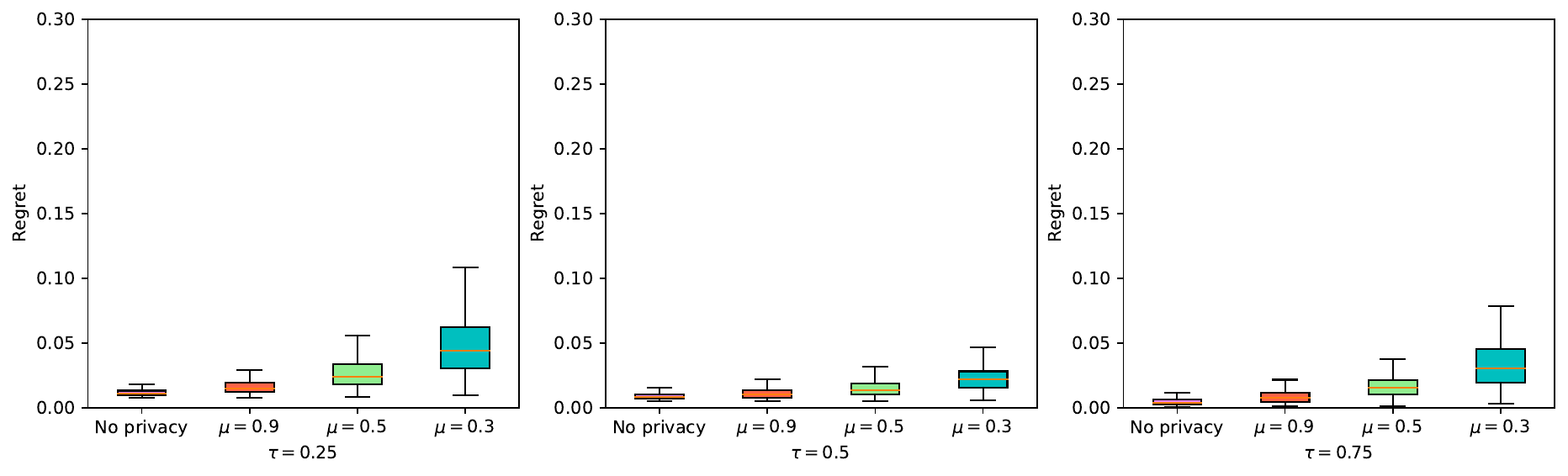}}
  {Regrets of different estimators when sample size equals 400 and $\varepsilon\sim 0.9 \cN(0,1)+0.1\cN(0,100)$ \\ \label{gmm-box}}
{Boxplots of regret with different privacy levels over 300 replications when the sample size is 400.} 
\end{figure}

For the hyper-parameters in the noisy gradient descent method, we set $T = 10$, $B = 2$ and $\sigma = \lceil 2\bar{\tau}BT^{1/2}/\mu\rceil$, where $\mu \in \{ 0.3,0.5,0.9 \}$ is privacy level and $
\tau \in \{ 0.25,0.5,0.75 \}$.  The step size $\eta_0$ is chosen via backtracking line search. As suggested in \cite{HPTZ2020}, the bandwidth $\varpi$ is taken to be $\sqrt{ \tau (1-\tau)}\cdot \{ (p + \log n)/n \}^{2/5}$. The final output $\BFbeta^{(T)}$ is $\mu$-GDP according to Theorem~\ref{prop:GDP}. We fix $p=5$ and let the sample size increase from 100 to 500. Figures~\ref{Fig2}--\ref{Fig4} present plots of the logarithmic $\ell_2$-error and regret versus the sample size under different error distributions and privacy levels, averaged over 300 repetitions. The regret of $\BFbeta^{(T)}$, defined as  $\E [ C(\BFx^\T \BFbeta^{(T)} ,d)]-\E [C(\BFx^\T \BFbeta^*,d)]$ where the expectation is taken over the joint distribution of $(d, \BFx)$, is evaluated using an additional data set of size one million. Figures~\ref{norm-box}--\ref{gmm-box} display the boxplots of regrets under different error distributions and privacy levels based on 300 repetitions with a sample size of 400. Table~\ref{Table_boxplot} reports the corresponding average regrets and standard deviations.

\begin{table}
\TABLE
{Synthetic data analysis\label{Table_boxplot}}
{ \begin{tabular}{ccccc}
\hline & Non-private& $\mu = 0.9$&$\mu = 0.5$&$\mu = 0.3$  \\
\hline
$\varepsilon\sim \cN(0,1)$ & 0.004 &0.009 &0.017 &0.038 \\
& (0.002) &(0.006) &(0.011) &(0.026)\\
$\varepsilon\sim t_3$ & 0.012 &0.017 &0.027 &0.052 \\
& (0.003) &(0.006) &(0.012) &(0.034)\\
$\varepsilon\sim 0.9 \cN(0,1)+0.1\cN(0,100)$ & 0.006 &0.01  &0.019 &0.04 \\
& (0.003) &(0.005) &(0.011) &(0.026)\\
\hline
\end{tabular} }
{Synthetic data analysis: mean and standard deviation of regret of different estimators in 3 different models, over 300
replications when the sample size equals 400.}
\end{table}

From Figures~\ref{Fig2}--\ref{Fig4} we see that both the estimation errors and regrets decrease as the number of observations grows, as expected. The spacing of these error curves further illustrates the impact of privacy. From Table~\ref{Table_boxplot} we observe that the variability of regrets is low when the sample size is reasonably large. These numerical results also highlight the robustness of newsvendor loss minimization against heavy-tailed error distributions.

\subsection{Real Data Example}

We demonstrate the effectiveness of the proposed privacy-preserving algorithm using the restaurant data from \cite{buttler2022meta}. This dataset comprises demand data for main ingredients at a casual restaurant in Stuttgart over approximately 750 days.  The restaurant manager needs to decide on the amount of ingredients to defrost overnight to prepare meals, considering that leftover ingredients result in holding costs. Therefore, we formulate the problem of determining the optimal amount of ingredients to defrost as a newsvendor problem. It is worth noting that during the data collection period, the store manager's strategy was to maintain a service level of nearly 100\%, rendering the issue of censored demand negligible.

\begin{table}
\TABLE
{Restaurant data analysis\label{Table1}}
{ \begin{tabular}{ccccc}
\hline & Non-private& $\mu = 0.9$&$\mu = 0.5$&$\mu = 0.3$  \\
\hline
 $b=50$ & 313.08& 315.87 &316.71 &317.49 \\
 $b=70$ & 365.67 &365.75 &367.09 &369.32 \\
 $b=90$  & 405.45& 405.22& 407.47& 410.43 \\
 $b=120$  & 452.45& 453.07& 456.21& 459.89\\
\hline
\end{tabular} }
{Restaurant data analysis: average 
    out-of-sample cost of the private estimator and 
    non-private estimator.}
\end{table}

In our analysis, we utilize the private algorithm to determine the optimal strategy for defrosting the amount of lamb (which is the ingredient with the highest demand) to minimize costs and maximize performance. We compare the outcomes obtained from the private algorithm with those from the standard non-private algorithm. Our model incorporates three distinct features: (1) calendric features, which include binary variables indicating holidays or non-holidays extracted from the date; (2) lag features, incorporating demand information from the previous periods, such as demand from exactly one week ago and exactly two weeks ago; and (3) weather features, encompassing rain and temperature data. We assume a per-unit (per kilogram) holding cost $h$ for lamb of \textdollar30. We consider four different values for the lost-sales penalty cost $b$: \textdollar50, \textdollar70, \textdollar90, and \textdollar120 per kilogram. These values correspond to gross profit margins (excluding labor costs) of roughly 62.5\%, 70\%, 75\%, and 80\% respectively, which are similar to the gross profit margin of a financially viable restaurant, estimated to be around 70\%. We use a training dataset of $n = 552$ past demand observations to train our model and evaluate its performance by measuring the out-of-sample error on a separate testing dataset consisting of $184$ observations.
 
The algorithm hyper-parameters are set as $T = 10$, $B = 2$ and $\sigma = \lceil 2\bar{\tau}BT^{1/2}/\mu\rceil$ with privacy level $\mu \in \{ 0.3, 0.5, 0.9\}$. We conduct 100 random partitions of the dataset into training and testing data and summarize the average out-of-sample cost across these partitions. Table~\ref{Table1} presents the out-of-sample cost of our private estimator (at different privacy levels) and the naive non-private estimator for various choices of $b$. The average cost of our private algorithm is at most 2\% higher than the cost of the non-private algorithm. This indicates that the proposed algorithm can be effectively used by the restaurant to predict future demands while maintaining a reasonable level of privacy protection, albeit at a slightly higher cost.

\section{Concluding Remarks and Discussions}

In this paper, we investigate the learning of privacy-preserving optimal policies for feature-based newsvendor problems with unknown demand. We consider the problem within the framework of $f$-differential privacy, a recently proposed approach that extends the classical $(\epsilon, \delta)$-differential privacy with several appealing features. To address the challenge of nonsmoothness associated with the newsvendor loss function, we propose a new noisy gradient algorithm based on convolution smoothing.  We provide privacy-preserving guarantees for the $T$-step output of the proposed algorithm and establish rigorous finite-sample high-probability bounds for estimation error and regret. Importantly, we demonstrate that a reasonable level of privacy protection can be achieved without sacrificing performance compared to the clairvoyant policy with known demand distribution but without privacy protection.

A future endeavor is to find proper conditions on the mini-batch size $m$ under which the noisy SGD estimators are consistent. If $m$ is fixed, there will be a non-vanishing noise term in noisy SGD. Thus, we might not have consistent noisy SGD estimators unless $m\to \infty$, and the cost of privacy might not be negligible unless we also have that $m^2 /n \to \infty$. These problems deserve further attention in future research. From the practical perspective, the choice of appropriate mini-batch size is critical and nontrivial to ensure high-quality performance. In the newsvendor problem, when the data size is not on the scale of millions, the full gradient descent method remains computationally efficient and exhibits fast geometric convergence.

%
%
%
\ACKNOWLEDGMENT{The authors are grateful to the department editor, associate editor,
and two anonymous referees for their extensive and constructive
comments. The research of Zhao and Wang was partially supported by Grant NSF FRGMS-1952373.}

\begin{APPENDICES}

\section{Technical Lemmas}

This section presents several technical lemmas that are essential for certifying that events $\cE_0(B)$, $\cE_1(\delta_0,\delta_1)$ and $\cE_2(R)$ defined in \eqref{def:E0}--\eqref{def:E2} hold with high probability for properly chosen $(B, \delta_0, \delta_1, R)$.
For $\BFbeta\in \RR^p$, after a change of variable $\BFdelta = \BFbeta - \BFbeta^*$ we define
\#
	 \hat R_\varpi(\BFdelta) &=\hQ_\varpi(\BFbeta) - \hQ_\varpi(\BFbeta^* ) - \langle \nabla \hQ_\varpi(\BFbeta^*) , \BFbeta - \BFbeta^* \rangle  ,  \label{def:Rh} \\
	  \hat B_\varpi(\BFdelta) & =\hQ_\varpi(\BFbeta^*) - \hQ_\varpi(\BFbeta ) - \langle \nabla \hQ_\varpi(\BFbeta ) , \BFbeta ^*- \BFbeta  \rangle , \label{def:Bh}
\#
and their population counterparts $R_\varpi(\BFdelta)  = \E \{  \hat R_\varpi(\BFdelta) \}$ and $B_\varpi(\BFdelta)  = \E \{  \hat B_\varpi(\BFdelta) \}$.

Lemma~\ref{lem:population.loss.curvature} provides lower bounds for the first-order Taylor series remainder of the population (smoothed) loss in a neighborhood of $\BFbeta^*$, that is, $R_\varpi(\cdot)$ and $B_\varpi(\cdot)$. 
Lemma~\ref{lem:loss.diff} provides upper bounds (in high probability) for the fluctuation of the empirical loss $|\hat   D_\varpi(\BFdelta) -  D_\varpi(\BFdelta)|$ uniformly over $\BFdelta$ in a compact set excluding a local neighborhood of the origin. 
Lemma~\ref{lem:grad.diff} establishes uniform convergence of the gradient and Hessian
of the smoothed quantile loss to their population counterparts, as long as the sample size scales linearly in the number of features.
 
Note that under Condition~\ref{cond.subgaussian},  $m_q = \sup_{\BFu \in \mathbb{S}^{p-1} } \E | \langle \BFw, \BFu\rangle|^q$ ($q=3,4$)  are bounded and only depend on $\upsilon_1$.

\begin{lemma}  \label{lem:population.loss.curvature}
Conditions~\ref{cond.kernel}--\ref{cond.density} ensure that 
\$
   \min\big\{ 	R_\varpi(\BFdelta ) , B_\varpi(\BFdelta) \big\} \geq 0.5 \big( f_l - l_0 \kappa_1 \varpi \big) \| \BFdelta \|_{\Sigma}^2 ~\mbox{ for all } \BFdelta \in \Theta_\Sigma(1).
\$
\end{lemma}

 \begin{lemma} \label{lem:loss.diff}
Let $r > \delta >0$.  For any $u\geq 0$, the following bounds hold:
\begin{enumerate}
\item[(i)] With probability at least $1-e^{-u}$,
\#
\hat D_\varpi(\BFdelta )   - D_\varpi(\BFdelta)   \leq  C_0  \upsilon_1 r \sqrt{\frac{p + u}{n}}   ~~\mbox{ for all }  \BFdelta \in \Theta_\Sigma(r) ; \label{loss.diff.max.bound}
\#
\item[(ii)]  With probability at least $1-e^{-u}$,
 \#
	\hat D_\varpi(\BFdelta )   - D_\varpi(\BFdelta)   \leq  C_1 \upsilon_1\sqrt{\frac{p+ \log_2(r/\delta ) + u}{n}}  \cdot \| \BFdelta \|_{\Sigma} ~~\mbox{ for all }  \BFdelta \in \Theta_\Sigma(r)\setminus {\Theta}_\Sigma(\delta) , \label{loss.diff.uniform.bound}
\#
\end{enumerate}
where $C_0, C_1>0$ are absolute constants. The same upper bounds in \eqref{loss.diff.max.bound} and \eqref{loss.diff.uniform.bound}  also apply to $ D_\varpi(\BFdelta)  - \hat D_\varpi(\BFdelta)$.  
\end{lemma}

\begin{lemma}  \label{lem:grad.diff}
Assume Conditions~\ref{cond.kernel}--\ref{cond.density} hold and let $R>0$ be a fixed constant. For any $u\geq 0$, the following bounds hold with probability at least $1-3e^{-u}$ as long as $n\gtrsim \upsilon_1^4 (p+u)$: $\max_{1\leq i\leq n} \| \Sigma^{-1/2} \BFx_i \|_2  \leq    1 + \upsilon_1    \sqrt{p-1} +\upsilon_1     \sqrt{2(u+\log n)} $, 
\#
& \sup_{\BFbeta \in \Theta^*_\Sigma(R)}   \|  \nabla \hat Q_\varpi(\BFbeta) - \nabla Q_\varpi(\BFbeta)    \|_{\Sigma^{-1}} \leq C_2  \upsilon_1 \sqrt{\frac{p \log (n/\varpi )  +u}{n}} + ( f_u \varpi + 2\kappa_u ) \frac{R}{n  }   \label{grad.unif.convergence} 
\#
and
\#
  &  \sup_{\BFbeta \in \Theta^*_\Sigma(R)}  \| \nabla^2 \hQ_\varpi(\BFbeta) - \nabla^2 Q_\varpi(\BFbeta) \|_{\Sigma^{-1}} \nn \\
  & \lesssim  \upsilon_1^2 \bigg\{   \sqrt{\frac{p\log (n/ \varpi ) + u}{ n \varpi }} + \frac{p\log (n/\varpi )+ u}{n \varpi }\bigg\} +   \big\{ \upsilon_1 (p+ \log n+ u)^{1/2}   + l_0  m_3  \varpi^2 \big\} \frac{R}{n^2} .   \label{hessian.unif.convergence}
\# 
Moreover,  $\|  \nabla   Q_\varpi(\BFbeta^*) \|_{\Sigma^{-1}}  \leq 0.5 l_0 \kappa_2 \varpi^2$ and $ \sup_{\BFbeta \in \mathbb{R}^p} \|  \nabla Q_\varpi(\BFbeta)    \|_{\Sigma^{-1}} \leq \bar \tau = \max(\tau, 1-\tau)$.

\end{lemma}

The following lemma provides upper bounds for the i.i.d. standard normal random vectors $\{ \BFg_t\}_{t=0}^{T-1}$ in the noisy gradient descent algorithm. In particular, inequality \eqref{weighted.chi-square.concentration} slightly improves the tail bound in Lemma~11 of \cite{CWZ2021}.

\begin{lemma} \label{lem:max.normal.l2}
Let $\BFg_0,  \BFg_1,\ldots,  \BFg_{T-1}$ ($T \geq 1$) be independent random vectors from $\cN(\BFzero,   \BFI_p)$. Then, for any $z\geq 0$,
\#
	 P \bigg\{  \max_{0\leq t\leq T-1} \| \BFg_t \|_2 \geq   \sqrt{p} + \sqrt{2(\log T + z ) }   \bigg\} \leq e^{-z} .  \label{max.chi-square.concentration}
\#
Moreover, for any $\rho\in (0,1)$, we have with probability at least $1-e^{-z}$ that
\#
	\sum_{t=0}^{T-1} \rho^t \| \BFg_t \|_2^2 \leq \frac{p}{1-\rho} + 2 \sqrt{\frac{p z}{1-\rho^2}} + 2 z   \leq \frac{2p}{1-\rho} +\bigg(\frac{1}{1+\rho} + 2 \bigg)z . \label{weighted.chi-square.concentration}
\#
\end{lemma}

\begin{lemma} \label{lem:GD.compact}
Let $z\geq 0$. In addition to \eqref{parameters.cond1},  assume
\#
   n \geq B_T \sigma /\Delta   ,  \label{parameters.cond3}
\#
where $B_T :=  \sqrt{p} + \sqrt{2(\log T + z ) }$ and  $\Delta =  \phi_1 -  \delta_0 - b^* \in (0, f_l/2)$. 
Then,  conditioning on  $\cE_0(B) \cap \cE_1(\delta_0, \delta_1) \cap \cE_2(2)$,   the  noisy gradient descent iterates satisfy $\BFbeta^{(t)} \in \Theta_\Sigma^*(1)$ for all $t= 1,\ldots, T$ with probability (over $\{ \BFg_t \}_{t=0}^{T-1}$) at least $1-e^{-z}$.
\end{lemma}  

\begin{lemma} \label{lem:initial.value}
Let $R_0 = \| \BFbeta^{(0)} - \hat \BFbeta_\varpi \, \|_\Sigma$, and assume $ \| \hat \BFbeta_\varpi - \BFbeta^* \|_\Sigma \leq r_0$ and $\eta_0 \in (0, 1/(2f_u)]$.  For any $T_0\geq 2$ and $z\geq 0$, assume the sample size satisfies
\$
  n \geq   \frac{e-1}{4-e} (2 \bar \tau B + 1) \max\{ 1, \eta_0 / R_0 \}^2  T_0  B_{T_0}\sigma     ,
\$
where $B_{T_0} :=  \sqrt{p} + \sqrt{2(\log T_0 + z ) }$. Then, conditioned on the event $\cE_0(B) \cap  \cE_2(R)$ with $R\geq2R_0   + r_0$, we have
\$
	 \| \BFbeta^{(t)} - \hat \BFbeta_\varpi \, \|_\Sigma^2 \leq (1+1/T_0)^t R_0^2 +   \{ (1+1/T_0)^t - 1   \} ( 2 \bar \tau B + 1  )   \eta_0^2 \frac{   T_0  B_{T_0}\sigma  }{n}  , \ \  1\leq t\leq T_0 
\$
with probability (over i.i.d. normal vectors $\{ \BFg_t\}_{t=0}^{T_0-1} $) at least $1- e^{-z}$. In particular, $\max_{1\leq t\leq T_0} \| \BFbeta^{(t)} - \hat \BFbeta_\varpi \, \|_\Sigma \leq  2 R_0$.
\end{lemma}

\section{Proof of Theorems}
\subsection{Proof of Theorems~\ref{thm:high.prob.bound} and \ref{thm:regret.bound}}

The high probability bound in Theorem~\ref{thm:high.prob.bound} is a direct consequence of Theorem \ref{thm:convergence} and Proposition~\ref{prop:event012} by taking $z=2\log n$ therein. To obtain the second bound under expectation, note that
$$
     \| \BFbeta^{(1)} - \BFbeta^{(0)} \|_\Sigma 
     \leq  \bigg\| \frac{\eta_0}{n} \sn \{ \bar K_{\varpi}(\BFx_i^\T \BFbeta^{(0)}) - \tau \} \overbar \BFw_i  + \frac{\eta_0 \sigma }{n}\BFg_0 \bigg\|_2  \leq \bar \tau \eta_0 B  + \frac{\eta_0 \sigma }{n} \| \BFg_0 \|_2 ,
$$
where we have used the property that $\max_{1\leq i\leq n}\| \overbar \BFw_i \|_2 \leq B$. Similarly, it can be shown that for each $t\geq 1$, $\|  \BFbeta^{(t)} - \BFbeta^{(t-1)} \|_\Sigma \leq \bar \tau \eta_0 B + \frac{\eta_0 \sigma}{n} \| \BFg_t \|_2$. With $\sigma = 2 \bar \tau B T^{1/2} / \mu$, applying the triangle inequality repeatedly yields
$$
    \|  \BFbeta^{(T)} - \BFbeta^{(0)} \|_\Sigma \leq \sum_{t=1}^T \|  \BFbeta^{(t)} - \BFbeta^{(t-1)} \|_\Sigma \leq \bar \tau \eta_0 B T +  2 \bar{\tau}\eta_0 B T^{1/2} \frac{ 1 }{\mu n} \sum_{t=0}^{T-1}  \| \BFg_t \|_2 . 
$$
Initialized at any $\BFbeta^{(0)}$ satisfying $\| \BFbeta^{(0)}  - \BFbeta^* \|_\Sigma \leq 1$, it holds almost surely that
\begin{align}
 \|  \BFbeta^{(T)} - \BFbeta^* \|_\Sigma \leq 1 + \bar \tau \eta_0 B T +  2 \bar \tau \eta_0 B T^{1/2} \frac{ 1}{\mu n} \sum_{t=0}^{T-1}  \| \BFg_t \|_2  . \label{crude.err.bound}
\end{align}
From the first high probability bound, we see that there exists an event $\cE$ such that $P(\cE) \geq 1 - C_1 n^{-2}$ and on $\cE$ it holds
$$
    \|  \BFbeta^{(T)} - \BFbeta^* \|_\Sigma \leq r_n\asymp   T^{1/2}  \eta_0 \frac{p+\log n}{  \mu n} + \frac{1}{f_l}\sqrt{ \frac{p\log n}{n}} .
$$
By combining the above two bounds, with one being almost surely true and the other holding with high probability, we further derive that
\begin{align*}
   &  \E \|  \BFbeta^{(T)} - \BFbeta^* \|_\Sigma = \E \{ \|  \BFbeta^{(T)} - \BFbeta^* \|_\Sigma \mathbbm{1}(\cE) \} +  \E \{ \|  \BFbeta^{(T)} - \BFbeta^* \|_\Sigma \mathbbm{1}(\cE^{{\rm c}}) \} \\ 
   & \leq r_n + (1 + \bar \tau \eta_0 B T ) P(\cE^{{\rm c}}) +  2 \bar \tau \eta_0  B T^{1/2}\frac{1}{\mu n} \sum_{t=0}^{T-1}   \E  \{ \| \BFg_t \|_2  \mathbbm{1}(\cE^{{\rm c}})  \} \\ 
   & \leq r_n + C_1 \frac{1 + \bar \tau \eta_0 B T}{n^2} +  2 \bar \tau \eta_0 B T^{1/2} \frac{ 1}{\mu n}  \sum_{t=0}^{T-1}  ( \E\| \BFg_t\|_2^2 )^{1/2} P(\cE^{{\rm c}})^{1/2} \\ 
   & \leq r_n + C_1 \frac{1 + \bar \tau \eta_0 B T}{n^2} +  2 \bar \tau \eta_0 \sqrt{C_1 p}    \frac{B T^{3/2}}{\mu n^2}     .
\end{align*}
Taking $B \asymp \sqrt{p+\log n}$ and $T\asymp \log n$, the second and third terms on the right-hand side are dominated by $r_n$ in order. This proves the claimed upper bound under expectation.

To prove the regret bound in Theorem~\ref{thm:regret.bound},  recall that the optimal coefficient $\BFbeta^*$ in the linear demand model satisfies the first-order condition $\nabla Q(\BFbeta^*) = 0$. Under Condition~\ref{cond.density}, the Hessian matrix $\nabla^2 Q(\BFbeta) =  \E \{ f_{\varepsilon | \BFx} ( \BFx^\T (\BFbeta - \BFbeta^*) ) \BFx \BFx^\T \}$ satisfies $\| \Sigma^{-1/2} \nabla^2 Q(\BFbeta)  \Sigma^{-1/2}\|_2 \leq f_u \| \E (\BFw \BFw^\T ) \|_2 = f_u$ for all $\BFbeta \in \RR^p$. Therefore, applying the mean value theorem yields $Q(\BFbeta) -  Q(\BFbeta^*) = Q(\BFbeta) -  Q(\BFbeta^*) - \langle \nabla Q(\BFbeta^*) , \BFbeta - \BFbeta^* \rangle   \leq 0.5 f_u \| \BFbeta - \BFbeta^* \|_\Sigma^2$ for any $\BFbeta \in \RR^p$. This, joint with the property that $\|  \BFbeta^{(T)} - \BFbeta^*\|_\Sigma \lesssim  f_l^{-1}\sqrt{(p\log n) /n} + \sqrt{\log n} \,\eta_0 (p+\log n)/(\mu n)$ holds with high probability, proves the claimed regret bound. 

To bound the expected regret, by \eqref{crude.err.bound} and the assumption $\eta_0\leq 1/\max\{ 2f_u, f_l+\bar \tau\}$ we have 
\begin{align*}
    \| \BFbeta^{(T)} - \BFbeta^* \|_\Sigma^2 &  \leq 2(1+  \bar \tau \eta_0 B T  )^2 +  8 \bar \tau^2 \eta_0^2 B^2 T  \frac{1}{(\mu n)^2 } \bigg( \sum_{t=0}^{T-1} \| \BFg_t \|_2 \bigg)^2 \\
    & \leq 2(1+  0.5 \bar \tau f_u^{-1} B T )^2 + 2\bar \tau^2 \bigg( \frac{BT}{ f_u\mu n} \bigg)^2\sum_{t=0}^{T-1} \| \BFg_t \|_2^2 .
\end{align*}
This, together with the inequality $\E \| \BFg_t \|_2^4 \leq p \sum_{j=1}^p \E g_{t,j}^4 = 3 p^2$, implies
\begin{align*}
 &  \E  \| \BFbeta^{(T)} - \BFbeta^* \|_\Sigma^2  \leq r_n^2 + (1+  0.5 \bar \tau f_u^{-1} B T)^2 \frac{2 C_1}{n^2} +  2 \bar \tau^2 \sqrt{3C_1}\frac{p}{n} \bigg( \frac{BT}{f_u \mu n} \bigg)^2 .
\end{align*}
 Once again, the leading term on the right-hand side is $r_n^2$ (in order). This proves the claimed expected regret bound. \Halmos

\subsection{Proof of Theorem~\ref{thm:convergence}}
 
For simplicity, we write $\cE_0 = \cE_0(B)$, $\cE_1=\cE_1(\delta_0, \delta_1)$ and $\cE_2 = \cE_2(2)$ for the events defined in \eqref{def:E0}--\eqref{def:E2}, where $\delta_0,\delta_1>0$ satisfy the constraints in \eqref{parameters.cond1}.  
 
To control the random perturbations in noisy gradient descent, for any $z>0$ we define 
\#
 \cG = \cG(z) =   \bigg\{   \max_{ 0 \leq t\leq  T-1} \| \BFg_t \|_2 \leq  B_T  \bigg\} \bigcap  \bigg\{ \sum_{t=0}^{T-1} (1-\epsilon)^t  \| \BFg_{T-1-t} \|_2^2 \leq 2p/\epsilon + 3z  \bigg\} \label{def:G}
\#
where $B_T  := \sqrt{p} + \sqrt{2(\log T + z) }$ and $\epsilon :=\eta_0 \phi_1$. It thus follows from Lemma~\ref{lem:max.normal.l2} that $P\{ \cG (z)   \} \geq 1-2 e^{-z}$.
In the following, we prove \eqref{def:r*} by conditioning on these events.
  
Lemma~\ref{lem:GD.compact} shows that starting at an initial estimate $\BFbeta^{(0)} \in \Theta_\Sigma^*(1)$ and conditioning on $\cE_0 \cap \cE_1 \cap \cE_2 \cap \cG$, all the successive iterates will stay in $ \Theta_\Sigma^*(1)$. Next, we establish a contraction property for the noisy gradient descent iterates. Define
\$
     \wt \BFbeta^{(t+1)}  = \BFbeta^{(t)} -  \eta_0 \Sigma^{-1} \nabla  \hat Q_\varpi(\BFbeta^{(t)}) ~~\mbox{ and }~~ \BFh_t =   \frac{\eta_0 \sigma}{ n}\BFg_t , \ \ t = 0, 1, \ldots, T-1,
\$
and note that $\BFbeta^{(t+1)} = \wt \BFbeta^{(t+1)} + \Sigma^{-1/2}  \BFh_t$. Note that \eqref{parameters.cond3} is guaranteed by \eqref{sample.size.requirement}. Lemma~\ref{lem:GD.compact} ensures that $\BFbeta^{(t)} \in \Theta_\Sigma^*(1)$ for all $t= 0 ,1 ,\ldots, T$. Similarly, it can be shown that the non-private gradient descent iterates $\wt \BFbeta^{(t)}$ also stay in the ball $\Theta_\Sigma^*(1)$ for $t=1,\ldots, T$.
For simplicity, set
\$
	\BFdelta^{(t)} = \BFbeta^{(t )} - \BFbeta^* ,  \quad \wt \BFdelta^{(t )} = \wt \BFbeta^{(t)} - \BFbeta^* ~~\mbox{ for }~ t= 0 , 1, \ldots, T.
\$
To upper bound $\| \BFdelta^{(t+1)}   \|_\Sigma $ at each iteration,  we split the analysis into two cases. 

\noindent
{\bf Case 1}:  Suppose the non-private gradient descent iterate $\wt \BFbeta^{(t+1)}$ is in $\Theta_\Sigma(r_0)$. Then
\#
  \| \BFdelta^{(t+1)}   \|_\Sigma \leq r_0 + \| \BFh_t \|_2.  \label{case1}
\#

\noindent
{\bf Case 2}: Assume otherwise $\wt \BFbeta^{(t+1)} \notin \Theta_\Sigma(r_0)$.
For the previously defined $\epsilon =\eta_0 \phi_1  \in (0, 1/2]$,
\#
  \| \BFdelta^{(t+1)}   \|_\Sigma^2  
	& =  \| \wt \BFdelta^{(t+1)}  \|_\Sigma^2 + \| \BFh_t \|_2^2 + 2 \langle \Sigma^{ 1/2} \wt  \BFdelta^{(t+1)}  ,  \BFh_t \rangle \nn \\
	& \leq  ( 1 + \epsilon ) \| \wt \BFdelta^{(t+1)}   \|_\Sigma^2 + (1+1/\epsilon) \| \BFh_t \|_2^2 \nn \\
	& = (1+\epsilon)  \| \BFdelta^{(t)}   \|_\Sigma^2   + (1+1/\epsilon) \| \BFh_t \|_2^2   \label{case2.1} \\
	&~~~~~+ 2\eta_0 (1+\epsilon)     \bigg\{   \underbrace{\frac{\eta_0}{2}  \| \nabla \hat Q_\varpi(\BFbeta^{(t)}  )\|_{\Sigma^{-1}}^2 -  \langle \BFbeta^{(t)} - \BFbeta^* , \nabla \hat Q_\varpi(\BFbeta^{(t)} ) \rangle }_{  \Pi  } \bigg\}  .  \nn
\#
To bound $\Pi$, we use the local strong convexity and smoothness properties of $\hQ_\varpi(\cdot)$ as in \eqref{restricted.rsc.one-side} and \eqref{restricted.smootheness}. The latter implies
\$
 \hat Q_\varpi(\wt \BFbeta^{(t+1 )}) - \hat Q_\varpi(  \BFbeta^{(t)} )  \leq \langle \nabla \hat Q_\varpi( \BFbeta^{(t)} ) , \wt \BFbeta^{(t+1)} - \BFbeta^{(t)} \rangle + f_u \| \wt \BFbeta^{(t+1)} - \BFbeta^{(t)} \|_\Sigma^2 .
\$
For the former, we assume without loss generality that $r_1 = (\delta_0 + \delta_1) /\phi_1  \geq  1/n$. Otherwise if $r_1<1/n$, \eqref{restricted.rsc.one-side} implies $\hQ_\varpi(\BFbeta^*) - \hQ_\varpi(\BFbeta)  -  \langle \nabla \hQ_\varpi(\BFbeta), \BFbeta^* - \BFbeta \rangle  \geq    \phi_1   (  \| \BFbeta - \BFbeta^* \|_{\Sigma}^2 -   n^{-1} \| \BFbeta - \BFbeta^* \|_{\Sigma}  )$ for all $\BFbeta \in \Theta_\Sigma^*(1)\setminus  {\Theta}_\Sigma^*(1/n)$. Then we obtain \eqref{def:r*} with $r_1$ replaced by $1/n$.

Since $r_1  \geq  1/n$, it holds for every $\BFbeta^{(t)}\in \Theta^*_\Sigma(1/n)$ that
\$
 \hat Q_\varpi(\BFbeta^* ) - \hat Q_\varpi(\BFbeta^{(t)}) - \langle \nabla \hat Q_\varpi(\BFbeta^{(t)}) , \BFbeta^* - \BFbeta^{(t)} \rangle \geq 0 \geq  \phi_1 \| \BFbeta^{(t)} - \BFbeta^* \|_\Sigma^2 - (\delta_0+\delta_1) \| \BFbeta^{(t)} - \BFbeta^* \|_\Sigma .
\$
Otherwise if $\BFbeta^{(t)} \in \Theta^*_\Sigma(1)\setminus \Theta^*_\Sigma(1/n)$,
\$
 \hat Q_\varpi(\BFbeta^* ) - \hat Q_\varpi(\BFbeta^{(t)})  -  \langle \nabla \hat Q_\varpi(\BFbeta^{(t)}) , \BFbeta^* - \BFbeta^{(t)} \rangle \geq  \phi_1 \| \BFbeta^{(t)} - \BFbeta^* \|_\Sigma^2 - (\delta_0+\delta_1) \| \BFbeta^{(t)} - \BFbeta^* \|_\Sigma .
\$
 Together, the above upper and lower bounds imply
\$
 &  \hat Q_\varpi(\wt \BFbeta^{(t+1 )})  -  \hat Q_\varpi(\BFbeta^* )  \\  
 & \leq \langle \nabla \hat Q_\varpi(\BFbeta^{(t)} )  , \wt \BFbeta^{(t+1)} - \BFbeta^*   \rangle  + f_u \| \wt \BFbeta^{(t+1)} - \BFbeta^{(t)} \|_\Sigma^2 -  \phi_1 \| \BFbeta^{(t)} - \BFbeta^* \|_\Sigma^2  + (\delta_0+\delta_1) \| \BFbeta^{(t)} - \BFbeta^* \|_\Sigma \\
 & = \langle  \nabla \hat Q_\varpi(\BFbeta^{(t)} )  , \wt \BFbeta^{(t+1)} - \BFbeta^*   \rangle  + f_u \eta_0^2  \|  \nabla   \hat Q_\varpi(\BFbeta^{(t)} ) \|_{ \Sigma^{-1}}^2 -  \phi_1 \| \BFbeta^{(t)} - \BFbeta^* \|_\Sigma^2  + (\delta_0+\delta_1) \| \BFbeta^{(t)} - \BFbeta^* \|_\Sigma \\
  & = \langle  \nabla  \hat Q_\varpi(\BFbeta^{(t)} )  ,  \BFbeta^{(t)} - \BFbeta^*   \rangle   - \eta_0 ( 1 - f_u \eta_0)    \|   \nabla   \hat Q_\varpi(\BFbeta^{(t)} ) \|_{ \Sigma^{-1}}^2    -  \phi_1 \| \BFbeta^{(t)} - \BFbeta^* \|_\Sigma^2  + (\delta_0+\delta_1) \| \BFbeta^{(t)} - \BFbeta^* \|_\Sigma  \\
  & \leq \langle  \nabla\hat Q_\varpi(\BFbeta^{(t)} )  ,  \BFbeta^{(t)} - \BFbeta^*   \rangle   -   \frac{ \eta_0}{2}   \| \nabla    \hat Q_\varpi(\BFbeta^{(t)} ) \|_{ \Sigma^{-1}}^2    -  \phi_1 \| \BFdelta^{(t)}   \|_\Sigma^2  + (\delta_0+\delta_1) \| \BFdelta^{(t)}   \|_\Sigma ,
\$
where the last inequality holds when $\eta_0 \leq 1/(2f_u)$. Since $\wt \BFbeta^{(t+1)} \in \Theta^*_\Sigma(1) \setminus \Theta^*_\Sigma(r_0)$, the restricted lower bound \eqref{restricted.loss.difference} yields
\$
\hat Q_\varpi(\wt \BFbeta^{(t+1 )})  -  \hat Q_\varpi(\BFbeta^* )  \geq   \phi_1  \| \wt \BFbeta^{(t+1 )} - \BFbeta^* \|_\Sigma \cdot  \big(    \| \wt \BFbeta^{(t+1 )} - \BFbeta^* \|_\Sigma - r_0 \big) \geq 0.
\$
We hence conclude that $\Pi \leq   -  \phi_1 \| \BFdelta^{(t)}  \|_\Sigma^2  + (\delta_0+\delta_1) \| \BFdelta^{(t)}   \|_\Sigma$. Substituting this into \eqref{case2.1} implies
\#
 & \| \BFdelta^{(t+1)}   \|_\Sigma^2 \nn \\
& \leq   \big\{  1 + \epsilon - 2\eta_0 \phi_1  (1+\epsilon)  \big\} \| \BFdelta^{(t)}   \|_\Sigma^2 +  (  1 + 1/\epsilon  ) \| \BFh_t \|_2^2 +  2\eta_0   (1+\epsilon) (\delta_0+\delta_1) \| \BFdelta^{(t)}  \|_\Sigma \nn \\
&\leq   \big\{  1 +  \epsilon - 2\eta_0 \phi_1  (1+\epsilon)  \big\} \| \BFdelta^{(t)}  \|_\Sigma^2 +  (  1 + 1/\epsilon  ) \| \BFh_t \|_2^2  \nn \\
& ~~~~~~~~~~~~~~~~~~~~~~~~~~~~~~~~~~~~~~~~~~~~~~~~~+  (1+\epsilon) (\eta_0 \phi_1)^2 \| \BFdelta^{(t)}   \|_\Sigma^2  + (1+\epsilon)(\delta_0 + \delta_1)^2  /\phi_1^2  \nn \\
& = (1+\epsilon) (1-\eta_0 \phi_1 )^2 \| \BFdelta^{(t)}   \|_\Sigma^2 +  (  1 + 1/\epsilon  ) \| \BFh_t \|_2^2+ (1+\epsilon) r_1^2 . \label{case2}
\#
On the other hand, observe that 
$$
(r_0 + \| \BFh_t \|_2   )^2 = r_0^2 + \| \BFh_t\|_2^2 + 2 r_0 \| \BFh_t \|_2 \leq  (  1 + 1/\epsilon  ) \| \BFh_t \|_2^2 +  (1+\epsilon) r_0^2 .
$$
Recall that $\epsilon = \eta_0   \phi_1 \in (0, 1/2]$. It then follows from \eqref{case1} and \eqref{case2} that
\#
  \| \BFdelta^{(t+1)}   \|_\Sigma^2 \leq  (1-\epsilon)   \| \BFdelta^{(t)}  \|_\Sigma^2 +   (  1 + 1/\epsilon ) \| \BFh_t \|_2^2 +  (1+\epsilon ) (r_0 \vee r_1)^2, \nn
\#
for $t=0,1, \ldots , T-1$. This recursive bound further implies
\#
\| \BFbeta^{(T)} - \BFbeta^* \|_\Sigma^2  & \leq  (1-\epsilon )^T  +(1+1/\epsilon )   \bigg\{   \Big( \frac{\eta_0 \sigma}{n} \Big)^2  \underbrace{ \sum_{t=0}^{T-1} (1-\epsilon )^t \| \BFg_{T-1-t} \|_2^2 }_{\leq \, 2p/\epsilon + 3z ~{\rm on }~ \cG} + \, (r_0 \vee r_1)^2 \bigg\} . \label{convergence.rate1}
\#

Let $T = T_1 + T_2$  with $T_1  \geq  2 \log (n) / \log ((1-\epsilon)^{-1})$, and $r^*$ be as in \eqref{def:r*}.
Then,  conditioned on $\cE_0 \cap \cE_1 \cap \cE_2 \cap \cG$, we obtain
$\| \BFbeta^{(t)} - \BFbeta^* \|_\Sigma    \leq   r^*$ for all $ T_1 \leq t \leq T$, as desired. In other words, $r^*$ given in \eqref{def:r*} characterizes the convergence rate of the noisy gradient descent iterators $\BFbeta^{(t)}$ to the true parameter $\BFbeta^*$.

Turning to the non-private empirical risk minimizer $\hat \BFbeta_\varpi \in \argmin_{\bbeta \in \mathbb{R}^p} \hat Q_\varpi(\BFbeta)$, we claim that $\| \hat \BFbeta_\varpi - \BFbeta^* \|_\Sigma \leq r_0 = (\delta_0+b^*)/\phi_1$ conditioning on $\mathcal{E}_1$. To see this, note that for any $\BFdelta \in \Theta_\Sigma(1)$, 
$$
    \hat D_{\varpi} (\BFdelta) = R_\varpi(\BFdelta) + \langle \nabla Q_\varpi(\BFbeta^*) , \BFdelta \rangle + \{   \hat D_\varpi(\BFdelta) -    D_\varpi(\BFdelta) \} \geq \phi_1 \| \BFdelta \|_\Sigma^2 - (\delta_0 + b^*) \| \BFdelta \|_\Sigma .
$$
By \eqref{parameters.cond1}, $ \hat D_{\varpi} (\BFdelta) \geq \phi_1 - \delta_0 - b^* >0$ for all $\BFdelta \in \partial \Theta_\Sigma(1)$--the boundary of $\Theta_\Sigma(1)$. On the other hand, note that $\hat \BFdelta = \hat \BFbeta_\varpi - \BFbeta^*$ satisfies $\hat D_{\varpi} (\hat \BFdelta_\varpi) \leq 0$. By the convexity of $\hat Q_\varpi(\cdot)$, we must have $\hat \BFbeta_\varpi \in \Theta_\Sigma^*(1)$. Otherwise if $\| \hat \BFbeta_\varpi - \BFbeta^* \|_\Sigma >1$, there exists some $\alpha \in (0, 1)$ such that $\tilde \BFbeta = \alpha \hat \BFbeta_\varpi + (1-\alpha) \BFbeta^*$ lies on the boundary of $\Theta_\Sigma^*(1)$, that is, $\| \tilde \BFbeta - \BFbeta^* \|_\Sigma =1$. Then $\tilde \BFdelta := \tilde \BFbeta - \BFbeta^*$ satisfies $ \hat D_{\varpi}(\tilde \BFdelta) >0$. However, by the convexity of $\hat D_\varpi$,
$$
\hat D_{\varpi}(\tilde \BFdelta) = \hat D_{\varpi}(\alpha \hat  \BFdelta_\varpi + (1-\alpha) \BFzero) \leq \alpha \hat D_{\varpi}(  \hat  \BFdelta ) + (1-\alpha) \hat D_{\varpi}(\BFzero)  \leq 0.
$$
This leads to a contradiction. As a result, from \eqref{restricted.loss.difference} it can be deduced that
$$
   0 \geq   \hat D_\varpi(\hat \BFdelta) \geq \phi_1 \| \hat \BFdelta \|_\Sigma^2 - (\delta_0 + b^*) \| \hat \BFdelta \|_\Sigma ,
$$
 thereby verifying the claim.

Finally, we prove the convergence of $\BFbeta^{(t)}$ towards the non-private estimator $\hat \BFbeta_\varpi$, for which it has been shown that $\| \hat \BFbeta_\varpi - \BFbeta^* \|_\Sigma\leq r_0$ conditioned on $\mathcal{E}_1$. By the triangle inequality, $\| \BFbeta^{(t)} - \hat \BFbeta_\varpi \|_\Sigma \leq \| \BFbeta^{(t)} - \BFbeta^* \|_\Sigma + \| \hat \BFbeta_\varpi - \BFbeta^* \|_\Sigma \leq r^*+ r_0$ for all $T_1\leq t\leq T$ conditioned on $\mathcal{E}_0 \cap \mathcal{E}_1 \cap \mathcal{E}_2 \cap \mathcal{G}$. To improve this ``crude'' bound, the key element is the refined restricted strong convexity property, guaranteed by the event $\mathcal{E}_3 = \mathcal{E}_3(r, \phi_2)$ with $r= r^* + r_0$.

Thus far we have shown that $\{ \tilde{\BFbeta}^{(t)} \}_{t=0, 1, \ldots, T} \subseteq \Theta^*_\Sigma(1)$ and $\{ \BFbeta^{(t)} \}_{t=T_1, \ldots, T } \subseteq \Theta^*_\Sigma(r^*)$. Moreover, note that inequality \eqref{case2.1} remains valid if $\BFbeta^*$ is replaced by $\hat \BFbeta_\varpi$, that is, 
\begin{align*}
    \| \BFbeta^{(t+1)} - \hat \BFbeta_\varpi \|_\Sigma^2   & \leq (1+\epsilon) \| \BFbeta^{(t)} - \hat \BFbeta_\varpi \|_\Sigma^2 + (1+1/\epsilon) \| \hb_t \|_2^2  \\
     & ~~~~ + 2 \eta_0 ( 1+\epsilon) \bigg\{ \frac{\eta_0}{2} \| \nabla \hat Q_{\varpi}(\BFbeta^{(t)} ) \|_{\Sigma^{-1}}^2 - \langle \BFbeta^{(t)} - \hat{\BFbeta}_\varpi , \nabla \hat{Q}_{\varpi}(\BFbeta^{(t)} ) \rangle \bigg\} .
\end{align*}
Conditioned on $\mathcal{E}_3(r,\phi_2)$ with $r=r^*+r_0 \geq 2r_0$, we have, for $t=T_1, T_1+1,\ldots, T-1$,
\begin{align*}
    \hat Q_{\varpi}(\tilde \BFbeta^{(t+1)} ) = \hat Q_\varpi (\BFbeta^{(t)} \leq \langle \nabla \hat Q_\varpi (\BFbeta^{(t)}) , \tilde \BFbeta^{(t+1)} - \BFbeta^{(t)} \rangle + f_u \| \tilde\BFbeta^{(t+1)} - \BFbeta^{(t)} \|_\Sigma^2  \\
    \mbox{ and }~ \hat Q_\varpi(\hat \BFbeta_\varpi) - \hat Q_\varpi(\BFbeta^{(t)} ) \geq \langle \nabla \hat Q_\varpi(\BFbeta^{(t)} ) , \hat \BFbeta_\varpi - \BFbeta^{(t)} \rangle + \phi_2 \| \BFbeta^{(t)} - \hat\BFbeta_\varpi \|_\Sigma^2 .
\end{align*}
Following a similar argument, and recall that $\hat \BFbeta_\varpi$ minimizes $\hat Q_\varpi(\cdot)$, we obtain
\begin{align*}
    0 & \leq \hat Q_\varpi (\tilde \BFbeta^{(t+1)} - \hat Q_\varpi(\hat \BFbeta_\varpi) \\
    & \leq \langle \nabla \hat Q_\varpi (\BFbeta^{(t)} ) , \BFbeta^{(t)} - \hat \BFbeta_\varpi \rangle - \frac{\eta_0}{2} \| \nabla \hat Q_\varpi(\BFbeta^{(t)}
 \|_{\Sigma^{-1}}^2 - \phi_2 \| \BFbeta^{(t)} - \hat \BFbeta_\varpi \|_\Sigma^2 .
 \end{align*}
Putting together the pieces yields the recursive bound 
$$
 \| \BFbeta^{(t+1)} - \hat \BFbeta_\varpi \|_\Sigma^2 \leq (1-\epsilon) \| \BFbeta^{(t)} - \hat \BFbeta_\varpi \|_\Sigma^2 + (1+1/\epsilon)  \| \hb_t \|_2^2, \quad t=T_1, T_1+1, \ldots, T. 
$$
Provided $T_2 \geq \log(n)/\log((1-\epsilon)^{-1})$, conditioning on $\mathcal{E}_0\cap \mathcal{E}_1\cap \mathcal{E}_2 \cap \mathcal{E}_3 \cap \mathcal{G}$ it follows that
\begin{align*}
    \| \BFbeta^{(T)} - \hat \BFbeta_\varpi \|_\Sigma^2 & \leq (1-\epsilon)^{T_2} \| \BFbeta^{(T_1)} -\hat \BFbeta_\varpi \|_\Sigma^2 + (1+1/\epsilon) \bigg( \frac{\eta_0\sigma}{n} \bigg)^2 \sum_{t=0}^{T_2-1} (1-\epsilon)^t \| \gb_{T-1-t} \|_2^2  \\
    & \leq \frac{r^2}{n} + (1+1/\epsilon) (2p/\epsilon + 3z) \bigg(\frac{\eta_0 \sigma}{n} \bigg)^2 ,
\end{align*}
as claimed. This completes the proof of the theorem. \Halmos

\subsection{Proof of Theorem~\ref{thm:initial.convergence}}


To begin with, recall from the proof of Theorem~\ref{thm:convergence} that conditioning on $\cE_0(B) \cap \cE_1(\delta_0, \delta_1) \cap \cE_2(2)$, the non-private empirical risk minimizer $\hat \BFbeta_\varpi$ satisfies $\| \hat \BFbeta_\varpi - \BFbeta^* \|_\Sigma \leq r_0 = (\delta_0 + b^*) / \phi_1 < 1$.  For any $T_0\geq 2$ and $z\geq 0$,  set
\$
\cG_0(z) = \bigg\{ \max_{0\leq t \leq T_0 - 1 } \| \BFg_t \|_2 \leq  B_{T_0} : = \sqrt{p} + \sqrt{2(\log T_0 + z ) } \bigg\}, 
\$
so that $P \{ \cG_0(z) \} \geq  1- e^{-z}$. From Lemma~\ref{lem:initial.value} we see that conditioned on $\cE_0(B) \cap \cE_2(R) \cap \cG_0(z)$ with $R= 2R_0 + r_0$,  the iterates $\{ \BFbeta^{(t)} \}_{t=1,\ldots, T_0}$ satisfy
\#
  \| \BFbeta^{(t)} - \hat \BFbeta_\varpi  \|_\Sigma^2 \leq e R_0^2 + (e-1) (2  \bar \tau B + 1/2)   \frac{ T_0 B_{T_0} \sigma }{n} \eta_0^2  \leq  4 R_0^2   \label{GD.crude.bound}
\#
as long as $n\geq    \frac{e-1}{4-e}  ( 2  \bar \tau B + 1/2 )   T_0  B_{T_0} \sigma$.
Consequently, $\| \BFbeta^{(t)} - \BFbeta^* \|_\Sigma \leq \| \BFbeta^{(t)} - \hat \BFbeta_\varpi \|_\Sigma + \|  \hat \BFbeta_\varpi - \BFbeta^* \|_\Sigma \leq R$, and hence  $\BFbeta^{(t)} \in \Theta^*_\Sigma(R)$ for all $t=0, 1, \ldots, T_0$.

Keep the notation from the proof of Theorem~\ref{thm:convergence}, and note that $\BFbeta^{(t+1)} = \BFbeta^{(t)} - \eta_0  \Sigma^{-1} \nabla \hQ_\varpi(\BFbeta^{(t)} )  - \Sigma^{-1/2} \BFh_t $, where $\BFh_t = \frac{ \eta_0  \sigma}{n} \BFg_t$ and $\eta_0 \leq 1/(2f_u)$. Conditioned on $\cG_0(z)$, 
\#
	 \max_{0\leq t\leq T_0 -1 } \| \BFh_t \|_2 \leq e_{{\rm priv}} := \frac{ \eta_0  B_{T_0} \sigma}{n}  < \frac{ \eta_0}{2}. \label{ht.uniform.bound}
\# 
By \eqref{restricted.smootheness} and the convexity of $\hQ_\varpi(\cdot)$, for each $t = 0 , 1, \ldots, T_0 -1$ we have
\#
	\hQ_\varpi(\BFbeta^{(t+1)} ) &   \leq \hQ_\varpi(\BFbeta^{(t)} )   + \langle \nabla \hQ_\varpi(\BFbeta^{(t)}) , \BFbeta^{(t+1)} - \BFbeta^{(t)} \rangle + f_u  \| \BFbeta^{(t+1)} - \BFbeta^{(t)} \|_\Sigma^2  \nn \\
&  = \hQ_\varpi(\BFbeta^{(t)} )   -  \langle \nabla \hQ_\varpi(\BFbeta^{(t)}) ,   \eta_0 \Sigma^{-1} \nabla \hQ_\varpi(\BFbeta^{(t)} )  + \Sigma^{-1/2} \BFh_t \rangle  \nn \\ 
& ~~~~+ f_u \|   \eta_0  \Sigma^{-1} \nabla \hQ_\varpi(\BFbeta^{(t)} )  + \Sigma^{-1/2} \BFh_t \|_\Sigma^2 \nn \\
& \leq \hQ_\varpi(\BFbeta^{(t)} )  - \eta_0 ( 1- f_u \eta_0 )  \|  \nabla \hQ_\varpi(\BFbeta^{(t)} ) \|_{\Sigma^{-1}}^2   \nn \\
& ~~~~   +  ( 1 + 2 f_u \eta_0  ) \| \BFh_t \|_2  \|  \nabla \hQ_\varpi(\BFbeta^{(t)} ) \|_{\Sigma^{-1}} + f_u    \| \BFh_t \|_2^2 \nn \\ 
   & \leq \hQ_\varpi(\BFbeta^{(t)} )  - \frac{\eta_0}{2 }  \|  \nabla \hQ_\varpi(\BFbeta^{(t)} ) \|_{\Sigma^{-1}}^2 +    (2 \bar \tau B +  f_u e_{{\rm priv}}  ) e_{{\rm priv}} \label{loss.different.bound} \\
   & \leq  \hQ_\varpi( \hat \BFbeta_\varpi )  +  \underbrace{  \langle \nabla \hQ_\varpi(\BFbeta^{(t)} ) , \BFbeta^{(t)}  - \hat \BFbeta_\varpi \rangle - \frac{\eta_0}{2 }  \|  \nabla \hQ_\varpi(\BFbeta^{(t)} ) \|_{\Sigma^{-1}}^2  }_{=: \, \Pi_t}+   (2 \bar \tau B + f_u e_{{\rm priv}}   ) e_{{\rm priv}} . \nn
\#
Moreover,
\$
	\Pi_t  & = \frac{1}{2\eta_0 }  \Big\{   2\eta_0 \langle \nabla \hQ_\varpi(\BFbeta^{(t)} ) , \BFbeta^{(t)}  - \hat \BFbeta_\varpi \rangle -  \eta_0^2   \|  \nabla \hQ_\varpi(\BFbeta^{(t)} ) \|_{\Sigma^{-1}}^2 \Big\} \\
	& =   \frac{1}{2\eta_0 }   \Big\{     \|  \BFbeta^{(t)}  - \hat \BFbeta_\varpi \,  \|_\Sigma^2  -  \|  \BFbeta^{(t)} - \eta_0  \Sigma^{-1} \nabla  \hQ_\varpi(\BFbeta^{(t)} ) - \hat \BFbeta \, \|_{\Sigma}^2 \Big\} \\
	& =  \frac{1}{2\eta_0 }   \Big\{     \|  \BFbeta^{(t)}  - \hat \BFbeta_\varpi   \|_\Sigma^2  -  \|  \BFbeta^{(t+1)} + \Sigma^{-1/2} \BFh_t  - \hat \BFbeta_\varpi   \|_{\Sigma}^2 \Big\} \\
	& =  \frac{1}{2\eta_0 }   \Big\{     \|  \BFbeta^{(t)}  - \hat \BFbeta_\varpi  \|_\Sigma^2  - \|  \BFbeta^{(t+1)}   - \hat \BFbeta_\varpi \|_{\Sigma}^2 -   \|    \BFh_t  \|_2^2   - 2 \langle \BFh_t , \Sigma^{1/2} ( \BFbeta^{(t+1)}  - \hat \BFbeta_\varpi )  \rangle \Big\} \\
	& \leq   \frac{1}{2\eta_0 }   \Big\{     \|  \BFbeta^{(t)}  - \hat \BFbeta_\varpi  \|_\Sigma^2  - \|  \BFbeta^{(t+1)}   - \hat \BFbeta_\varpi  \|_{\Sigma}^2 \Big\} +  \frac{1}{\eta_0}  \| \BFh_t  \|_2  \| \BFbeta^{(t+1)}   - \hat \BFbeta_\varpi \|_\Sigma .
\$
Summing over $t=0, 1, \ldots, T_0-1$ gives
\$
	&	\sum_{t=0}^{T_0-1} \big\{  \hQ_\varpi(\BFbeta^{(t+1)} ) - \hQ_\varpi( \hat \BFbeta_\varpi  ) \big\}  \\
	& \leq  \frac{1}{2\eta_0 }\Big\{     \|  \BFbeta^{(0)}  - \hat \BFbeta_\varpi  \|_\Sigma^2  - \|  \BFbeta^{(T_0)}   - \hat \BFbeta_\varpi\|_{\Sigma}^2 \Big\} \nn \\ 
 & ~~~~ + \frac{1}{\eta_0}\sum_{t=0}^{T_0-1}  \| \BFbeta^{(t+1)}   - \hat \BFbeta_\varpi \|_\Sigma \cdot  \| \BFh_t \|_2 +  T_0  (2 \bar \tau B + f_u  e_{{\rm priv}}  )   e_{{\rm priv}} ,
\$
On the other hand, applying \eqref{loss.different.bound} repeatedly yields
\$
 T_0  \hQ_\varpi(\BFbeta^{(T_0)} )   \leq \sum_{t=1}^{T_0 }  \hQ_\varpi(\BFbeta^{(t )} )   + \frac{T_0(T_0-1)}{2}  (2 \bar \tau B +  f_u e_{{\rm priv}}  ) e_{{\rm priv}} . 
\$
Combining the last two bounds with \eqref{GD.crude.bound} and \eqref{ht.uniform.bound} gives that
\$
\hQ_\varpi(\BFbeta^{(T_0)} )  - \hQ_\varpi( \hat \BFbeta_\varpi ) & \leq \frac{1}{T_0}    \sum_{t=1}^{T_0 }  \big\{ \hQ_\varpi(\BFbeta^{(t )} ) -  \hQ_\varpi( \hat \BFbeta_\varpi ) \big\} +  \frac{ T_0-1 }{2}  (2 \bar \tau B +  f_u e_{{\rm priv}}  ) e_{{\rm priv}} \\
& \leq \frac{1}{2  T_0\eta_0 } \| \BFbeta^{(0)} - \hat \BFbeta_\varpi \|_\Sigma^2 +  \frac{ 2R_0 e_{{\rm priv}} }{\eta_0 }    + (T_0+1) (  \bar \tau B + f_u  e_{{\rm priv}} /2 )  e_{{\rm priv}} \\
& < \frac{R_0^2}{2  T_0 \eta_0}   +    \frac{2 R_0 B_{T_0} \sigma }{n}  + (  \bar \tau B + 1/4 )  (T_0+1) \eta_0 \frac{ B_{T_0} \sigma }{n} .
\$
Under condition \eqref{T0.n.requirement} on $(T_0, n)$, it follows that
\$
\hQ_\varpi(\BFbeta^{(T_0)} ) <   \hQ_\varpi( \hat \BFbeta_\varpi ) + \Delta \leq \hQ_\varpi(\BFbeta^* ) + \Delta  .
\$ 

Finally,  define $\hat \Delta_1 = \inf_{\BFbeta \notin \Theta^*_\Sigma(1) } \hQ_\varpi(\BFbeta) - \hQ_\varpi(\BFbeta^* ) $, so that any $\BFbeta$ such that $\hQ_\varpi(\BFbeta) < \hQ_\varpi(\BFbeta^*) + \hat \Delta_1$ must satisfy $\| \BFbeta - \BFbeta^* \|_\Sigma \leq 1$.
Recall that $\hat \BFbeta_\varpi = \argmin_{\BFbeta \in \RR^p} \hQ_\varpi(\BFbeta) \in \Theta^*_\Sigma(r_0)\subseteq \Theta^*_\Sigma(1)$. By the convexity of $\hQ_\varpi(\cdot)$, the infimum over $\RR^p\setminus  \Theta^*_\Sigma(1)$ must be achieved on the boundary $\partial \Theta^*_\Sigma(1)$.
On the other hand, Proposition~\eqref{prop:smooth.landscape} ensures that conditioned on $\cE_1(\delta_0, \delta_1)$,  $\hQ_\varpi(\BFbeta  )  - \hQ_\varpi(   \BFbeta^* )  \geq \Delta = \phi_1 - \delta_0 - b^*$ for any $\BFbeta \in \partial \Theta^*_\Sigma(1)$, thus implying 
$$
	\hat \Delta_1 \geq \Delta > \hQ_\varpi(\BFbeta^{(T_0)} )  -  \hQ_\varpi(\BFbeta^* )  . 
$$
Therefore, we must have $\| \BFbeta^{(T_0)} - \BFbeta^* \|_\Sigma \leq 1$, as claimed. \Halmos

\section{Proof of Propositions}

\subsection{Proof of Proposition~\ref{prop:smooth.landscape}}

\noindent
{\sc Proof of \eqref{restricted.loss.difference}}. For $\BFbeta  \in \RR^p$, write $\BFdelta = \BFbeta - \BFbeta^*$, and note that $\hat D_\varpi(\cdot)$ given in \eqref{def:Dh} satisfies
\#
 \hat D_\varpi(\BFdelta )   &  =  \langle \nabla Q_\varpi(\BFbeta^*), \BFdelta \rangle + R_\varpi(\BFdelta) - \{ D_\varpi(\BFdelta) -\hat D_\varpi(\BFdelta) \} \nn \\
& \geq  R_\varpi(\BFdelta) - \|  \nabla   Q_\varpi(\BFbeta^*) \|_{\Sigma^{-1}} \cdot   \|  \BFdelta \|_{\Sigma} -   \{ D_\varpi(\BFdelta) - \hat D_\varpi(\BFdelta )   \}  \nn \\
& =  R_\varpi(\BFdelta) -  b^*   \|  \BFdelta \|_{\Sigma} -   \{ D_\varpi(\BFdelta) - \hat D_\varpi(\BFdelta )   \} . \label{loss.diff.lower.bound}
\#
By Lemma~\ref{lem:population.loss.curvature}, $R_\varpi(\BFdelta) \geq \phi_1 \| \BFdelta \|_\Sigma^2$ for all $\BFdelta \in \Theta_\Sigma(1)$; and conditioned on $\cE_1$, $D_\varpi(\BFdelta) - \hat D_\varpi(\BFdelta )  \leq \delta_0 \| \BFdelta \|_\Sigma$ for all $\delta \in \Theta_\Sigma(1) \setminus \Theta_\Sigma(1/n)$. Substituting these two bounds into \eqref{loss.diff.lower.bound} proves the first part of \eqref{restricted.loss.difference}.

To prove the second part of \eqref{restricted.loss.difference}, consider the decomposition
\#
  \hat  Q_\varpi( \BFbeta ) - \hat Q_\varpi(\BFbeta^*)   =   \hat R_\varpi(\BFdelta ) +   \langle \nabla \hat Q_\varpi (\BFbeta^*), \BFbeta- \BFbeta^* \rangle  , \nn
\#
where $\hat R_\varpi(\BFdelta) = \hat D_\varpi(\BFdelta) - \langle \nabla \hQ_\varpi(\BFbeta^* ), \BFdelta \rangle$.  By the mean value theorem,
\$
  \hat R_\varpi(\BFdelta)   =  \int_0^1 \langle  \nabla \hQ_\varpi(\BFbeta^* + t \BFdelta ) - \nabla \hQ_\varpi(\BFbeta^*) , \BFdelta \rangle {\rm d}t .
\$
By Lemma~8 in \cite{LW2015} and  the convexity of $\hQ_\varpi(\cdot)$, we have
\$
  \langle \nabla \hQ_\varpi(\BFbeta^* + t \BFdelta ) - \nabla \hQ_\varpi(\BFbeta^*) ,  t \BFdelta \rangle   
 \geq \frac{1}{s}   \langle  \nabla \hQ_\varpi(\BFbeta^* +  t \cdot s \BFdelta ) - \nabla \hQ_\varpi(\BFbeta^*) ,  t \cdot   s \BFdelta \rangle  
\$
for any $s\in (0,1]$, hence implying $ \hat R_\varpi(\BFdelta)  \geq      s^{-1}  \hat R_\varpi(s \BFdelta )$.  Consequently,  for any $\BFbeta  \in \Theta^*_\Sigma(1)^{{\rm c}}$,  we set $s= 1/\| \BFdelta \|_{\Sigma}\in (0,1)$, $\BFdelta_1 = \BFdelta / \| \BFdelta \|_{\Sigma} \in \partial \Theta_\Sigma(1)$, and obtain that
\$
 \hQ_\varpi(\BFbeta) - \hQ_\varpi(\BFbeta^*) &  \geq   \| \BFdelta \|_{\Sigma} \cdot   \hat R_\varpi(\BFdelta_1)  +  \| \BFdelta \|_{\Sigma} \cdot   \langle \nabla  \hQ_\varpi(\BFbeta^*),  \BFdelta_1 \rangle   \\
 &  =  \| \BFdelta \|_{\Sigma} \cdot  \big\{ \hQ_\varpi(\BFbeta^* +\BFdelta_1 ) - \hQ_\varpi(\BFbeta^*) \big\} .
\$
Combining this with the earlier lower bound when $\BFbeta \in \Theta^*_\Sigma(1)$ proves the second part of \eqref{restricted.loss.difference}.

\medskip
\noindent
{\sc Proof of \eqref{restricted.rsc.one-side}}.  The goal is to bound  $\hat B_\varpi( \BFdelta)$ defined in \eqref{def:Bh} from below uniformly over $\BFdelta$ in a compact set. Note that
 \$
\hat  B_\varpi(\BFdelta)  &  =   - \hat D_\varpi(\BFdelta) + \langle  \nabla Q_\varpi(\BFbeta) , \BFdelta \rangle+ \langle \nabla \hat Q_\varpi(\BFbeta) - \nabla Q_\varpi(\BFbeta) , \BFdelta \rangle  \nn \\
& =  \underbrace{  - D_\varpi(\BFdelta) + \langle  \nabla Q_\varpi(\BFbeta) , \BFdelta \rangle }_{=  B_\varpi(\BFdelta) }  - \{ \hat D_\varpi(\BFdelta) - D_\varpi(\BFdelta) \} + \langle \nabla \hat Q_\varpi(\BFbeta) - \nabla Q_\varpi(\BFbeta) , \BFdelta \rangle   \nn \\
& \geq B_\varpi(\BFdelta) - \{   \hat D_\varpi(\BFdelta )   - D_\varpi(\BFdelta)  \} -  \|  \nabla \hat Q_\varpi(\BFbeta) - \nabla Q_\varpi(\BFbeta)    \|_{\Sigma^{-1}} \cdot \|\BFdelta \|_{\Sigma}.
\$
Again, from Lemma~\ref{lem:population.loss.curvature} we see that $B_\varpi(\BFdelta) \geq \phi_1 \| \BFdelta \|_\Sigma^2$ for all $\BFdelta \in \Theta_\Sigma(1)$. Conditioned on $\cE_1=\cE_1(\delta_0,\delta_1)$, the following bounds
\$
  \hat D_\varpi(\BFdelta )   - D_\varpi(\BFdelta)   \leq \delta_0 \| \BFdelta \|_\Sigma~~\mbox{ and }~~
 \|  \nabla \hat Q_\varpi(\BFbeta) - \nabla Q_\varpi(\BFbeta)    \|_{\Sigma^{-1}}  \leq  \delta_1
\$
hold uniformly over $\BFdelta \in \Theta_\Sigma(1) \setminus \Theta_\Sigma(1/n)$ and $\BFbeta \in \Theta_\Sigma^*(1)$, respectively. Putting together the pieces yields \eqref{restricted.rsc.one-side}.

\medskip
\noindent
{\sc Proof of \eqref{restricted.smootheness}}. 
For $\BFbeta_1, \BFbeta_2\in \Theta^*_\Sigma(2)$, applying a second-order Taylor series expansion yields
\$
\hQ_\varpi(\BFbeta_2 ) - \hQ_\varpi(\BFbeta_1) - \langle \nabla \hQ_\varpi(\BFbeta_1) , \BFbeta_2 - \BFbeta_1 \rangle  = \frac{1}{2} (\BFbeta_2 - \BFbeta_1)^\T \nabla^2 \hQ_\varpi(t\BFbeta_1 + (1-t)\BFbeta_2 ) (\BFbeta_2 - \BFbeta_1) 
\$
for some $t \in [0,1]$. Since $t\BFbeta_1 + (1-t)\BFbeta_2 \in  \Theta^*_\Sigma(2)$, the right-hand side is further bounded by
\$
 \frac{1}{2} \| \BFbeta_2 -\BFbeta_1 \|_{\Sigma}^2 \cdot   \sup_{\BFbeta \in  \Theta^*_\Sigma(2) } \| \Sigma^{-1/2} \nabla^2 \hQ_\varpi( \BFbeta ) \Sigma^{-1/2}  \|_2 = \frac{1}{2} \| \BFbeta_2 -\BFbeta_1 \|_{\Sigma}^2 \cdot   \sup_{\BFbeta \in  \Theta^*_\Sigma(2) } \|  \nabla^2 \hQ_\varpi( \BFbeta )   \|_{\Sigma^{-1}}  .
\$
By the triangle inequality,
\$
    \sup_{\BFbeta \in  \Theta^*_\Sigma(2) } \|  \nabla^2 \hQ_\varpi( \BFbeta )   \|_{\Sigma^{-1}}  \leq  \underbrace{    \sup_{\BFbeta \in  \Theta^*_\Sigma(2) }  \|  \nabla^2 \hQ_\varpi( \BFbeta )  - \nabla^2 Q_\varpi( \BFbeta ) \|_{\Sigma^{-1}}   }_{\leq ~f_u {\rm~conditioned~on~} \cE_2(2) } ~+ ~   \sup_{\BFbeta \in  \Theta^*_\Sigma(2) } \|  \nabla^2 Q_\varpi( \BFbeta )   \|_{\Sigma^{-1}}  .
\$
For the population Hessian, note that  $\Sigma^{-1/2}  \nabla^2 Q_\varpi( \BFbeta ) \Sigma^{-1/2}    = \E   \{ K_\varpi(\varepsilon - \BFx^\T \BFdelta ) \BFw \BFw^\T  \}$, where $\BFdelta =  \BFbeta - \BFbeta^*$ and $\E    \{ K_\varpi(\varepsilon -  \BFx^\T \BFdelta ) | \BFx  \} = \int_{-\infty}^\infty K(v) f_{\varepsilon|\BFx}(  \BFx^\T \BFdelta + h v ) {\rm d} v$. Hence, it follows from Condition~\ref{cond.density} that $\| \nabla^2 Q_\varpi( \BFbeta )   \|_{\Sigma^{-1}}\leq f_u$ for all $\BFbeta \in \RR^p$. Combining this with the above two bounds proves \eqref{restricted.smootheness}.  \Halmos

\subsection{Proof of Proposition~\ref{prop:event012}}

The claimed bounds follow immediately from Lemmas~\ref{lem:loss.diff} and \ref{lem:grad.diff}. \Halmos

\subsection{Proof of Proposition~\ref{prop5}}

For each pair $(\BFbeta_1, \BFbeta_2)$ of parameters, it follows from a second-order Taylor series expansion that
\begin{align*}
	 &  \hat{Q}_\varpi( \bbeta_1 )  - \hat{Q}_\varpi( \bbeta_2  )   - \langle \nabla \hat{Q}_\varpi(\bbeta_2) , \bbeta_1 - \bbeta_2 \rangle   \\
& =   \frac{1}{2} \BFdelta^\T \nabla \hat{Q}_\varpi\big( (1-t)\bbeta_1 + t\bbeta_2 \big) \BFdelta  = \frac{1}{2 n} \sn  K_\varpi\big(  \varepsilon_i - \langle \xb_i,  (1-t)\bbeta_1 + t \bbeta_2 - \bbeta^* \rangle \big) \langle \xb_i, \BFdelta \rangle^2   \\
& = \frac{1}{2 n} \sum_{i=1}^n  K_\varpi\big(  \varepsilon_i - t \langle\xb_i, \BFdelta \rangle - \langle \xb_i, \bbeta_1 - \bbeta^* \rangle \big) \langle \xb_i, \BFdelta \rangle^2
\end{align*}
for some $t\in [0,1]$, where $\BFdelta = \bbeta_2- \bbeta_1$. For each $i$, define the event 
\$
 \mathcal{F}_i = \big\{ | \varepsilon_i | \leq \varpi/4 \big\} \cap \big\{ |  \langle \xb_i, \bbeta_1 - \bbeta^* \rangle | \leq \varpi/4 \big\} \cap \big\{ |\langle \xb_i, \BFdelta \rangle | \leq \| \BFdelta \|_\Sigma \cdot \varpi/(2 r) \big\} ,
\$
on which $| t \langle\xb_i, \BFdelta \rangle + \langle \xb_i, \bbeta_1 - \bbeta^* \rangle | \leq \varpi /2 + \varpi /4= 3\varpi/4$ for all $\bbeta_2 \in \bbeta_1 + \Theta_\Sigma(r)$. Consequently.
\$
\hat Q_\varpi( \bbeta_1 )  - \hat Q_\varpi( \bbeta_2  )   - \langle \nabla \hat Q_\varpi(\bbeta_2) , \bbeta_1 - \bbeta_2 \rangle \geq \frac{\kappa_l }{2n \varpi} \sn \langle \xb_i, \BFdelta \rangle^2 \mathbbm{1}(\cF_i) ,
\$
where $\kappa_l = \min_{|u|\leq 1} K(u)$.
For $R>0$, define  functions  $\varphi_R(u)=u^2 \mathbbm{1}(|u|\leq R/2) + \{ u\sign(u) -R \}^2 \mathbbm{1}( R/2 < |u| \leq R)$ and $\psi_R(u)= \mathbbm{1}(|u|\leq R/2) + \{ 2 - (2/R) \sign(u)\} \mathbbm{1} (R/2 < |u| \leq R)$, which are smoothed versions of $u\mapsto u^2 \mathbbm{1}(|u|\leq R)$ and $u\mapsto \mathbbm{1}(|u|\leq R)$.
Moreover,  note that $\varphi_{cR}(cu) = c^2 \varphi_{R}(u)$ for any $c > 0$ and $\varphi_0(u) = 0$. The right-hand side of the above inequality can be further bounded from below by
\#
 & \frac{\kappa_l}{2 n \varpi } \sn   \mathbbm{1}\big(|\varepsilon_i| \leq \varpi /4 \big)  \cdot \varphi_{\| \BFdelta \|_\Sigma \cdot \varpi/(2 r)} (\langle \xb_i, \BFdelta \rangle  )  \cdot \psi_{\varpi/4} (\langle \xb_i, \bbeta_1 - \bbeta^* \rangle )   \nn \\
 & = \kappa_l \| \BFdelta \|_\Sigma^2 \cdot 
 \frac{1}{2}  \underbrace{ \frac{1}{  n \varpi } \sn   \mathbbm{1}\big(|\varepsilon_i| \leq \varpi/4\big)  \cdot \varphi_{   \varpi/(2r)} (\langle \xb_i, \BFdelta /\| \BFdelta \|_\Sigma\rangle  ) \cdot  \psi_{\varpi/4} (\langle \xb_i, \bbeta_1 - \bbeta^* \rangle )  }_{ =: V_n(\bbeta_1, \bbeta_2) } . \label{def:Vn}
\#

In the following, we bound $\E  V_n(\bbeta_1, \bbeta_2) $ and $ V_n(\bbeta_1, \bbeta_2)  - \E  V_n(\bbeta_1, \bbeta_2) $, respectively. Write $\vb = \Sigma^{1/2} \BFdelta / \| \BFdelta \|_\Sigma \in \mathbb{S}^{p-1}$, we have
\$
 & \E  \big\{ \varphi_{\varpi/(2 r)} ( \langle  \wb_i , \vb \rangle  ) \cdot  \psi_{\varpi/4} (\langle \xb_i, \bbeta_1 - \bbeta^*\rangle ) \cdot \mathbbm{1}\big(|\varepsilon_i|\leq \varpi/4 \big) \big\}  \\
 & \geq     \frac{1}{2} f_l' \varpi \cdot  \E  \big\{ \langle \wb_i, \vb \rangle^2  \mathbbm{1}\big(  | \langle \wb_i, \vb \rangle | \leq \varpi /(4 r)\big)  \cdot  \psi_{\varpi/4} (\langle \xb_i, \bbeta_1  - \bbeta^*\rangle )    \big\} \\ 
 & \geq   \frac{1}{2}   f_l' \varpi \cdot  \bigg\{  1-  \E  \langle \wb_i, \vb \rangle^2  \mathbbm{1}\big(  | \langle \wb_i, \vb\rangle | > \varpi /(4 r)\big)     -    \E    \langle \wb_i, \vb \rangle^2  \mathbbm{1}\big(  | \langle \xb_i, \bbeta_1 - \bbeta^* \rangle | >  \varpi / 8 \big)  
 \bigg\} \\
 & \geq  \frac{1}{2} f_l' \varpi \cdot  \bigg\{  1-   m_4^{1/2}   P\big(  | \langle \wb_i, \vb   \rangle | > \varpi / (4 r) \big)^{1/2}   -    m_4^{1/2}  P\big(  | \langle \xb_i, \bbeta_1 - \bbeta^* \rangle | > \varpi/ 8 \big)^{1/2} 
 \bigg\}    .
\$
Under Condition~\ref{cond.subgaussian} with $\upsilon_1\geq 1$, it can be shown that for any $\vb \in \mathbb{S}^{p-1}$ and $\bbeta_1 \in \Theta^*_\Sigma(r/2)$,
\$
P\big(  | \langle \wb_i,  \vb  \rangle | > \varpi/ (4 r) \big) \leq 2  \exp\bigg\{ \frac{1}{2  }-\frac{1}{2} \frac{\varpi^2}{(4 \upsilon_1 r )^2 }  \bigg\} ~\mbox{ and } \\
P\big(  | \langle \xb_i, \bbeta_1 - \bbeta^* \rangle | > \varpi/ 8 \big) \leq 2  \exp\bigg\{ \frac{1}{2  }-\frac{1}{2} \frac{\varpi^2}{(4 \upsilon_1  r )^2 }  \bigg\}.
\$
Let $\varpi/ r  \geq 16 (m_4 \vee 3)^{1/4}\upsilon_1 $.  It then follows from a numerical calculation that
\$
 \frac{1}{4}\bigg\{  1-   m_4^{1/2}   P\big(  | \langle \wb_i, \vb   \rangle | > \varpi / (4 r) \big)^{1/2}   -    m_4^{1/2}  P\big(  | \langle \xb_i, \bbeta_1 - \bbeta^* \rangle | > \varpi / 8 \big)^{1/2} 
 \bigg\}  \geq  c_0 \approx  0.248
\$ 
holds uniformly over $\vb \in \mathbb{S}^{p-1}$ and $\bbeta_1 \in \bbeta^* + \Theta_\Sigma(r)$. Putting together the pieces yields
\#
    \inf_{  \bbeta_1 \in \bbeta^* +  \Theta_\Sigma(r/2), \, 
\bbeta_2 \in \bbeta_1 + \Theta_\Sigma(r) }  \frac{1}{2} \E  V_n (\bbeta_1 , \bbeta_2 )  \geq     c_0 f_l'  . \label{mean.Vn.lbd}
\#

Next we upper bound the supremum
$$
\Omega(r)    : = \sup_{ \bbeta_1 \in \bbeta^* + \Theta_\Sigma(r/2), \,  \bbeta_2 \in \bbeta_1 + \Theta_\Sigma(r)  }   \big\{   -  V_n(\bbeta_1, \bbeta_2 )  +   \E   V_n(\bbeta_1, \bbeta_2 ) \big\} .
$$ 
Write $V_n(\bbeta_1, \bbeta_2) = (1/n) \sn v_i(\bbeta_1, \bbeta_2 )$, where $v_i(\bbeta_1, \bbeta_2 ) =   \varpi^{-1}\mathbbm{1}\big(|\varepsilon_i| \leq \varpi/4\big)  \cdot \varphi_{   \varpi/(2r)} (\langle \wb_i, \vb \rangle  ) \cdot  \psi_{\varpi/4} (\langle \xb_i, \bbeta_1 - \bbeta^* \rangle )$ satisfies 
\$
	 0 \leq v_i(\bbeta_1, \bbeta_2) \leq  \frac{\varpi}{(4 r)^2} ~~\mbox{ and }~~
\E  v^2_i(\bbeta_1, \bbeta_2) \leq \frac{ f_u   m_4}{2 \varpi}.
\$
Applying Theorem~7.3 in \cite{B2003}, a refined version of Talagrand’s inequality, we obtain that for any $t>0$,
\#
\Omega(r)   \leq  \E  \Omega(r)   +  \big\{ \E  \Omega(r)   \big\}^{1/2}  \frac{1}{2r} \sqrt{\frac{ \varpi t}{n}}  +  (f_u m_4)^{1/2}  \sqrt{\frac{t}{ n \varpi }}  + \frac{\varpi}{(4r)^2} \frac{t}{3 n } \label{talagrand.concentration}
\#
holds with probability at least $1-e^{-t}$. To bound the expectation $\E  \Omega(r)$, using Rademacher symmetrization and Lemma~4.5 in \cite{LT1991}, we obtain
\#
\E  \Omega(r) \leq 2 \cdot \sqrt{\frac{\pi}{2}} \cdot \E \bigg(  \sup_{\bbeta_1, \bbeta_2}   \mathbb{G}_{\bbeta_1, \bbeta_2 }    \bigg),  \label{gaussian.complexity}
\#
where the supremum is taken over $( \bbeta_1 , \bbeta_2)  \in \Theta^*_\Sigma(r/2) \times \{ \bbeta_1 + \Theta_\Sigma(r)\}$---including $(\bbeta^*, \bbeta^*)$, 
$$ 
\mathbb{G}_{\bbeta_1, \bbeta_2 }  = \frac{1}{n \varpi } \sn g_i \cdot \chi_i \cdot \varphi_{\varpi /(2 r)} ( \langle  \wb_i , \vb \rangle  ) \cdot  \psi_{\varpi/4} (\langle \xb_i, \bbeta_1 - \bbeta^*\rangle )  ~\mbox{ with }~ \chi_i  = \mathbbm{1}\big( | \varepsilon_i | \leq \varpi /4 \big) ,
$$
and $g_1, \ldots, g_n$ are independent standard normal variables. Conditional on $\{(y_i,\xb_i)\}_{i=1}^n$,  $\mathbb{G}_{\bbeta_1, \bbeta_2 } $ is a centered Gaussian process and $\mathbb{G}_{\bbeta^*, \bbeta^* }  = 0$. 
For any two admissible pairs $(\bbeta_1, \bbeta_2)$ and $(\bbeta_1',\bbeta_2')$, write $\vb = \Sigma^{1/2} (  \bbeta_2 - \bbeta_1 )/ \| \bbeta_2 - \bbeta_1 \|_\Sigma$, $\vb' = \Sigma^{1/2} ( \bbeta_2' -\bbeta_1') / \| \bbeta'_2 - \bbeta'_1 \|_\Sigma$, and note that
\begin{align*}
\mathbb{G}_{\bbeta_1, \bbeta_2 } - \mathbb{G}_{\bbeta'_1, \bbeta'_2 }  & =  \GG_{\bbeta_1, \bbeta_2 } -  \GG_{\bbeta_1' ,  \bbeta'_1 +\BFdelta } +  \GG_{\bbeta_1'  , \bbeta'_1 +\BFdelta  }  - \GG_{\bbeta'_1, \bbeta'_2 }  \\
& = \frac{1}{n \varpi} \sn g_i \cdot \chi_i \cdot  \varphi_{\varpi/(2 r)} (\langle \wb_i, \vb \rangle ) \big\{ \psi_{\varpi/4} ( \langle \xb_i, \bbeta_1 - \bbeta^* \rangle ) - \psi_{\varpi/4} ( \langle \xb_i, \bbeta'_1 - \bbeta^* \rangle  ) \big\} \\
&~~~~ + \frac{1}{n \varpi} \sn g_i \cdot \chi_i \cdot  \psi_{\varpi/4} ( \langle \xb_i, \bbeta'_1 - \bbeta^* \rangle  )   \big\{ \varphi_{\varpi/(2 r)} (\langle \wb_i, \vb \rangle ) - \varphi_{\varpi/(2 r)} (\langle \wb_i, \vb' \rangle )  \big\} .
\end{align*}
Recall that $\varphi_R$ and $\psi_R$ are, respectively, $R$- and $(2/R)$-Lipschitz continuous, and $\varphi_R(u) \leq (R/2)^2$. It follows that
\$
 & \E^* \big( \GG_{\bbeta_1, \bbeta_2 } -  \GG_{\bbeta_1' ,  \bbeta'_1 +\BFdelta }  \big)^2 \nn \\
& \leq   \frac{1}{(n \varpi)^2 }  \bigg( \frac{8}{\varpi} \bigg)^2\bigg( \frac{\varpi}{4r} \bigg)^4 \sn \chi_i  \langle \xb_i, \bbeta_1 - \bbeta_1' \rangle^2  =  \frac{1}{4 r^4 n^2 } \sn \chi_i  \langle \xb_i, \bbeta_1 - \bbeta_1'  \rangle^2 
\$
and 
\$
 & \E^* \big(    \GG_{\bbeta_1'  , \bbeta'_1 +\BFdelta  }  - \GG_{\bbeta'_1, \bbeta'_2 }  \big)^2 \nn \\
& \leq \frac{1}{(n \varpi)^2 } \bigg( \frac{\varpi}{2r} \bigg)^2  \sn  \chi_i \big( \langle \wb_i, \vb\rangle - \langle \wb_i , \vb' \rangle \big)^2  = \frac{1}{4 r^2 n^2 }\sn   \chi_i\langle \wb_i, \vb - \vb' \rangle^2 .
\$
Together, the last two displays imply
\$
\E^* \big( \mathbb{G}_{\bbeta_1, \bbeta_2 } - \mathbb{G}_{\bbeta'_1, \bbeta'_2 }  \big)^2 \leq  
  \frac{1}{2 r^4 n^2 } \sn \chi_i  \langle \xb_i, \bbeta_1 -  \bbeta_1'  \rangle^2  +  \frac{1}{2r^2 n^2 }\sn  \chi_i \langle \wb_i, \vb - \vb' \rangle^2.
\$
Define another (conditional) Gaussian process $\{ \mathbb{Z}_{\bbeta_1, \bbeta_2} \}$ as 
\$
\mathbb{Z}_{\bbeta_1, \bbeta_2} =  \frac{1}{2^{1/2} r^2 n } \sn g_i'  \chi_i \langle  \xb_i, \bbeta_1 - \bbeta^* \rangle + \frac{1}{2^{1/2} r n} \sn g''_i \chi_i \frac{\langle \xb_i, \bbeta_2 - \bbeta_1 \rangle }{\| \bbeta_2 - \bbeta_1 \|_\Sigma} ,
\$
where $\{ g'_i \}$ and $\{g_i''\}$ are two dependent copies of $\{ g_i\}$. The above calculations show that
\$
 \E^* \big( \mathbb{G}_{\bbeta_1, \bbeta_2 } - \mathbb{G}_{\bbeta'_1, \bbeta'_2 }  \big)^2  \leq \E^* \big( \mathbb{Z}_{\bbeta_1, \bbeta_2 } - \mathbb{Z}_{\bbeta'_1, \bbeta'_2 }  \big)^2  .
\$
 Using Sudakov-Fernique's Gaussian comparison inequality (see, e.g. Theorem~7.2.11 in \cite{V2018}) gives
\#
 \E^* \bigg( \sup_{\bbeta_1, \bbeta_2}  \mathbb{G}_{\bbeta_1, \bbeta_2 }  \bigg) 
 \leq  \E^* \bigg( \sup_{\bbeta_1, \bbeta_2}   \mathbb{Z}_{\bbeta_1, \bbeta_2 }   \bigg) . \label{gaussian.comparison}
\#
The same bound also applies to unconditional expectations of the two suprema.  For the random process $\{ \mathbb{Z}_{\bbeta_1, \bbeta_2 } \}$, using the bound $P(|\varepsilon_i|\leq \varpi/4|\xb_i ) \leq f_u \varpi/2$ (a.s.) we derive that
\$
 & \E  \bigg( \sup_{\bbeta_1, \bbeta_2}   \mathbb{Z}_{\bbeta_1, \bbeta_2 }   \bigg)  \\
 & \leq \frac{1}{2^{1/2} r^2} \sup_{\bbeta_1 \in \Theta^*_\Sigma(r/2) } \| \bbeta_1 -\bbeta^* \|_\Sigma  \cdot \E \bigg\| \frac{1}{n} \sn  g_i' \chi_i \wb_i \bigg\|_2 +  \frac{1}{2^{1/2} r }  \E  \bigg\| \frac{1}{n} \sn  g''_i \chi_i \wb_i \bigg\|_2  \\
& \leq f_u^{1/2} \frac{1}{4 r}     \sqrt{\frac{ \varpi p}{n}} +  f_u^{1/2} \frac{ 1}{2r }  \sqrt{\frac{\varpi p}{n}} = f_u^{1/2}\frac{3  }{4 r} \sqrt{\frac{ \varpi p}{n}}.
\$
This, joint with \eqref{talagrand.concentration}, \eqref{gaussian.complexity} and \eqref{gaussian.comparison}, implies that, with probability at least $1-e^{-t}$,
\#
 \Omega(r)  & \leq  \frac{5}{4} \E  \Omega(r) + (f_u m_4)^{1/2} \sqrt{\frac{t}{n \varpi}} + (4+1/3)\frac{\varpi t}{(4r)^2 n}  \nn \\
& \leq   (f_u m_4)^{1/2} \sqrt{\frac{t}{n \varpi}}  + 15 (\pi/2)^{1/2}  f_u^{1/2}\frac{1}{ 8 r} \sqrt{\frac{\varpi p}{ n}} +   (4+1/3)\frac{\varpi t}{16 r^2 n}   . \label{concentration.Vn.ubd}
\#

Finally, combining \eqref{def:Vn}, \eqref{mean.Vn.lbd} and \eqref{concentration.Vn.ubd}, we conclude that for all $\bbeta_1 \in \Theta^*_\Sigma(r/2)$ and $\bbeta_2 \in \bbeta_1+\Theta_\Sigma(r)$ with $r = \varpi/(16 \max\{m_4, 3 \}^{1/4} \upsilon_1)$,
\$
 & \frac{  \hQ_h( \bbeta_1 )  - \hQ_h( \bbeta_2  )   - \langle \nabla \hQ_h(\bbeta_2) , \bbeta_1 - \bbeta_2 \rangle }{ \kappa_l  \| \bbeta_1- \bbeta_2 \|_\Sigma^2 }  \\
 & ~~~~~~~~~~~~~~~~~ \geq  c_0  f_l' -  \frac{1}{2} (f_u m_4)^{1/2} \sqrt{\frac{t}{n \varpi}}  -    C_1   f_u^{1/2}  m_4^{1/4} \upsilon_1 \sqrt{\frac{ p }{n \varpi }}  -  C_2 m_4^{1/2} \upsilon_1^2 \frac{t}{  n \varpi  } 
\$
holds with probability at least $1-e^{-t}$, where $C_1, C_2>0$ are absolute constants.  Let $n\varpi$ be sufficiently large---$n \varpi \gtrsim  m_4^{1/2}\upsilon_1^2 (p+t)$---so that the right-hand side is bounded from below by $c_0  f_l' /2$. This completes the proof. 
 \Halmos

\section{Proofs of Technical Lemmas}

\subsection{Proof of Lemma~\ref{lem:loss.approxi}}

For simplicity, define the re-scaled kernel $W(u) = K_\varpi(u)=K(u/\varpi) / \varpi$, which is also symmetric and non-negative.
Note that $\rho_\tau(u)= |u|/2 + (\tau-1/2) u$ and $(\rho_\tau * W)(u) = (1/2) \int_{-\infty}^\infty |v| W(u-v){\rm d}v +  (\tau-1/2) u $. It thus suffices to prove that for any $u\in \mathbb{R}$,
$$
    |u| \leq \int_{-\infty}^\infty |v| W(u-v){\rm d}v \leq |u| + \kappa_1 \varpi .
$$
The upper bound follows from the triangle inequality and a change of variable that
\begin{align*}
   & \int_{-\infty}^\infty |v| W(u-v){\rm d}v    \leq \int_{-\infty}^\infty |u-v| W(u-v){\rm d}v + \int_{-\infty}^\infty |u | W(u-v){\rm d}v  \\ 
   & = |u| + \frac{1}{\varpi} \int_{-\infty}^\infty |u-v| K\bigg(\frac{u-v}{\varpi} \bigg) {\rm d} v  = |u| +  \varpi \int_{-\infty}^\infty |s| K(s) {\rm d} s  = |u| + \kappa_1 \varpi .
\end{align*}
 
To prove the lower bound, we first assume $u\geq 0$. Utilizing the change of variable and the properties $\int W(u){\rm d}u=1$ and $\int u W(u){\rm d}u =0$,  we derive that
\begin{align*}
   &  \int_{-\infty}^\infty |v| W(u-v){\rm d}v = \int_{-\infty}^\infty | u + w| W(w) {\rm d} w \\ 
    & = \int_{-u}^{\infty} (u+w) W(w) {\rm d} w - \int_{-\infty}^{-u} (u+w) W(w) {\rm d} w \\ 
    & =  \int_{-u}^{\infty} (u+w) W(w) {\rm d} w +  \int_{u}^{\infty } (-u+w) W(w) {\rm d} w \\
    & = u \int_{-u}^u W(w) {\rm d} w + \int_{-u}^\infty w W(w) {\rm d} w + \int_u^\infty w W(w) {\rm d}w  \\ 
    & = u -  \int_u^\infty  u W(w) {\rm d} w -  \int_{-\infty}^{-u}  u W(w) {\rm d} w+ \underbrace{\int_{-u}^\infty w W(w) {\rm d} w}_{= -\int_{-\infty}^{-u} w W(w){\rm d}w ~{\rm by~symmetry}} + \int_u^\infty w W(w) {\rm d}w  \\ 
    & = u + \underbrace{\int_u^\infty (w-u) W(w) {\rm d}w}_{\geq 0} - \underbrace{\int_{-\infty}^{-u} (w+u) W(w) {\rm d} w}_{\leq 0} \geq u .
\end{align*}
This proves the lower bound when $u\geq 0$. The case where $u<0$ can be proven using a similar argument, and therefore we omit the details for brevity. \Halmos

\subsection{Proof of Lemma~\ref{lem:population.loss.curvature}}

Starting with $R_\varpi(\BFdelta)$,  using a second-order Taylor series expansion yields
\#
	R_\varpi(\BFdelta) =  \frac{1}{2} \E \big\{  K_\varpi ( \varepsilon  -  t\langle \BFw  ,  \BFv \rangle     ) \langle \BFw  , \BFv \rangle^2 \big\} ~~\mbox{ for some } t\in [0,1] , \nn
\#
where $\BFw = \Sigma^{-1/2}\BFx$ and $\BFv = \Sigma^{1/2}\BFdelta= \Sigma^{1/2} (\BFbeta- \BFbeta^*)$.
By a change of variable and the Lipschitz continuity of $f_{\varepsilon |\BFx}(\cdot)$, we have
\$
 \E \big\{  K_\varpi ( \varepsilon  -  t\langle \BFw  ,  \BFv \rangle     ) | \BFx \big\} = \int_{-\infty}^\infty K(v) f_{\varepsilon |\BFx} ( t \langle \BFw, \BFv \rangle + \varpi v ) {\rm d} v \geq    f_{\varepsilon |\BFx} ( t \langle \BFw, \BFv \rangle  )  - l_0(\BFx)  \kappa_1 \varpi .
\$ 
It follows that for any $\BFdelta\in \RR^p$ satisfying $\| \BFdelta \|_{\Sigma}\leq 1$ (so that $\BFv \in \mathbb{B}^p(1)$),
\#
 & 2 R_\varpi(\BFdelta )  \geq \E \big\{ f_{\varepsilon|\BFx}(t \langle\BFw,\BFv\rangle) \langle \BFw, \BFv\rangle^2 \big\}  -  l_0 \kappa_1 \varpi \cdot  \E \langle \BFw, \BFv \rangle^2 \nn\\
  & = \E \big\{ f_{\varepsilon|\BFx}( \underbrace{ t \| \BFv \|_2    }_{\in [0,1]} \langle\BFw,\BFv/\| \BFv \|_2\rangle) \langle \BFw, \BFv/\| \BFv \|_2\rangle^2 \big\}  \cdot \| \BFv \|_2^2 -  l_0 \kappa_1 \varpi \cdot \| \BFv \|_2^2  \nn\\
 &   \geq  (f_l - l_0\kappa_1 \varpi )  \| \BFv \|_2^2 , \nn  
\# 
where the second inequality follows from Condition~\ref{cond.density}.
The same argument also applies to $B_\varpi(\BFdelta)$. We thus omit the details. \Halmos
 
\subsection{Proof of Lemma~\ref{lem:loss.diff}}

We only need to derive an upper bound for $ \hat D_\varpi(\BFdelta ) - D_\varpi(\BFdelta)$ uniformly over $\BFdelta \in \RR^p$ in a compact subset. The same argument applies to  $D_\varpi(\BFdelta) - \hat D_\varpi(\BFdelta)$.  For each sample $\BFz_i = (\BFw_i, \varepsilon_i)$ with $\BFw_i = \Sigma^{-1/2} \BFx_i$, define $d_\varpi(\BFv; \BFz_i) =  \ell_\varpi(\varepsilon_i - \langle \BFw_i, \BFv \rangle ) - \ell_\varpi(\varepsilon_i)$, so that $\hat D_\varpi(\BFdelta) = (1/n) \sn d_\varpi(\BFv; \BFz_i)$ for $\BFv = \Sigma^{1/2} \BFdelta$. Note that $\ell_\varpi(\cdot)$ is continuously differentiable with $|\ell'_\varpi(u) | \leq \bar \tau = \max(\tau, 1-\tau)$ for all $u$. Hence, for any $\BFz_i$ and $\BFv, \BFv' \in \RR^p$, $ | d_\varpi(\BFv; \BFz_i) -d_\varpi(\BFv'; \BFz_i) |  \leq \bar \tau | \langle \BFw_i, \BFv \rangle - \langle \BFw_i, \BFv'\rangle |$.
In other words, $d_\varpi(\BFv;\BFz_i)$ is $ \bar \tau$-Lipschitz continuous in $\langle \BFw_i, \BFv\rangle$.

For any fixed $r>0$ and some $\epsilon \in (0,1)$ to be determined, define the random variable 
$$
	\Delta_\epsilon(r)  =   \frac{1-\epsilon}{2\bar \tau r} n^{1/2}\sup_{\BFv \in \BB^p(r)}  \big\{  \hat  D_\varpi(\BFv) -  D_\varpi(\BFv) \big\} = \frac{1-\epsilon}{2\bar \tau r}   \sup_{\BFv \in \BB^p(r)} \bigg\{ \frac{1}{n^{1/2}} \sn  (1-\E) d_\varpi(\BFv;\BFz_i)  \bigg\} .
$$  
By Chernoff's inequality, for any $z \geq 0$,
\#
	P\{ \Delta_\epsilon(r) \geq z \} \leq  \exp \bigg[  - \sup_{\lambda \geq 0 }   \big\{  \lambda z - \log \E e^{\lambda  \Delta_\epsilon(r) }  \big\}  \bigg] . \label{chernoff.bd}
\#
To control the moment generating function $ \E e^{\lambda  \Delta_\epsilon(r) }$,  by Rademacher symmetrization we have 
\$
\E e^{\lambda  \Delta_\epsilon(r) } \leq  \E  \exp\bigg\{   (1-\epsilon) \frac{ 2 \lambda }{2 \bar \tau r }\sup_{\BFv \in \BB^p(r)}   \frac{1}{n^{1/2}}   \sn e_i  d_\varpi(\BFv ; \BFz_i)   \bigg\} , 
\$
where $e_1, \ldots, e_n$ are independent Rademacher random variables.  Recall that $d_\varpi(\BFv ; \BFz_i)$ is $ \bar \tau$-Lipschitz continuous in $\langle \BFw_i, \BFv  \rangle$,  and $d_\varpi(\BFv ; \BFz_i)=0$ if $\langle \BFw_i, \BFv \rangle = 0$. Then, applying the Ledoux-Talagrand contraction inequality (see Theorem~4.12 and inequality (4.20) in \cite{LT1991}) yields
\$
& \E  \exp\bigg\{   (1-\epsilon )\frac{   \lambda }{  \bar \tau r} \sup_{\BFv \in \BB^p(r)}    \frac{1}{n^{1/2}} \sn e_i  d_\varpi (\BFv ; \BFz_i)  \bigg\}  \\
&  \leq  \E  \exp\bigg\{  (1-\epsilon) \frac{\lambda }{r} \sup_{\BFv \in \BB^p(r)}    \frac{1}{n^{1/2}}  \sn e_i  \langle \BFw_i, \BFv \rangle   \bigg\}  \leq \E \exp \bigg\{   (1-\epsilon)\lambda  \bigg\|  \frac{1}{n^{1/2}} \sn e_i \BFw_i \bigg\|_2 \bigg\} .
\$
For this $\epsilon \in (0,1)$, there exists an $\epsilon$-net $\{ \BFu_1,  \ldots , \BFu_{N_\epsilon} \}$ of $\mathbb{S}^{p-1}$ with cardinality $N_\epsilon \leq (1+2/\epsilon)^p$ such that $\| \sn e_i \BFw_i  \|_2 \leq (1-\epsilon)^{-1} \max_{1\leq j\leq N_\epsilon} \sn e_i \BFu_j^\T\BFw_i $. Therefore,
\$
 \E \exp \bigg\{    (1-\epsilon) \lambda \bigg\| \frac{1}{n^{1/2}} \sn e_i \BFw_i \bigg\|_2 \bigg\} 
  \leq \sum_{j=1}^{N_\epsilon}   \E \exp \bigg(  \frac{\lambda}{n^{1/2}}  \sn e_i \BFu_j^\T\BFw_i  \bigg) .
\$
Write $S_j = n^{-1/2} \sn e_i \BFu_j^\T\BFw_i$, which is a sum of zero-mean random variables.
Note that $e_i \in \{-1, 1\}$ is symmetric, and  Condition~\ref{cond.subgaussian} ensures that
$\log \E \exp (  c e_i \BFu_j^\T \BFw_i ) \leq c^2 \upsilon_1^2/2$ for all $c\in \RR$. Consequently,
\$
 & \E \exp(\lambda S_j ) =  \prod_{i=1}^n \E \exp(  \lambda n^{-1/2} e_i \BFu_j^\T \BFw_i  ) \leq \prod_{i=1}^n e^{\lambda^2 \upsilon_1^2 / (2n )} = e^{\lambda^2 \upsilon_1^2 / 2} ,
\$
from which it follows that
\$
  \log  \E e^{\lambda  \Delta_\epsilon(r) }  \leq  \log N_\epsilon +  \frac{  1 }{ 2} \upsilon_1^2 \lambda^2 .
 \$
For any $u\geq 0$,  note that
\$
\sup_{\lambda \geq 0}  \big\{  \lambda z - \log \E e^{\lambda  \Delta_\epsilon(r) }  \big\} \geq -\log N_\epsilon  + \sup_{\lambda \geq 0}  \bigg(  \lambda  z  - \frac{ 1}{2} \upsilon_1^2\lambda^2 \bigg) = -\log N_\epsilon + z^2/(2\upsilon_1^2).
\$
Substituting this into \eqref{chernoff.bd} and taking $v= z^2/(2\upsilon_1^2)$, we obtain that with probability at least $1-\exp \{ p\log(1+2/\epsilon) - v \}$,
\#
	 \sup_{\BFv \in \BB^p(r)}   \big\{ \hat D_\varpi(\BFv )   - D_\varpi(\BFv)  \big\}  \leq \frac{2\bar \tau  \upsilon_1 }{1-\epsilon}   r   \sqrt{\frac{2v}{n}} . \label{Rh.pointwise.bound}
\# 
This proves \eqref{loss.diff.max.bound} by setting $\epsilon=2/(e^4-1)$ and $v= 4p +u$.

Via a peeling/slicing argument, next, we prove a uniform version of \eqref{Rh.pointwise.bound}, which holds for all $\BFv \in \BB^p(r_l, r_u) = \{ \BFv \in \RR^p: r_l \leq \| \BFv \|_2 \leq r_u\}$.  For some $\gamma>1$ to be determined, and positive integers $k =1 ,\ldots, N := \lceil \log (\frac{r_u}{r_l} )/\log(\gamma) \rceil$, define the sets $\Theta_k=\{ \BFv \in \RR^p:  \gamma^{k-1} r_l \leq \|\BFv \|_2 \leq \gamma^{k } r_l \}$, so that $\BB^p(r_l, r_u) \subseteq \cup_{k=1}^N \Theta_k$.
Then,
\$
 & 	P\bigg\{   \exists \BFv  \in \BB^p(r_l, r_u) \mbox{ s.t. }   \hat  D_\varpi(\BFv )   -  D_\varpi(\BFv)  >  \frac{2\sqrt{2}\gamma}{1-\epsilon} \bar \tau \upsilon_1   \| \BFv \|_2  \sqrt{\frac{ v}{n}}    \bigg\} \\
& \leq \sum_{k=1}^N  	P\bigg\{   \exists \BFv \in \Theta_k  \mbox{ s.t. }   \hat D_\varpi(\BFv )   -  D_\varpi(\BFv)   >\frac{2\sqrt{2}\gamma }{1-\epsilon}  \bar \tau  \upsilon_1  \gamma^{k-1} r_l    \sqrt{\frac{v}{n}}     \bigg\} \\
& \leq   \sum_{k=1}^N  	P\bigg\{  \sup_{\BFv  \in \BB^p(\gamma^k r_l) }   \hat D_\varpi(\BFv )   -  D_\varpi(\BFv)    >  \frac{2\sqrt{2}}{1-\epsilon} \bar \tau  \upsilon_1 \gamma^{k } r_l    \sqrt{\frac{v}{n}}    \bigg\} \\
& \stackrel{({\rm i})}{\leq } \sum_{k=1}^N   \exp\big\{ p \log(1+2/\epsilon) - v \big\} \leq  \lceil \log (\tfrac{r_u}{r_l})/\log(\gamma) \rceil  \exp\big\{ p \log(1+2/\epsilon) - v  \big\} ,
\$
where inequality (i) follows from \eqref{Rh.pointwise.bound} with $r= \gamma^{k } r_l$ for $k=1,\ldots, N$.  Taking $\epsilon=2/(e^4-1)$, $\gamma = e^{1/e}$ and $v=4p+\log\{ e \log(\frac{r_u}{r_l}) \}+u$ yields that with probability at least $1- e^{-u}$,
\#
\hat D_\varpi(\BFv)  - D_\varpi(\BFv) \leq   4.25 \bar \tau \upsilon_1 \cdot \| \BFv \|_2 \sqrt{\frac{4 p + \log\{ e \log( r_u/r_l ) \} +  u}{n}} . \nn
\# 
This proves \eqref{loss.diff.uniform.bound} by taking $(r_l, r_u) = (\delta, r)$. \Halmos

\subsection{Proof of Lemma~\ref{lem:grad.diff}}

\noindent
{\sc Proof of \eqref{grad.unif.convergence}}. By a change of variable $\BFv =  \Sigma^{ 1/2} (\BFbeta - \BFbeta^*)$, define the centered gradient process
\#
  G_\varpi(\BFv )  =    \Sigma^{-1/2} \{ \hat Q_\varpi(\BFbeta) - Q_\varpi(\BFbeta) \}     =  \frac{1}{n}  \sn  (1-\E ) \underbrace{ \{  \bar{K}_\varpi(\BFw_i^\T \BFv - \varepsilon_i ) - \tau \} }_{=: \xi_{i,\BFv} } \BFw_i   . \nn
\# 
For any $\epsilon  \in (0, R)$, there exists an $\epsilon$-net $\{ \BFv_1, \ldots , \BFv_{N_\epsilon} \}$ of $\BB^p(R)$ with $N_\epsilon \leq (1+2R/\epsilon)^p$. For any $\BFv \in \BB^p(R)$, there exists some $1\leq j\leq N_\epsilon$ such that $\| \BFv - \BFv_j \|_2\leq \epsilon$. Then, by the triangle inequality,
\$
	 \|   G_\varpi(\BFv )  \|_2 \leq \|  G_\varpi (\BFv) - G_\varpi(\BFv_j ) \|_2 + \| G_\varpi(\BFv_j) \|_2. 
\$  

We first control the approximation error $ \|  G_\varpi (\BFv) - G_\varpi(\BFv_j ) \|_2 $. Using integration by parts and a change of variable, we have $\E(\xi_{i,\BFv} |\BFx_i ) = \int_{-\infty}^\infty K(u) F_{\varepsilon_i | \BFx_i }(\BFw_i^\T \BFv - h u ) {\rm d} u$. Condition~\ref{cond.density} ensures that
\$
   \| \E(\xi_{i,\BFv}  \BFw_i  ) - \E(\xi_{i,\BFv_j}  \BFw_i  ) \|_2  \leq   \sup_{\BFu \in\mathbb{S}^{p-1}}  \E  \big\{ f_u(\BFx)|\langle \BFw_i, \BFu\rangle \cdot \langle \BFw_i, \BFv -\BFv_j \rangle |\big\}  \leq  f_u  \epsilon.
\$
Turning to $\xi_{i,\BFv}  \BFw_i$ and $\xi_{i,\BFv_j}  \BFw_i$, since $K(\cdot)$ is bounded by $\kappa_u$, we have
\$
 \bigg\| \frac{1}{n}  \sn   \xi_{i,\BFv}   \BFw_i - \frac{1}{n}  \sn   \xi_{i,\BFv_j }   \BFw_i \bigg\|_2  & \leq \sup_{\BFu \in \mathbb{S}^{p-1}} \frac{1}{n}   \sn   |  \bar{K}_\varpi(\BFw_i^\T \BFv - \varepsilon_i) - \bar{K}_\varpi(\BFw_i^\T \BFv_j - \varepsilon_i)  | \cdot |\BFw_i^\T \BFu | \\
& \leq  \frac{\kappa_u}{\varpi} \sup_{\BFu \in \mathbb{S}^{p-1}} \frac{1}{n}   \sn   | \BFw_i^\T (\BFv - \BFv_j) | \cdot |\BFw_i^\T \BFu |  \\
& \leq \frac{\kappa_u  \epsilon }{\varpi}  \cdot  \lambda_{\max} \bigg(   \frac{1}{n} \sn \BFw_i \BFw_i^\T  \bigg) .
\$
Note that $\BFw_i = (1, \BFw_{i,-}^\T)^\T$, where $\BFw_{i,-} \in \RR^{p-1}$ are zero-mean sub-Gaussian random vectors.   By a standard covering argument, paired with Bernstein's inequality, it can be shown that (see, e.g. Theorem~4.6.1 in \cite{V2018}) with probability at least $1-(1/2)e^{-u}$,
\#
 \lambda_{\max} \bigg(   \frac{1}{n} \sn \BFw_i \BFw_i^\T  \bigg)    \leq  1 + C_1 \upsilon_1^2 \bigg(  \sqrt{\frac{p+ u}{n}} + \frac{p+u}{n}\bigg)  , \nn
\#
where $C_1 >0$ is an absolute constant. Provided $n\gtrsim \upsilon_1^4( p+u)$, it follows that with probability at least $1-(1/2)e^{-u}$, 
\#
     \|  G_\varpi (\BFv) - G_\varpi(\BFv_j ) \|_2 \leq f_u \epsilon + 2  \kappa_u \epsilon / \varpi   \label{approximation.error.bound}
\#
holds uniformly over all the pairs $(\BFv, \BFv_j)$ satisfying $\| \BFv - \BFv_j \|_2\leq \epsilon$.

It remains to deal with $\|  G_\varpi(\BFv_j)  \|_2$ for each $j=1,\ldots, N_\epsilon$. For simplicity, we write $\xi_{ij} = \xi_{i,\BFv_j} =\bar{K}_\varpi( \BFw_i^\T \BFv_j - \varepsilon_i ) - \tau$, and define
\$
 S_j = \frac{1}{n}  \sn (\xi_i - \E \xi_i) ~~\mbox{ and }~~ \bS_j = \frac{1}{n}   \sn \{ \xi_{ij}  \BFw_{i,-} -  \E( \xi_{ij}  \BFw_{i,-})\} \in \RR^{p-1}.
\$
Then, $\|  G_\varpi(\BFv_j)  \|_2^2 = S_j^2 + \| \bS_j  \|_2^2$. For $S_j$, note that $\xi_{ij} \in [-\tau, 1-\tau ]$. As a direct application of Hoeffding's inequality, we have that for any $v\geq 0$,
\#
	 P \bigg(  |S_j | \geq  \sqrt{\frac{\log 4  + v}{2n }}  \,\bigg) \leq   \frac{1}{2} e^{-v}. \label{intercept.grad.bound}
\#
Next we use a covering argument to bound $\| \bS_j  \|_2$. For any $\delta\in (0,1)$, there exists an $\delta$-net $\cM_\delta$ of $\mathbb{S}^{p-2}$ (unit sphere in $\RR^{p-1}$) with cardinality $M_\delta \leq (1+2/\delta)^{p-1}$ such that 
$$
 \|  \bS_j  \|_2 \leq (1-\delta)^{-1} \frac{2\bar \tau \upsilon_1}{n^{1/2}}\underbrace{ \max_{\BFu\in \cM_{\delta}} \frac{1}{2 \bar \tau \upsilon_1 n^{1/2}} \sn \langle \BFu, \xi_{ij} \BFw_{i,-} - \E( \xi_{ij} \BFw_{i,-} ) \rangle }_{=: \Delta_\delta },
$$
where $\bar \tau  = \max(1-\tau,\tau)$. For any $z \geq 0$, applying Chernoff's inequality gives 
\$
	P(\Delta_\delta \geq z ) \leq \exp\bigg\{ - \sup_{\lambda \geq 0}   \big(  \lambda z - \log \E e^{\lambda \Delta_\delta }   \big) \bigg\}  .
\$
By Rademacher symmetrization and independence,
\$
\E e^{\lambda \Delta_\delta }   \leq \E \exp\bigg(   \max_{\BFu\in \cM_{\delta}} \frac{\lambda}{ \bar \tau \upsilon_1 n^{1/2}} \sn e_i \xi_{ij} \BFu^\T \BFw_{i,-}  \bigg) \leq \sum_{u\in \cM_\delta } \prod_{i=1}^n \E \exp\bigg(    \frac{\lambda}{ \bar \tau \upsilon_1 n^{1/2}}   e_i \xi_{ij}  \BFu^\T \BFw_{i,-}   \bigg) ,
\$
where $e_i$'s are independent Rademacher random variables. Fix $\BFu \in \cM_\delta$, and set $\omega_i  = \BFu^\T \BFw_{i,-}   /\upsilon_1$. We further have
\$
  \E \exp\bigg(    \frac{\lambda}{ \bar \tau  n^{1/2}}   e_i \xi_{ij} \omega_i   \bigg)   = \sum_{k=0}^\infty \frac{1}{k!} \bigg( \frac{\lambda }{n^{1/2}}\bigg)^k  \underbrace{  \E (e_i \xi_{ij}  \omega_i  / \bar \tau )^k }_{= 0 ~{\rm if}~k~{\rm is~odd}} \leq  1+\sum_{\ell=1}^\infty \frac{1}{(2\ell)!} \bigg( \frac{\lambda }{n^{1/2}} \bigg)^{2\ell} \E (  \omega_i ^{2\ell} ).
 \$
 Recall from Condition~\ref{cond.subgaussian} that $\E (e^{t \omega_i }  )\leq e^{t^2/2}$ for all $t\in \RR$, and $\E (\omega_i^2) \leq 1$. For $\ell \geq 2$,
 \$
  \E( \omega_i^{2\ell} ) = 2\ell \int_0^\infty u^{2\ell -1} P(|\omega_i| \geq u) {\rm d} u \leq  4 \ell  \int_0^\infty  u^{2\ell-1} e^{-u^2/2} {\rm d}u = 2^{\ell+1} \ell!  .
 \$
Substituting this into the previous exponential moment bound yields
\$
 \E \exp\bigg(    \frac{\lambda}{ \bar \tau  n^{1/2}}   e_i \xi_{ij} \omega_i   \bigg)  \leq    1  +  \frac{\lambda^2}{n} +\sum_{\ell=2}^\infty \frac{4/3}{ \ell!} \bigg( \frac{\lambda^2 }{n } \bigg)^{ \ell} 
\leq \exp\bigg( \frac{2}{\sqrt{3}  n }  \lambda^2 \bigg), 
\$
which in turn implies
\$
 \log \E e^{\lambda \Delta_\delta } \leq \log M_\delta + \frac{2}{\sqrt{3}} \lambda^2 ~\mbox{ and }~ \sup_{\lambda\geq 0}  \big\{ \lambda z - \log \E e^{\lambda \Delta_\delta }  \big\}  \geq - \log M_\delta +  \frac{\sqrt{3}}{8} z^2 .
\$
Combining this with Chernoff's bound and a change of variable, we obtain that with probability at least $1-(1/2)e^{-v}$,
\#
 \|  \bS_j  \|_2^2  \leq \frac{ 32 \bar \tau^2\upsilon_1^2}{3^{1/2}(1-\delta)}   \frac{(p-1)\log(1+2/\delta) +\log 2 + v }{n} .  \label{coef.grad.bound}
\#
Together, \eqref{intercept.grad.bound} and \eqref{coef.grad.bound} with a properly chosen $\delta$, say, $\delta=2/(e^2-1)$, imply that $\| G_\varpi(\BFv_j  ) \|_2 \leq C_2 \upsilon_1 \sqrt{(p+v)/n}$ holds with probability at least $1-e^{-v}$, where $C_2>0$ is an absolute constant. Taking the union bound over $j=1,\ldots, N_\epsilon$, we have that with probability at least $1- \exp\{ p\log(1+2R/\epsilon) - v\}$,
\#
    \max_{1\leq j\leq N_\epsilon }\| G_\varpi(\BFv_j  ) \|_2 \leq C_2 \upsilon_1 \sqrt{\frac{p+ v }{n}} . \label{discretized.max.bound}
\#

Taking $\epsilon=R \varpi/n $ and $v=\log(2)+p\log(1+2  n/\varpi)+u$ in \eqref{approximation.error.bound} and \eqref{discretized.max.bound}, we obtain that  with probability at least $1-e^{-u}$,
\$
 \sup_{\BFv \in \BB^p(R) } \| G_\varpi(\BFv ) \|_2 \leq ( f_u \varpi +  2\kappa_u ) \frac{R}{n} + C_2 \upsilon_1 \sqrt{\frac{\log 2 + p\log ( e + 2e n/\varpi)+ u }{n}}  
\$
as long as $n\gtrsim \upsilon_1^4(p+u)$. This proves \eqref{grad.unif.convergence}.

For  the population gradient, noting that $|\bar{K}_\varpi( y_i - \BFx_i^\T \BFbeta ) - \tau| \leq \bar \tau$ for any $\BFbeta \in \RR^p$, we have $ \sup_{\BFbeta \in \RR^p}
\| \nabla Q_\varpi(\BFbeta) \|_{\Sigma^{-1}}   \leq \bar \tau \sup_{\BFu \in \mathbb{S}^{p-1}} \E | \langle \BFw, \BFu \rangle | \leq \bar \tau$. At $\BFbeta^*$, the bound $\| \nabla Q_\varpi(\BFbeta^*) \|_{\Sigma^{-1}} \leq 0.5 l_0 \kappa_2 \varpi^2$ follows from (C.2) in \cite{HPTZ2020}.

\medskip
\noindent
{\sc Proof of \eqref{hessian.unif.convergence}}.  For $\BFbeta \in \RR^p$, define after a change of variable $\BFv =  \Sigma^{1/2} ( \BFbeta - \BFbeta^*)$ that
\$
 \hat \BFH_\varpi(\BFv)  = \Sigma^{-1/2} \nabla^2 \hQ_\varpi( \BFbeta ) \Sigma^{-1/2}  = \frac{1}{n} \sn K_\varpi(\varepsilon_i - \langle \BFw_i, \BFv \rangle ) \BFw_i \BFw_i^\T~~\mbox{ and }~~ \BFH_\varpi(\BFv) =  \E \{ \hat \BFH_\varpi(\BFv) \}  .
\$
To upper bound $\sup_{\BFv \in \BB^p(R)}  \| \hat \BFH_\varpi(\BFv)  - \BFH_\varpi(\BFv)  \|_2 $, we use a similar discretization argument as in the proof of \eqref{grad.unif.convergence} and obtain
\$
 & \sup_{\BFv \in \BB^p(R)}  \| \hat \BFH_\varpi(\BFv)  - \BFH_\varpi(\BFv)  \|_2  \\
 & \leq \frac{l_K \epsilon }{\varpi^2} \max_{1\leq i\leq n} \| \BFw_i \|_2 \cdot  \bigg\| \frac{1}{n} \sn \BFw_i \BFw_i^\T \bigg\|_2  \\
 &~~~~ + \underbrace{  \max_{1\leq j\leq N_\epsilon }  \| \hat \BFH_\varpi(\BFv_j)  - \hat  \BFH_\varpi(\BFv_j)  \|_2  }_{{\rm sampling~error}}+   \underbrace{  \max_{1\leq j\leq N_\epsilon }  \| \BFH_\varpi(\BFv_j)  - \BFH_\varpi(\BFv_j)  \|_2 }_{{\rm~approximation~error}} ,
\$
where $l_K$ is the Lipschitz constant of the kernel $K(\cdot)$, $\{ \BFv_1, \ldots, \BFv_{N_\epsilon}\}$ is an $\epsilon$-net of $\BB^p(R)$ and $N_\epsilon \leq (1+2R/\epsilon)^p$.
For the last term on the right-hand side, the Lipschitz continuity of $f_{\varepsilon |\BFx}(\cdot)$ ensures that
\$
\max_{1\leq j\leq N_\epsilon }  \| \BFH_\varpi(\BFv_j)  - \BFH_\varpi(\BFv_j)  \|_2 \leq l_0 m_3 \epsilon , 
\$
where $m_3 = \sup_{\BFu \in \mathbb{S}^{p-1}} \E |\langle \BFw, \BFu \rangle|^3$.

We have already shown that $\lambda_{\max} ( n^{-1} \sn \BFw_i \BFw_i^\T ) \leq 2$ with probability at least $1-(1/2)e^{-u} $ as long as $n\gtrsim \upsilon_1^4(p+u)$. Moreover, applying
Theorem 2.1 in \cite{HKZ2012} to each $\BFw_{i,-}$ yields that  with probability at least $1-e^{-z}$,
\$
 \| \BFw_{i,-} \|_2^2 \leq \upsilon_1^2 \Big\{   p-1 + 2\sqrt{(p-1) z} + 2 z  \Big\} 
 \leq \upsilon_1^2 \Big( \sqrt{p-1} + \sqrt{2 z} \Big)^2 .
\$
Taking the union bound over $i=1,\ldots, n$ and setting $z= \log n + u$, we obtain that with probability $1- e^{-u}$,
\$
	\max_{1\leq i\leq n} \| \Sigma^{-1/2} \BFx_i \|_2 = \max_{1\leq i\leq n} \| \BFw_i  \|_2  \leq 1 + \upsilon_1  \Big\{ \sqrt{p-1} + \sqrt{2 \log n +2 u  } \Big\} \leq C \upsilon_1 ( p + \log n + u )^{1/2} .
\$

Turning to $ \max_{1\leq j\leq N_\epsilon }  \| \hat \BFH_\varpi(\BFv_j)  - \hat  \BFH_\varpi(\BFv_j)  \|_2$, it has been shown in the proof of Proposition~3.2 in \cite{HPTZ2020} (see, e.g. (C.61)) that with probability at least $1- e^{-u}$,
\$
 \max_{1\leq j\leq N_\epsilon }  \| \hat \BFH_\varpi(\BFv_j)  - \hat  \BFH_\varpi(\BFv_j)  \|_2 \lesssim  \upsilon_1^2 \bigg\{ \sqrt{\frac{p \log(R/\epsilon )   + u}{n \varpi }} + \frac{p \log(R/\epsilon )  + u}{n \varpi }   \bigg\}.
\$
Finally, as long as $n\gtrsim \upsilon_1^4(p+u)$, taking $\epsilon = (\varpi/n)^2R$ yields the bound
\$
& \sup_{\BFv \in \BB^p(R)}  \| \hat \BFH_\varpi(\BFv)  - \BFH_\varpi(\BFv)  \|_2 \\ 
&\lesssim   \upsilon_1^2 \bigg\{ \sqrt{\frac{p \log(n /\varpi  )   + u}{n \varpi }} + \frac{p \log(n/\varpi )  + u}{n \varpi }   \bigg\}   +      \big\{ \upsilon_1 (p+ \log n+ u)^{1/2}   + l_0  m_3  \varpi^2 \big\}  \frac{R }{n^2}.
\$
This proves \eqref{hessian.unif.convergence}, and thus completes the proof.  \Halmos

\subsection{Proof of Lemma~\ref{lem:max.normal.l2}}

For each $t=0,1,\ldots, T-1$,  we apply the concentration inequality for Lipschitz functions of standard normal random variables and obtain that
\$
 P \big( \| \BFg_t\|_2   \geq  \sqrt{p} + \sqrt{2z} \big) \leq e^{-z}, \ \ \mbox{ valid for any } z\geq 0.	
\$
Combining this with the union bound (over $t=0,1,\ldots, T-1$) yields \eqref{max.chi-square.concentration}.

Note that  $\| \BFg_t \|_2^2$ follows the chi-square distribution $\chi_p^2$, which is a special case of the gamma distribution $\Gamma(p/2,1/2)$. The centered variable, $\| \BFg_t \|_2^2- p$, is known to be sub-gamma with parameters $v=2p$ and $c=2$ \citep{BLM2013}.  Let $Z= \sum_{t=0}^{T-1} \rho^t  ( \| \BFg_t\|_2^2 - p )$. For each $t$ and $0<\lambda <1/c$,
\$
 \log \E e^{\lambda\rho^t  ( \| \BFg_t\|_2^2 - p ) } \leq \frac{v \lambda^{2} \rho^{2t} }{2(1 - c \lambda\rho^t )} \leq \frac{v \lambda^2 \rho^{2t} }{2(1-c\lambda)}.
\$
By independence,
\$
\log \E e^{\lambda Z } = \sum_{t=0}^{T-1}  \log \E e^{\lambda\rho^t  ( \| \BFg_t\|_2^2 - p ) } \leq \frac{v \lambda^2  \sum_{t=0}^{T-1}\rho^{2t} }{2(1-c\lambda)} \leq \frac{v}{1-\rho^2} \frac{  \lambda^2  }{2(1-c\lambda)} .
\$
In other words, the centered variable $Z$ is sub-gamma with parameters $(v/(1-\rho^2), c)=(2p/(1-\rho^2), 2)$. Applying Chernoff's bound to $Z$ (see, e.g. Section~2.4 of \cite{BLM2013}) yields that, for any $z>0$,
\$
 P \bigg(  Z > 2 \sqrt{\frac{ p z }{1-\rho^2}} + 2 z  \bigg) \leq e^{-z}.
\$
This, combined with the elementary inequality $\sum_{t=0}^{T-1} \rho^t \leq 1/(1-\rho)$, proves \eqref{weighted.chi-square.concentration}. \Halmos

 \subsection{Proof of Lemma~\ref{lem:GD.compact}}
 
To begin with, note that conditioned on event $\cE_0(B) \cap \cE_1 (\delta_0, \delta_1)$ with $\delta_0 + b^* < \phi_1$,  
\$
\sup_{\BFbeta \in  \Theta_\Sigma^*(1) } \| \nabla \hQ_\varpi(\BFbeta) \|_{\Sigma^{-1}}  & \leq \delta_1+ \bar \tau  , \\
\hQ_\varpi(  \BFbeta  ) -  \hQ_\varpi( \BFbeta^*  ) & \geq ( \underbrace{ \phi_1 - \delta_0  - b^* }_{= \Delta }) \| \BFbeta - \BFbeta^* \|_{\Sigma} ~\mbox{ for all }~ \BFbeta \in \Theta_\Sigma^*(1)^\cc
\$
and $\BFbeta^{(t+1)} = \BFbeta^{(t)} - \eta_0 \Sigma^{-1} \nabla \hQ_\varpi(\BFbeta^{(t)})  -   \eta_0  \sigma  \Sigma^{-1/2}\BFg_t / n$ for $t=0,1,\ldots, T-1$.

Now assume that $ \| \BFbeta^{(t)}  -\BFbeta^* \|_{\Sigma} \leq 1$ for some $t\geq 0$. 
Recall that $\delta_1 < \phi_1 < 0.5 f_l$. Hence, conditioned further on $ \cG$ defined in \eqref{def:G}, we have
\$
 \|   \BFbeta^{ (t+1)} - \BFbeta^* \|_{\Sigma}  & \leq \| \BFbeta^{(t)} - \BFbeta^* \|_{\Sigma} + \eta_0  \| \nabla \hQ_\varpi(\BFbeta^{(t)} ) \|_{{\Sigma}^{-1}} +  \frac{ \eta_0 \sigma }{ n }  \| \BFg_t\|_2  \\
 & \leq 1 + (   f_l +  \bar \tau )  \eta_0     \leq 2  ,
\$
where we used the assumptions $\eta_0 \leq 1/ (f_l + \bar \tau)$ and $  n \geq 2  B_T \sigma/f_l$ in the last inequality.  Proceeding via proof by contradiction,   suppose $\|  \BFbeta^{(t+1)} - \BFbeta^* \|_{\Sigma} > 1$ so that
\$
    \Delta \cdot  \| \BFbeta^{(t+1)} - \BFbeta^* \|_{\Sigma} \leq \hQ_\varpi(   \BFbeta^{(t+1)} ) -  \hQ_\varpi( \BFbeta^*  ) .
\$
For the right-hand side, we have
\$
     &   \hQ_\varpi(   \BFbeta^{(t+1)} ) -  \hQ_\varpi( \BFbeta^*  ) =  \hQ_\varpi(  \BFbeta^{(t+1 )} ) - \hQ_\varpi(\BFbeta^{(t)} ) + \hQ_\varpi(\BFbeta^{(t)} ) -  \hQ_\varpi( \BFbeta^*  )  \\
 & \stackrel{({\rm i})}{\leq } \langle  \nabla \hQ_\varpi(\BFbeta^{(t)} ),  \BFbeta^{(t+1)} - \BFbeta^{(t)}  \rangle + f_u  \|  \BFbeta^{(t+1)} - \BFbeta^{(t)}  \|_{\Sigma}^2 - \langle \nabla \hQ_\varpi(\BFbeta^{(t)} ) , \BFbeta^* - \BFbeta^{(t)}  \rangle \\
 & = \frac{1}{\eta_0} \langle     \BFbeta^{(t)}  -    \BFbeta^{(t+1)}   ,    \BFbeta^{(t+1)} - \BFbeta^* \rangle_{\Sigma} + f_u  \|   \BFbeta^{(t+1)} - \BFbeta^{(t)}  \|_{\Sigma}^2  +  \frac{\sigma }{ n}\langle   \Sigma^{-1/2}  \BFg_t  ,    \BFbeta^{(t+1)} - \BFbeta^* \rangle   \\ 
 & = \frac{1}{2 \eta_0} \| \BFbeta^{(t)}  - \BFbeta^* \|_{\Sigma}^2 - \frac{1}{2\eta_0 } \|   \BFbeta^{(t+1)} - \BFbeta^* \|_{\Sigma}^2 - \frac{1}{2 \eta_0} \|   \BFbeta^{(t+1)} - \BFbeta^{(t)}  \|_{\Sigma}^2  \\ 
 &~~~~+ f_u   \|   \BFbeta^{(t+1)} - \BFbeta^{(t)}  \|_{\Sigma}^2 +     \frac{ \sigma }{ n} \langle   \Sigma^{-1/2}  \BFg_t  ,    \BFbeta^{(t+1)} - \BFbeta^* \rangle  \\
 & \stackrel{({\rm ii})}{\leq }  \frac{1}{2 \eta_0} \| \BFbeta^{(t)}  - \BFbeta^* \|_{\Sigma}^2 - \frac{1}{2\eta_0 } \|  \BFbeta^{(t+1)} - \BFbeta^* \|_{\Sigma}^2 +     \frac{  \sigma}{  n} \|   \BFbeta^{(t+1)} - \BFbeta^* \|_{\Sigma} \cdot    \|  \BFg_t \|_2  \\
 & \stackrel{({\rm iii})}{\leq }   \frac{1}{2 \eta_0} \| \BFbeta^{(t)}  - \BFbeta^* \|_{\Sigma}^2 - \frac{1}{2\eta_0 } \|  \BFbeta^{(t+1)} - \BFbeta^* \|_{\Sigma}^2 +  \frac{ B_T \sigma }{ n}   \|   \BFbeta^{(t+1)} - \BFbeta^* \|_{\Sigma} ,
\$
where inequality (i) follows from the restricted smoothness property (\ref{restricted.smootheness}), inequality (ii) holds if $\eta_0 \leq 1/(2 f_u)$, and inequality (iii) uses conditioning on $ \cG$.  
Provided that $B_T \sigma/ n \leq \Delta$,  combining the above lower and upper bounds on $   \hQ_\varpi(   \BFbeta^{(t+1)} ) -  \hQ_\varpi( \BFbeta^*  ) $  yields
\$
  \|   \BFbeta^{(t+1)} - \BFbeta^* \|_{\Sigma}^2 \leq  \| \BFbeta^{(t)}  - \BFbeta^* \|_{\Sigma}^2  \leq  1 ,
\$ 
which leads to a contradiction. Therefore, starting from an initial value $\BFbeta^{(0)} \in \Theta^*(1)$,  and conditioning on event $\cE_0 \cap  \cE_1 \cap \cG$ with properly chosen parameters, we must have $\| \BFbeta^{(t )} - \BFbeta^* \|_{\Sigma}\leq 1$ for all $t= 1,\ldots, T$. \Halmos

\subsection{Proof of Lemma~\ref{lem:initial.value}}
 
Recall that conditioning on $\cE_0 \cap \cE_1$,  $\hat \BFbeta_\varpi \in \Theta^*_\Sigma(r_0)$ with $r_0 = (\delta_0 + b^*)/\phi_1$, $\BFbeta^{(t+1)} = \BFbeta^{(t)} - \eta_0 \Sigma^{-1} \nabla \hQ_\varpi(\BFbeta^{(t)})  +  \eta_0  \sigma  \Sigma^{-1/2}\BFg_t / n$ and $\| \nabla^2 \hQ_\varpi(\BFbeta ) \|_{\Sigma^{-1}} \leq 2 f_u$ for all $\BFbeta \in \Theta^*_\Sigma(R)$. The upper bound on the Hessian also implies
\$
	\langle \nabla \hQ_\varpi(\BFbeta_1) -  \nabla \hQ_\varpi(\BFbeta_2) , \BFbeta_1 - \BFbeta_2 \rangle \geq \frac{1}{2f_u } \|  \nabla \hQ_\varpi(\BFbeta_1) -  \nabla \hQ_\varpi(\BFbeta_2) \|_{\Sigma^{-1}}^2 , \ \ \BFbeta_1, \BFbeta_2 \in  \Theta^*_\Sigma(R) .
\$
Provided  $\eta_0 \leq 1/(2f_u)$ and $R \geq R_0 + r_0$,  applying this bound with $(\BFbeta_1, \BFbeta_2) = (\BFbeta^{(0)} ,  \hat \BFbeta_\varpi  )$ we obtain  
\$
	 \| \BFbeta^{(1)} - \hat \BFbeta_\varpi \,  \|_\Sigma^2  & =  \| \BFbeta^{(0)}  -   \eta_0 \Sigma^{-1} \nabla \hQ_\varpi(\BFbeta^{(0)})  +  \eta_0  \sigma  \Sigma^{-1/2}\BFg_0  / n- \hat \BFbeta_\varpi  \,  \|_\Sigma^2  \\
	 &= \| \BFbeta^{(0)} - \hat \BFbeta_\varpi  \,  \|_\Sigma^2   + \eta_0^2  \| \Sigma^{-1/2} \nabla \hQ_\varpi(\BFbeta^{(0)}) - \sigma \BFg_0 / n \|_2^2  \\
	 &~~~~~- 2 \eta_0 \big\langle  \BFbeta^{(0)} -\hat \BFbeta_\varpi , \nabla \hQ_\varpi(\BFbeta^{(0)})  - \sigma   \Sigma^{ 1/2}\BFg_0 / n \big\rangle \\
	 & \leq \| \BFbeta^{(0)} - \hat \BFbeta_\varpi   \,  \|_\Sigma^2  + \eta_0^2  \| \nabla \hQ_\varpi(\BFbeta^{(0)})  \|_{\Sigma^{-1}}^2 +      \bigg( \frac{\eta_0 \sigma}{n} \bigg)^2 \| \BFg_0 \|_2^2 +   2 \frac{\eta_0^2 \sigma}{n}  \| \nabla \hQ_\varpi(\BFbeta^{(0)})  \|_{\Sigma^{-1}} \| \BFg_0 \|_2     \\
	 &~~~~~  - \frac{\eta_0}{f_u} \|  \nabla \hQ_\varpi(\BFbeta^{(0)})  \|_{\Sigma^{-1}}^2  + 2 \frac{ \eta_0 \sigma }{n}  \| \BFbeta^{(0)} - \hat \BFbeta_\varpi  \,  \|_\Sigma \| \BFg_0 \|_2 \\
	 & \leq \| \BFbeta^{(0)} - \hat \BFbeta_\varpi   \,  \|_\Sigma^2 - \frac{\eta_0}{2 f_u }\|  \nabla \hQ_\varpi(\BFbeta^{(0)})  \|_{\Sigma^{-1}}^2   +       2 (R_0 +   \eta_0 \bar \tau B)   \frac{  \eta_0 \sigma }{n}   \| \BFg_0 \|_2 + \bigg( \frac{\eta_0 \sigma}{n} \bigg)^2 \| \BFg_0 \|_2^2   .
\$

For any given $T_0\geq 1$,  write $R_t = \| \BFbeta^{(t)} - \hat \BFbeta_\varpi   \|_\Sigma$ for $t=1, \ldots, T_0$.  Provided $R \geq \max_{0\leq t\leq T_0-1} R_t + r_0$, it can similarly shown that for any $0\leq t\leq T_0-1$,
\$
	& \| \BFbeta^{(t+1)} - \hat \BFbeta_\varpi   \|_\Sigma^2   \\
 & \leq \| \BFbeta^{(t)} - \hat \BFbeta_\varpi   \,  \|_\Sigma^2 - \frac{\eta_0}{2 f_u }\|  \nabla \hQ_\varpi(\BFbeta^{(t)})  \|_{\Sigma^{-1}}^2   +       2 (R_t +   \eta_0 \bar \tau B)   \frac{  \eta_0 \sigma }{n}   \| \BFg_t \|_2 + \bigg( \frac{\eta_0 \sigma}{n} \bigg)^2 \| \BFg_t \|_2^2  .
\$
Given $z\geq 0$,  by Lemma~\ref{lem:max.normal.l2} we have that with probability at least $1-e^{-z}$,
\$	
	\max_{0\leq t\leq T_0-1} \| \BFg_t \|_2 \leq B_{T_0} := \sqrt{p} + \sqrt{2(\log T_0 + z ) } .
\$
Define $e_{{\rm priv}} = \eta_0 B_{T_0} \sigma/n$. 
For some $\epsilon\in(0,1)$ to be determined, the above recursive bound implies
\$
	R_{t+1}^2 \leq (1+\epsilon) R_t^2 + (1+1/\epsilon)  e_{{\rm priv}}^2 + C_0 e_{{\rm priv}} , \ \ t = 0, 1, \ldots, T_0 -1
\$
and hence
\$
R_t^2 &  \leq (1+\epsilon)^{t} R_0^2 + \big\{ (1+1/\epsilon) e_{{\rm priv}}^2 + C_0 e_{{\rm priv}} \big\} \sum_{k=0}^{t-1} (1+\epsilon)^k   \\
& \leq (1+\epsilon)^{t} R_0^2 + \frac{(1+\epsilon)^{t}  -1}{\epsilon} \big\{ (1+1/\epsilon) e_{{\rm priv}}^2 + C_0 e_{{\rm priv}} \big\}, \ \ t= 1, \ldots, T_0, 
\$
where $C_0 = 2 \eta_0 \bar \tau B$. Provided $T_0\geq 2$ and 
$$
	n\geq    \frac{e-1}{4-e}  ( 2  \bar \tau B + 1/2 ) \max\{ 1 ,  \eta_0/R_0 \}^2 T_0  B_{T_0} \sigma >  2 (T_0+1)    B_{T_0} \sigma ,
$$
taking $\epsilon=1/T_0 \in (0,1)$ we obtain that 
\$
 R_t^2&  \leq e R_0^2 + (e-1)    \big\{  (T_0+1) e_{{\rm priv}}  + C_0\big\}  T_0 e_{{\rm priv}}  \\
 & \leq e R_0^2 + (e-1)   ( 2   \bar \tau B  +1/2 )  \frac{ T_0  B_{T_0} \sigma }{n}  \eta_0^2  \leq e R_0^2 + (4-e) R_0^2 = 4 R_0^2
\$
for all $t=1, \ldots, T_0$, as claimed. \Halmos

 \end{APPENDICES}




\bibliographystyle{informs2014}
\bibliography{ref1}


\end{document}